\documentclass[twoside]{article}

\usepackage[utf8]{inputenc}
\usepackage[T1]{fontenc}
\usepackage{url}
\usepackage{booktabs}
\usepackage{amsfonts}
\usepackage{nicefrac}
\usepackage{microtype}
\usepackage[hidelinks]{hyperref} 
\hypersetup{
  colorlinks,
  citecolor=blue,
  linkcolor=blue,
  urlcolor=blue}
\usepackage[inline]{enumitem}
\usepackage[table, dvipsnames]{xcolor}
\usepackage{graphicx}
\usepackage{amsmath}
\usepackage{float}
\usepackage{adjustbox}
\usepackage{caption} 
\usepackage{subcaption}
\usepackage{mathtools, nccmath}
\usepackage{tikz}
\usepackage{algorithm}
\usepackage[algo2e, ruled,vlined,boxed,linesnumbered]{algorithm2e}
\usepackage[noend]{algorithmic}
\SetArgSty{textnormal}
\usepackage{listings}
\usepackage[colaction]{multicol}
\usepackage{lineno}

\newcommand\multicollinenumbers{%
 \linenumbers
 \def\makeLineNumber{\docolaction{}{}{}}}

\usepackage{wrapfig}
\usepackage{enumitem}
\usepackage{makecell}
\usepackage{wrapfig}

\definecolor{darkblue}{HTML}{1A254B}
\definecolor{lightblue}{HTML}{A7BED3}
\definecolor{blue}{HTML}{114083}
\definecolor{green}{HTML}{81B5AE}
\definecolor{pink}{HTML}{F2545B}
\definecolor{red}{HTML}{A4243B}

\newcommand{\defeq}{\vcentcolon=}

\def\Ncal{\mathcal{N}}

\def\RR{\mathbb{R}}

\DeclarePairedDelimiterX{\dotp}[2]{\langle}{\rangle}{#1, #2}

\usepackage[toc,page]{appendix}      % for appendix

\usepackage{amsmath}		% for AMS macros
\usepackage{amssymb}		% for AMS symbols
\usepackage{amsfonts}		% for AMS fonts
\usepackage{amsthm}		% for theorems
\usepackage{mathtools}		% for advanced math

\usepackage{acronym}		% for acronyms
\newcommand{\acli}[1]{\textit{\acl{#1}}}		% for italicized acro
\newcommand{\aclip}[1]{\textit{\aclp{#1}}}		% for italicized acro (plural)
\newcommand{\acdef}[1]{\textit{\acl{#1}} \textup{(\acs{#1})}\acused{#1}}		% for acro def
\newcommand{\acdefp}[1]{\textit{\aclp{#1}} \textup{(\acsp{#1})}\acused{#1}}	% for acro def (plural)

\usepackage{makecell}

\newcommand{\afterhead}{.}
\usepackage{titlesec}

\newcommand{\para}[1]{\paragraph{\textbf{#1\afterhead}}}

\usepackage{quoting}			% for managing spaces with quotations
\quotingsetup{vskip=\medskipamount}

\usepackage{stmaryrd}		% for extra symbols
\usepackage{wasysym}		% for extra symbols

\usepackage[sort&compress,capitalize,nameinlink]{cleveref}		% for cleveref formatting
\crefname{assumption}{Assumption}{Assumptions}
\crefname{algo}{Algorithm}{Algorithms}
\crefname{example}{Example}{Examples}
\crefname{method}{Method}{Methods}
\creflabelformat{assumption}{\upshape(#2#1#3\upshape)}

\crefname{assumptionenum}{Assumption}{Assumptions}
\creflabelformat{assumptionenum}{#2#1#3}

\crefname{item}{}{}
\creflabelformat{item}{#2#1#3}

\crefname{eq}{}{}
\creflabelformat{eq}{\upshape(#2#1#3\upshape)}

\usepackage{thmtools}		% for theorem tools
\usepackage{thm-restate}		% for restating theorems

\newtheorem{theorem}{Theorem}		% for theorems
\newtheorem{corollary}{Corollary}		% for corollaries
\newtheorem{lemma}{Lemma}		% for lemmas
\newcommand{\debug}[1]{#1}		% for removing macro coloring

\newcommand{\newmacro}[2]{\newcommand{#1}{\debug{#2}}}		% for shorthand definitions
\DeclarePairedDelimiter{\braces}{\{}{\}}		% for braces
\DeclarePairedDelimiter{\bracks}{[}{]}		% for brackets
\DeclarePairedDelimiter{\parens}{(}{)}		% for parentheses

\DeclarePairedDelimiterX{\inner}[2]{\langle}{\rangle}{#1, #2}		% for scalar product
\DeclarePairedDelimiter{\norm}{\lVert}{ \rVert}		% for norm
\DeclarePairedDelimiterXPP{\twonorm}[1]{}{\lVert}{\rVert}{}{#1}		% for L2 norm
\DeclarePairedDelimiterXPP{\dnorm}[1]{}{\lVert}{\rVert}{_{\ast}}{#1}		% for dual norm
\DeclarePairedDelimiterX{\braket}[2]{\langle}{\rangle}{#1,#2}		% for brakets
\DeclarePairedDelimiterX{\setdef}[2]{\{}{\}}{#1:#2}		% for set builder notation
\DeclarePairedDelimiterXPP{\exclude}[1]{\mathopen{}\setminus}{\{}{\}}{}{#1}

\newcommand{\alt}[1]{#1'}		% for alternates

\newcommand{\R}{\mathbb{R}}		% for reals
\newcommand{\C}{\mathbb{C}}		% for complex numbers (may clash)

\DeclareMathOperator{\dist}{dist}		% for distance
\DeclareMathOperator{\tr}{tr}		% for trace
\newmacro{\coef}{\lambda}		% for coefficient
\newmacro{\dd}{\:\mathrm{d}}		% for integrators
\newmacro{\intR}{\int_{\R^{\vdim}}}		% for integration over full domains
\newmacro{\intRR}{\int_{\R^{\vdim}  \times \R^{\vdim}  }}		% for integration over double full domains
\newmacro{\nn}{\nonumber}		% for equations

\newcommand{\subs}{\leftarrow}      % for substitution

\newcommand{\ddt}{\frac{\mathrm{d}}{\mathrm{d}t}}		% for Leibniz
\newcommand{\del}{\partial}		% for derivatives
\newcommand{\eps}{\varepsilon}		% for better epsilon
\newmacro{\pexp}{p}		% for first exponent
\newmacro{\qexp}{q}		% for second exponent
\newmacro{\rexp}{r}		% for third exponent

\newcommand{\const}{ \mathrm{const.} }

\newcommand{\cf}{cf.\xspace}		% for consistency
\newcommand{\eg}{e.g.,\xspace}		% for consistency
\newcommand{\ie}{i.e.,\xspace}		% for consistency
\newmacro{\mfd}{\mathcal{M}}		% for metric tensor
\newmacro{\curve}{\gamma}          % for curves
\newmacro{\sect}{\mathcal{K}}    % for sectional curvatures

\newcommand{\from}{\colon}		% for function definition
\newcommand{\eqdef}{\eqqcolon}		% for reverse definition

\newmacro{\sset}{\mathcal{S}}		% for generic set

\newmacro{\points}{\mfd}		% for Riemannian RM
\newmacro{\intpoints}{\points^{\circ}}		%for point set interior
\newmacro{\point}{x}		% for generic point
\newmacro{\pointalt}{\alt\point}		% for alternate point

\newmacro{\dpoints}{\mathcal{W}}		% for second point set (duals, etc.)
\newmacro{\dpoint}{w}		% for second generic point
\newmacro{\dpointalt}{\alt\dpoint}		% for second alternate variable

\newmacro{\base}{p}		% for reference point
\newmacro{\basealt}{q}		% for alternate reference point

\newmacro{\open}{\mathcal{U}}		% for open sets
\newmacro{\closed}{\mathcal{C}}		% for closed sets
\newmacro{\cpt}{\mathcal{K}}		% for compact sets
\newmacro{\nbhd}{\mathcal{U}}		% for neighborhoods

\newmacro{\start}{1}		% for start index
\newmacro{\halfafterstart}{3/2}		% for second index
\newmacro{\afterstart}{2}		% for second index
\newmacro{\running}{\start,\afterstart,\dotsc}		% for running index
\newmacro{\halfrunning}{\start,\halfafterstart,\dotsc}

\newmacro{\runalt}{k}		% for running sequence index
\newmacro{\run}{n}		% for main sequence index
\newmacro{\nRuns}{T}		% for total number of runs
\newmacro{\runs}{\mathcal{\nRuns}}		% for set of runs

\newmacro{\state}{Z}		% for main iterate
\newmacro{\dstate}{Y}		% for other iterate

\newmacro{\vecspace}{\R^{\vdim}}		% for generic vector space

\newmacro{\coord}{i}		% for index
\newmacro{\vdim}{d}		% for dimension
\newmacro{\vvec}{v}		% for generic vector
\newmacro{\bvec}{e}		% for basis vector
\newmacro{\bvecs}{\mathcal{E}}		% for basis vectors

\newmacro{\subspace}{\mathcal{W}}		% for subspace
\newmacro{\wvec}{w}		% for generic subspace vector
\newmacro{\subdim}{m}		% for subspace dimension

\newmacro{\tanhull}{\mathcal{Z}}		% for tangent hull
\newmacro{\tanvec}{z}		% for tangent vectors

\newcommand{\dual}[1]{#1^{\ast}}		% for dual variables
\newmacro{\dspace}{\dual\vecspace}		% for dual space
\newmacro{\dvec}{v}		% for dual vector
\newmacro{\dbvec}{\eps}		% for dual basis vectors

\newmacro{\mat}{M}		% for generic matrix
\newmacro{\eye}{I}		% for identity matrix

\newcommand{\mg}{\succ}		% for positive-definite
\newcommand{\SPD}{\mathbb{S}_{++}^\vdim}		% for Symmetric Positive Definite cone 

\DeclareMathOperator{\ex}{\mathbb{E}}		% for expectations
\DeclareMathOperator{\prob}{\mathbb{P}}		% for probability
\newmacro{\seed}{\omega}		% for seed
\newmacro{\seeds}{\Omega}		% for seed space
\newmacro{\history}{\mathcal{H}}		% for filtrations

\newmacro{\sample}{\omega}		% for samples
\newmacro{\samples}{\Omega}		% for sample space
\newmacro{\filter}{\mathcal{F}}		% for filtrations
\newmacro{\probspace}{(\samples,\filter,\prob)}		% for probability space

\newmacro{\event}{\mathcal{E}}       % for event
\newmacro{\eventalt}{\mathcal{H}}       % for alternate event
\newmacro{\mean}{\mu}		% for mean of distribution
\newmacro{\sdev}{\sigma}		% for mean of distribution
\newmacro{\variance}{\sdev^{2}}		% for mean of distribution

\newmacro{\dkl}{D_{\mathrm{KL}}}		% for Kullback Leibler
\providecommand\given{\vert}		% empty command for conditionals

\DeclarePairedDelimiterXPP{\exof}[1]{\ex}{[}{]}{}{%		% for conditional expectations
\renewcommand\given{\nonscript\,\delimsize\vert\nonscript\,\mathopen{}} #1}

\DeclarePairedDelimiterXPP{\probof}[1]{\prob}{(}{)}{}{%		% for conditional probabilities
\renewcommand\given{\nonscript\:\delimsize\vert\nonscript\:\mathopen{}} #1}

\newmacro{\gmat}{g}		% for metric tensor
\newmacro{\gdist}{\dist_{\gmat}}
\newmacro{\ball}{\mathbb{B}}		% for balls
\newmacro{\sphere}{\mathbb{S}}		% for spheres

\newmacro{\mbase}{\mu}
\newmacro{\m}{\mbase_0}     % for mean of the first gaussian
\newmacro{\malt}{\mbase_\horizon}     % for mean of the second gaussian

\newmacro{\covarbase}{\Sigma}
\newmacro{\covar}{\covarbase_0}     % for covariance of the first gaussian
\newmacro{\covaralt}{\covarbase_\horizon}     % for covariance of the first gaussian

\newmacro{\ctime}{t}
\newmacro{\ctimealt}{s}
\newmacro{\horizon}{1}

\newmacro{\ratiosym}{r}    % for symbol of the the ratio between \ctime and \horizon
\newcommand{\ratio}[1][\ctime]{\ratiosym_{#1}}     % for ratio between \ctime and \horizon
\newcommand{\ratioc}[1][\ctime]{\bar{\ratiosym}_{#1}}     % for complementary ratio between \ctime and \horizon

\newmacro{\scalingbase}{\mathrm{QV}}     % for quadratic variation

\newmacro{\KLbase}{D_{\mathrm{KL}}}
\newcommand{\KL}[2]{ \KLbase\parens*{ #1 \Vert #2 } }

\newmacro{\sdebase}{ X }
\newcommand{\sde}[1][\ctime]{ \sdebase_{ #1 } }     % for marginal variables
\newcommand{\dsde}[1][\ctime]{ \dd\sdebase_{ #1 } }     % for marginal variables
\newmacro{\tinv}{\tau}

\newmacro{\testfbase}{u}
\newmacro{\generator}{\mathcal{L}_{\ctime}}

\newcommand{\Laplace}{\Delta}

\usepackage{xparse}
\NewDocumentCommand{\testf}{ O{\ctime} O{\point} }{ \testfbase\parens*{#1,#2} }

\newmacro{\dconst}{\lambda}     % for constant drift
\newmacro{\sconst}{\mathbf{v}}     % for constant shift
\newmacro{\vconst}{\omega}     % for constant volatility

\newmacro{\qvbase}{\mathrm{q}}

\newcommand{\qv}[1][\ctime]{\qvbase\parens*{ #1 } }
\newcommand{\qve}[1][\horizon]{\qvbase\parens*{ #1 } }
\newcommand{\dqv}[1][\ctime]{ \dot{\qvbase}\parens*{ #1 } }

\newmacro{\subVPfbase}{ \beta }

\newcommand{\aggtimeqr}[1][\ctime]{ \aggtimebase^4_{ #1 } }

\newmacro{\refsdebase}{ Y }     % for ref SDEs
\newcommand{\refsde}[1][\ctime]{ \refsdebase_{ #1 } }
\newcommand{\drefsde}[1][\ctime]{ \dd \refsdebase_{ #1 } }

\newmacro{\wiescalebase}{ g }

\newmacro{\QVbase}{ \mathrm{qv} }

\newmacro{\driftbase}{  c  }
\newcommand{\drift}[1][\ctime]{ \driftbase_{ #1 }  }%{ \driftbase\parens*{ #1 }  }

\newmacro{\shiftbase}{  \alpha  }
\newcommand{\shift}[1][\ctime]{ \shiftbase_{ #1 }  }

\newmacro{\volatbase}{  g  }
\newcommand{\volat}[1][\ctime]{ \volatbase_{#1}  }%{ \volatbase\parens*{ #1 }  }
\newcommand{\volatsq}[1][\ctimealt]{ \volatbase^2_{ #1 }  }%{ \volatbase^2\parens*{ #1 }  }

\newmacro{\refprobase}{\mathbb{Q}}     % for alternative general stochastic processes alphabet
\newcommand{\refpro}[1][\ctime]{\refprobase_{ #1 }}
\newmacro{\refjoint}{\refprobase_{\mathrm{0\horizon}} }    % for alternative general stochastic processes alphabet
\newcommand{\refapprox}[1][\ctime]{ {\refprobase}^{\solbase}_{ #1 }}

\newmacro{\Wienerbase}{\mathbb{W}}     % for reversible Wiener processes alphabet
\newcommand{\Wiener}[1][\ctime]{ \Wienerbase_{#1} }     % for reversible Wiener processes
\newcommand{\dWiener}[1][\ctimealt]{ \dd \Wienerbase_{#1} }     % for Wiener processes increments

\newmacro{\aggtimebase}{  \tau  }

\newcommand{\aggtime}[1][\ctime]{ \aggtimebase_{ #1 }  }
\newcommand{\aggtimeinv}[1][\ctimealt]{ \aggtimebase^{\ssstyle -1}_{ #1 }  }
\newcommand{\aggtimesq}[1][\ctimealt]{ \aggtimebase^2_{ #1 }  }
\newcommand{\aggtimesqinv}[1][\ctimealt]{ \aggtimebase^{-2}_{ #1 }  }
\newcommand{\daggtime}[1][\ctime]{ \dot{\aggtimebase}_{ #1 }  }
\newmacro{\mrsdebase}{ \eta}
\NewDocumentCommand{\mYcinit}{ O{\ctime}  }{  \mrsdebase\parens*{#1 } } %  \middle| \refsde[0] } }

\newmacro{\kernelbase}{ \kappa}
\NewDocumentCommand{\kernel}{ O{\ctime} O{\ctime'}  }{  \kernelbase\parens*{#1, #2} }% \middle| \refsde[0] } }

\newmacro{\intdasq}{ \int_0^\ctime {\aggtimesqinv[\ctimealt]}{\volatsq} \dd \ctimealt }
\newmacro{\intdasqT}{ \int_0^\horizon {\aggtimesqinv[\ctimealt]}{\volatsq} \dd \ctimealt }

\newmacro{\ssstyle}{\scriptscriptstyle}
\newmacro{\sssNcal}{\ssstyle\Ncal}

\newmacro{\solbase}{}
\newmacro{\Cstar}{C_{\sdev_{\solbase}}}

\newmacro{\pbase}{\mathbb{P}}     % for general stochastic processes alphabet
\newmacro{\Pinit}{\pbase_{{0}}}     % for initial marginal
\newmacro{\Pend}{\pbase_{{\horizon}}}     % for end marginal
\newcommand{\Pmargin}[1][\ctime]{ \pbase_{{#1}} }     % for general marginals
\newcommand{\Psol}[1][\ctime]{ \pbase^{\solbase}_{ #1 } }
\newmacro{\Pjoint}{ \pbase_{ \mathrm{0\horizon}} }

\newmacro{\distbase}{ \pbase }
\newmacro{\ini}{ {0} }
\newmacro{\distinit}{ \hat{\distbase}_{ \ini } }
\newmacro{\en}{ {\horizon} }
\newmacro{\distend}{ \hat{\distbase}_{ \en} }

\newcommand{\Xsol}[1][\ctime]{ X^{\solbase}_{ #1 } }

\newcommand{\dXsol}[1][\ctime]{ \dd X^{\solbase}_{ #1 } }

\newcommand{\meansol}[1][\ctime]{ \mu^{\solbase}_{ #1 } }
\newcommand{\dmeansol}[1][\ctime]{ \dot{\mu}^{\solbase}_{ #1 } }
\newcommand{\Sigmasol}[1][\ctime]{ \Sigma^{\solbase}_{ #1 } }
\newcommand{\Sigmasolinv}[1][\ctime]{ \Sigma^{\solbase-1}_{ #1 } }
\newcommand{\dSigmasol}[1][\ctime]{ \dot{\Sigma}^{\solbase}_{ #1 } }

\newcommand{\dratio}[1][\ctime]{\dot{\ratiosym}_{#1}}     % for ratio between \ctime and \horizon
\newcommand{\dratioc}[1][\ctime]{\dot{\bar{\ratiosym}}_{#1}}     % for complementary ratio between \ctime and \horizon

\newcommand{\cmeansol}[2]{ \mu^{\solbase}_{ #1 \vert #2 } }
\newcommand{\cSigmasol}[2]{ \Sigma^{\solbase}_{ #1 \vert #2 } }

\newmacro{\efftrbase}{\rho}    % for effectively scaling
\newcommand{\efftr}[1][\ctime]{  \efftrbase_{ #1 }  }

\newmacro{\paramf}{ \theta }
\newmacro{\SBfbase}{ Z }
\newmacro{\paramb}{ \phi }
\newmacro{\SBbbase}{ \hat{\SBfbase} }
\NewDocumentCommand{\SBf}{ O{\ctime} O{\point} O{\paramf} }{ \SBfbase_{#1}^{#3}\parens{#2} }
\NewDocumentCommand{\SBb}{ O{\ctime} O{\point} O{\paramb} }{ \SBbbase_{#1}^{#3}\parens{#2} }

\newmacro{\GSBfbase}{ f_{\sssNcal} } %f_{\scriptscriptstyle\textup{GSB}} }
\newmacro{\GSBbbase}{ \hat{\GSBfbase}}%^{\scriptscriptstyle\textup{rev}} }
\NewDocumentCommand{\GSBf}{ O{\ctime} O{\point} }{ \GSBfbase\parens*{#1,#2} }

\newmacro{\tshiftbase}{ \zeta }%[1][\ctime]{ \dot{\scalingbase}\parens*{ #1 } } 
\newcommand{\tshift}[1][\ctime]{ \tshiftbase\parens*{ #1 } }

\newcommand{\Div}[1][\point]{ \nabla_{ {#1} } \cdot }
\newmacro{\dt}{ \dd \ctime}

\newmacro{\loss}{\ell}
\NewDocumentCommand{\lossf}{ O{\point_{\horizon}} O{\paramf} }{  \loss\parens*{ #1; #2 }}

\NewDocumentCommand{\lossb}{ O{\point_{0}} O{\paramb} }{  \loss\parens*{ #1; #2 }}

\newmacro{\caching}{M}
\newmacro{\outeriter}{K_\textup{out}}
\newmacro{\inneriter}{K_\textup{in}}

\newmacro{\pretriterf}{K_{\paramf}}
\newmacro{\pretriterb}{K_{\paramb}}

\newmacro{\lrbase}{\gamma}
\newmacro{\lrf}{\lrbase_{\paramf}}
\newmacro{\lrb}{\lrbase_{\paramb}}

\newmacro{\Nconst}{  \parens*{2\pi}^{\frac{\vdim}{2}}  }
\newmacro{\Psta}{ \pbase_{{0\horizon}} }
\newmacro{\Pstasol}{ \pbase^\star_{{0\horizon}} }
\newmacro{\Qsta}{ \refprobase_{{0\horizon} }}

\newmacro{\tPsta}{ \tilde{\pbase}_{\mathrm{0\horizon}} }
\newmacro{\tPstasol}{ \tilde{\pbase}^{\star}_{\mathrm{0\horizon}} }

\newmacro{\tm}{\tilde{\mbase}_0}
\newmacro{\tcovar}{\tilde{\covarbase}_0}

\newmacro{\eqlaw}{  \overset{ \mathrm{law} }{=} }

\newmacro{\Cs}{  C_{\sdev_{\star}} }

\newmacro{\Pt}{ P_{\ctime} }
\newmacro{\Qt}{ Q_{\ctime} }

\newmacro{\St}{S_{\ctime}}
\newmacro{\Kth}{K_{\ctime,\ctime + h}}

\newcommand{\muc}[1][\ctime+h]{ \check{\mu}_{ #1 } }
\newcommand{\Sigmac}[1][\ctime+h]{ \check{\Sigma}_{  #1 } }

\newcommand{\Sigmacinv}[1][\ctime+h]{ \check{\Sigma}^{-1}_{  #1 } }

\DeclareMathOperator{\gradBW}{\mathrm{grad}}

\newmacro{\tSt}{ \tilde{\St} }
\newmacro{\dSt}{\dot{S}_t}

\newmacro{\fpotbase}{\Phi'}

\newmacro{\rfpotbase}{\hat{\Phi'}}

\newmacro{\potbase}{\Phi}
\newcommand{\pot}[1][\ctime]{\potbase_{#1}}

\newmacro{\rpotbase}{\hat{\Phi}}
\newcommand{\rpot}[1][\ctime]{\rpotbase_{#1}}

\newmacro{\measurebase}{ \rho }
\newcommand{\measure}[1][\ctime]{ \measurebase_{#1}  }

\newmacro{\pt}{\partial_{\ctime}}

\newmacro{\potsolbase}{ \Phi }
\newcommand{\potsol}[1][\ctime]{ \potsolbase_{#1} }
\newcommand{\vecsol}[1][\ctime]{ \nabla\potsolbase_{#1} }

\newmacro{\lyapbase}{\mathcal{L}}
\NewDocumentCommand{\lyap}{ O{\Sigmasol} O{\dSigmasol} }{ \lyapbase_{\ssstyle #1}\bracks{#2}  }
\NewDocumentCommand{\lyapinv}{ O{\Sigma} O{2\nabla F} }{ \lyapbase^{-1}_{\ssstyle #1}\bracks{#2}  }

\newmacro{ \ctimebar }{ \bar{\ctime}  }

\renewcommand{\C}{ C_{\sdev} }
\newmacro{\D}{ D_{\sdev} }

\newmacro{\covbase}{\Sigma}
\newcommand{\Sigcurve}[1][\ctime]{  \covbase_{#1}  }

\newcommand{\dSigcurve}[1][\ctime]{  \dot{\covbase}_{#1}  }
\newcommand{\ddSigcurve}[1][\ctime]{  \ddot{\covbase}_{#1}  }

\newcommand{\Sigcurveinv}[1][\ctime]{  \covbase^{-1}_{#1}  }

\newmacro{\kenergyBW}{ \frac{1}{2} \normBWsq[\dSigcurve][\Sigcurve] }

\newmacro{\penergyBWbase}{ \mathcal{U} }
\NewDocumentCommand{\penergyBW}{ O{\sdev} O{\Sigcurve} }{ \penergyBWbase_{#1}\parens*{#2}  }

\newmacro{\entropybase}{ H }
\newcommand{\entropy}[1][\covbase]{ \entropybase\parens*{ #1 } }

\newmacro{\fisherbase}{ {I} }

\newmacro{\fisherBWbase}{ \mathcal{I} }
\newcommand{\fisherBW}[1][\covbase]{ \fisherBWbase_{\sdev}\parens*{#1} }

\newmacro{\LagBW}{ \kenergyBW - \penergyBW }
\newmacro{\HamBW}{ \kenergyBW + \penergyBW }

\newmacro{\tspace}{\mathcal{T}_{\ssstyle \Sigma}\SPD}
\newmacro{\coder}{\nabla_{\ssstyle\dSigcurve}\dSigcurve}
\newmacro{\dderbase}{D}
\newmacro{\dder}{\dderbase_{\ssstyle X}X}
\newmacro{\tdriftbase}{ f }

\NewDocumentCommand{\tdrift}{ O{\ctime} O{\refsde} }{ \tdriftbase\parens*{#1,#2} }

\newmacro{\SB}{ {\scriptscriptstyle\textup{SB}} } %{\scriptscriptstyle\mathrm{SB}} }

\newmacro{\horbase}{M}

\newmacro{\Ht}{ \tilde{S}_{\ctime} }
\newmacro{\dHt}{ \dot{\tilde{S}}_{\ctime} }

\newmacro{\nconst}{ (2\pi)^{\frac{d}{2}} }
\newcommand{\drm}{\mathrm{d}}

\newcommand{\Coupling}{\Pi}

\newcommand{\coupling}{\pi}

\newcommand{\vect}{v_t}
\newcommand{\vectalt}{w_t}
\newcommand{\driftf}{f_t}

\NewDocumentCommand{\innerBW}{ O{\dot{\Sigcurve}} O{\dot{\Sigcurve}} O{\Sigcurve} }{ \inner{#1}{#2}_{\ssstyle #3} }
\NewDocumentCommand{\normBW}{ O{\dSigmasol} O{\Sigmasol} }{ \norm{#1}_{\ssstyle #2} }
\NewDocumentCommand{\normBWsq}{ O{\dSigmasol} O{\Sigmasol} }{ \norm{#1}^2_{\ssstyle #2} }

\newacro{LHS}{left-hand side}
\newacro{RHS}{right-hand side}
\newacro{iid}[i.i.d.]{independent and identically distributed}
\newacro{lsc}[l.s.c.]{lower semi-continuous}

\newacro{GAN}{generative adversarial network}
\newacro{NN}{neural network}
\newacro{FTRL}{``follow the regularized leader''}
\newacro{wp1}[w.p.$1$]{with probability $1$}

\newacro{SDE}{stochastic differential equation}
\newacro{SB}{Schr\"odinger bridge}
\newacro{GSB}[GSB]{Gaussian Schr\"odinger bridge}

\newacro{SGM}{score-based generative model}

\newacro{SMLD}{score matching with Langevin dynamics}
\newacro{DDPM}{denoising diffusion probabilistic model}

\newacro{OU}{Ornstein\textendash Uhlenbeck}
\newacro{BM}{Brownian motion}
\newacro{BDT}{Black–Derman–Toy}

\newacro{VESDE}[VESDE]{variance exploding \ac{SDE}}
\newacro{VPSDE}[VPSDE]{variance preserving \ac{SDE}}

\newacro{ESC}{embryonic stem cell}
\newacro{MEF}{mouse embryonic fibroblast}
\newacro{iPSC}{induced pluripotent stem cell}

\newmacro{\acroalg}{\textsc{GSBflow}}   % acronym for our overall algorithm
\usepackage[accepted]{aistats2023}
\usepackage[round]{natbib}

\begin{document}
\renewcommand\ttdefault{lmtt}
\newcommand*{\tabindent}{\hspace{3mm}}

% If your paper is accepted and the title of your paper is very long,
% the style will print as headings an error message. Use the following
% command to supply a shorter title of your paper so that it can be
% used as headings.
%
%\runningtitle{I use this title instead because the last one was very long}

% If your paper is accepted and the number of authors is large, the
% style will print as headings an error message. Use the following
% command to supply a shorter version of the authors names so that
% they can be used as headings (for example, use only the surnames)
%
%\runningauthor{Surname 1, Surname 2, Surname 3, ...., Surname n}

\twocolumn[

\aistatstitle{The Schr\"odinger Bridge between Gaussian Measures has a Closed Form}

\aistatsauthor{Charlotte Bunne$^*$ \And Ya-Ping Hsieh$^*$ \And  Marco Cuturi \And Andreas Krause}

\aistatsaddress{ETH Z\"urich \And ETH Z\"urich \And Apple$^\ddag$ \And ETH Z\"urich} ]

%----------------------------------------------------------------------
%%% ABSTRACT
%----------------------------------------------------------------------
\begin{abstract}

\newacro{OT}{optimal transport}

The static optimal transport (OT) problem between Gaussians seeks to recover an optimal map, or more generally a coupling, to morph a Gaussian into another. It has been well studied and applied to a wide variety of tasks. 
%, with and without entropic regularization. 
Here we focus on the dynamic formulation of OT, also known as the \emph{Schr\"odinger bridge} (SB) problem, % when regularized with entropy, 
which has recently seen a surge of interest in machine learning due to its connections with diffusion-based generative models. In contrast to the static setting, much less is known about the dynamic setting, even for Gaussian distributions.
%very little is currently known in closed form to the problem, even for Gaussians. 
In this paper, we %fill that gap by 
provide \emph{closed-form expressions for SBs between Gaussian measures}. In contrast to the static Gaussian OT problem, which can be simply reduced to studying convex programs, our framework for solving SBs requires significantly more involved tools such as Riemannian geometry and generator theory. 
%An important consequence of our results is that 
Notably, we establish that the solutions of SBs between Gaussian measures are themselves Gaussian processes with explicit mean and covariance kernels, and thus are readily amenable for many downstream applications such as generative modeling or interpolation.
%To demonstrate the utility, we 
To demonstrate the utility, we devise a new method for modeling the evolution of
single-cell genomics data and report significantly improved numerical stability compared to existing SB-based approaches.

\end{abstract}

%----------------------------------------------------------------------
%%% INTRODUCTION
%----------------------------------------------------------------------
\section{Introduction}
\label{sec:introduction}
%\vspace{-10pt}

\newmacro{\Ninit}{\Ncal_{\ini}}
\newmacro{\Nend}{\Ncal_{\en}}

%\citep{}
\acused{OT}

The \acdef{SB} \citep{leonard2013survey, chen2021stochastic}, alternatively known as the \emph{dynamic} entropy-regularized \acl{OT} (OT), has recently received significant attention from the machine learning community. In contrast to the classical \emph{static} \ac{OT} where one seeks a {coupling} between measures that minimizes the average cost \citep{villani2009optimal, peyre2019computational}, the goal of \acp{SB} is to find the optimal \emph{stochastic processes} that evolve a given measure into another. As such, \acp{SB} are particularly suitable for learning complex continuous-time systems, and have been successfully applied to a wide range of applications such as sampling \citep{bernton2019schr, huang2021schrodinger}, generative modeling \citep{chen2021likelihood,de2021diffusion,pmlr-v139-wang21l}, molecular biology \citep{holdijk2022path}, and mean-field games \citep{liu2022deep}.

\begin{figure}
    \centering
    \includegraphics[width=1.1\linewidth]{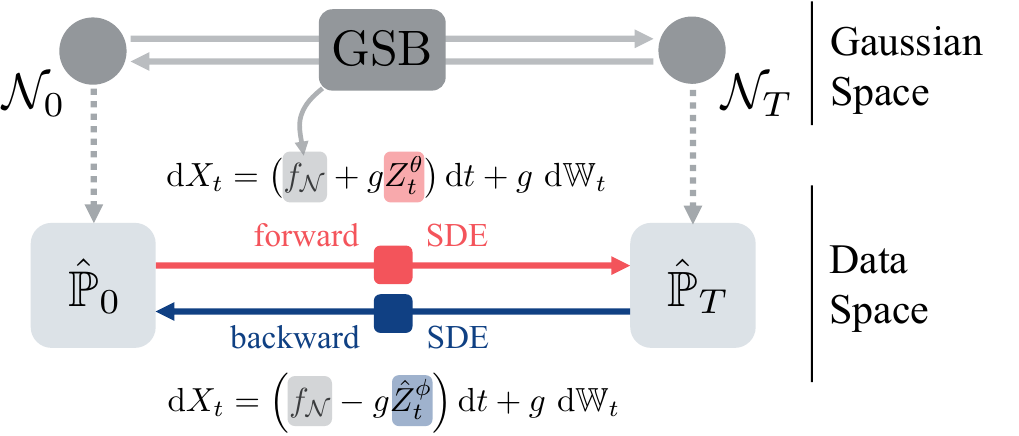}
    \caption{Solving the \ac{SB} problem between $\distinit$ and $\distend$ is notoriously difficult, because it requires learning the time-dependent drifts of two SDEs that respect the desired marginals, and a random initialization for these drifts is usually extremely far from satisfying that constraint. We propose a data-dependent procedure that relies first on Gaussian approximations of the data measures, which provide a closed-form drift $f_{\mathcal{N}}$ in \eqref{eq:GSB-forward} (the \ac{GSB}). We show that this facilitates the training of forward/backward drifts $\hat{Z}^\theta_t, \hat{Z}^\phi_t$.}%
    \label{fig:overview}
\end{figure}

Despite of these impressive achievements, a common limitation of the existing works is that the \acp{SB} are typically solved in a purely {numerical} fashion. In sharp contrast, it is well-known that many important \ac{OT} problems for \emph{Gaussian} measures admit \emph{closed-form} solutions, and the advantages of such solutions are numerous: they have inspired new learning methods \citep{rabin2011wasserstein, vayer2019sliced, bonneel2015sliced}, they can serve as the ground truth for evaluating numerical schemes \citep{janati2020entropic}, and they have lead to the discovery of a new geometry that is both rich in theory and application \citep{takatsu2010wasserstein}. 

\newpage
The goal of our paper is to continue this pursuit of closed-form solutions and thereby extending these advantages to \ac{SB}-based learning methods. For an overview of the method, see Fig.~\ref{fig:overview}. To this end, we make the following \textbf{contributions}: \vspace{8pt}
%As a central contribution, we derive a set of new expressions for \acdefp{GSB}, \ie\acp{SB} between Gaussian measures, that covers a wide range of applications. Furthermore, our analysis reveals a
\begin{enumerate}[leftmargin=.4cm,itemsep=.0cm,topsep=.0cm]
\item As our central result, we derive the closed-form expressions for \acfp{GSB}, \ie\acp{SB} between Gaussian measures. This is a challenging task for which all existing techniques fail, and thus we need to resort to a number of new ideas from entropic \ac{OT}, Riemannian geometry, and generator theory; see \cref{sec:overview}.
%, thereby filling the void of. %, that covers a wide range of applications.  %These formulas constitute the first non-trivial closed forms for \emph{dynamic} entropy-regularized \ac{OT}. %The proof combines a number of ideas from entropic optimal transport, Riemannian geometry, Gaussian analysis, and generator theory, which might be of independent interest; see \crefrange{sec:mechanics}{sec:results}.

\item We extend the deep connection between geometry and Gaussian \ac{OT} to \aclp{GSB}. In particular, our results can be seen as a vast generalization of the classical Bures-Wasserstein geodesics between Gaussian measures \citep{takatsu2010wasserstein, bhatia2019bures}, which is the foundation of many computational methods \citep{chewi2020gradient, altschuler2021averaging, han2021riemannian}.

\item Via a simple Gaussian approximation on real \emph{single-cell genomics} data, we numerically demonstrate that many benefits of the closed-form expressions in static \ac{OT} immediately carry over to \ac{SB}-based learning methods: We report improved numerical stability and tuning insensitivity when trained on benchmark datasets, which ultimately lead to an overall better performance.

\end{enumerate}

%----------------------------------------------------------------------
%%% PRELIMINARIES
%----------------------------------------------------------------------

\section{Preliminaries on Gaussian Optimal Transport Problems}
\label{sec:prelim}
Throughout this paper, let $\xi \sim \Ncal(\mu, \Sigma)$ and $\xi' \sim \Ncal(\mu', \Sigma')$ denote two given Gaussian random variables. By abusing the notation, we will continue to denote the measures of these Gaussians by $\xi$ and $\xi'$, respectively. We will also denote by $\Pi(\xi,\xi')$ the set of all their couplings.

\subsection{Static Gaussian Optimal Transport}
\label{sec:staticGOT}
The \emph{static} entropy-regularized \ac{OT} between Gaussians refers to the following minimization problem \citep{peyre2019computational}:
% An important class of problems which generalizes \eqref{eq:W2Gaussian}, is the static entropy-regularized \ac{OT} between Gaussians:
\begin{equation}
\label{eq:RegW2Gaussian}
%\tag{OT$_\sigma$}
\min_{\coupling \in \Coupling(\xi, \xi')} \int  \norm{\point-\pointalt}^2  \drm \coupling(\point, \pointalt) + 2\sigma^2 \KL{\pi}{ \xi \otimes \xi'  },
\end{equation}where $\xi\otimes\xi'$ denotes the product measure of $\xi$ and $\xi'$, and $\sigma \geq 0$ is a regularization parameter. When $\sigma = 0$, \eqref{eq:RegW2Gaussian} reduces to the classical 2-Wasserstein distance between $\xi$ and $\xi'$ \citep{villani2009optimal}, whose closed-form solution is classical \citep{dowson1982frechet, olkin1982distance}. The case for general $\sigma$ is more involved, and an analytical expression was only recently found \citep{bojilov2016matching, del2020statistical, janati2020entropic, mallasto2021entropy}: Setting% the specific form is quite convoluted, and thus we defer it to \cref{app:GaussianSB}.
\begin{equation}
\label{eq:Cstar}
D_\sdev \defeq \parens{ 4\Sigma^{\frac{1}{2}} \Sigma' \Sigma^{\frac{1}{2}} +  \sdev^4\eye  }^{\frac{1}{2}},\quad C_\sdev \defeq \frac{1}{2}\parens{\Sigma^{\frac{1}{2}} D_\sdev \Sigma^{-\frac{1}{2}} - \sdev^2\eye},
\end{equation}
then the solution $\pi^\star$ to \eqref{eq:RegW2Gaussian} is itself a Gaussian:
\begin{align}
\label{eq:StaticGaussianSB-sol}
\pi^\star \sim \Ncal \parens*{  \begin{bmatrix}
\mu \\
\mu'
\end{bmatrix},  \begin{bmatrix}
\Sigma&C_\sdev\\
C_\sdev^\top&\Sigma'
\end{bmatrix}}.
\end{align}

\subsection{Dynamic Gaussian Optimal Transport}
\label{sec:dynamicGOT-BB}

In the literature, \eqref{eq:RegW2Gaussian} is commonly referred to as the \emph{static} \ac{OT} formulation, since it merely asks \emph{where} the mass should be transported to (\ie $\pi(\point,\pointalt)$ dictates how much mass at $\point$ should be transported to $\pointalt$). In contrast, the more general problem of \emph{dynamic} Gaussian \ac{OT} seeks to answer \emph{how} the mass the should be transported:
\begin{align}
\label{eq:BBGaussian}
%\tag{dyn-OT$_\sigma$}
\min_{\ssstyle\substack{{\measure[0] = \xi,\ \measure[\horizon] = \xi'}}} \exof*{\int_{0}^\horizon  \frac{1}{2} \norm{\vect}^2 + \frac{\sigma^4}{8} \norm*{\nabla\log\measure}^2 \dt}.
\end{align}Here, the minimization is taken over all pairs $(\measure, \vect)$ where $\measure$ is an absolutely continuous curve of measures \citep{ambrosio2006gradient}, and $\vect:\R^\vdim\to\R^\vdim$ is such that the continuity equation holds:
\begin{equation}
\label{eq:Coneq}
\pt\measure = - \Div(\measure \vect),
\end{equation}
where $\left(\Div \vect\right)(\point) \coloneqq \sum_{i=1}^d \frac{\partial}{\partial \point_i}v_t^{i}(\point)$ denotes the divergence operator with respect to the $\point$ variable. It can be shown that, if $\rho^\star_t$ is the optimal curve for \eqref{eq:BBGaussian}, then the joint distribution of the end marginals $(\rho^\star_0, \rho_1^\star)$ coincides with \eqref{eq:StaticGaussianSB-sol}, hence the interpretation of $\rho_t^\star$ as the optimal \emph{trajectory} in the space of measures  \citep{chen2016relation, gentil2017analogy, chen2021stochastic, gentil2020dynamical}.

To our knowledge, the only work that has partially addressed the closed-form solution of \eqref{eq:BBGaussian} is \citet{mallasto2021entropy}, whose results are nonetheless insufficient to cover important applications such as generative modeling. In \cref{sec:results}, we will derive a vast generalization of the results in \citet{mallasto2021entropy} and provide a detailed comparison in \crefrange{sec:overview}{sec:mechanics}.

%----------------------------------------------------------------------
%%% OVERVIEW
%----------------------------------------------------------------------
% Overview
\section{The Gaussian Schr\"odinger Bridge Problem and Analysis Overview}
\label{sec:overview}
\newcommand{\Pstar}[1][\ctime]{\mathbb{P}^\star_{#1}}

The purpose of this section is to introduce the core objectives in our paper, the \aclip{GSB}, and establish their connection to the Gaussian \ac{OT} problems in \cref{sec:prelim}. To help the reader navigate our somewhat technical proofs in \crefrange{sec:mechanics}{sec:results},  we illustrate in \cref{sec:tere} the high-level challenges as well as our new techniques for solving \aclp{GSB}.

\subsection{Schr\"odinger Bridges as Dynamic Entropy-Regularized Optimal Transport}

Let $\nu, \nu'$ be two given measures and let $\refpro$ be an arbitrary stochastic process. In its most generic form, the \acl{SB} refers to the following constrained KL-minimization problem over all stochastic processes $\Pmargin$ \citep{leonard2013survey, chen2021stochastic}: 
\begin{equation}
\label{eq:SB}
%\tag{SB}
\min_{ \substack{ \Pinit = \nu, \; \Pend = \nu'} } \KL{\Pmargin}{\refpro}.% \quad \sdev >0 \textup{ and \Wiener
\end{equation}

In practice, $\nu$ and $\nu'$ typically arise as the (empirical) \emph{marginal} distributions of a complicated continuous-time dynamics observed at the starting and end times, and $\refpro$ is a ``prior process'' representing our belief of the dynamics before observing any data. The solution $\Pstar$ to \eqref{eq:SB} is thus interpreted as the best dynamics that conforms to the prior belief $\refpro$ while respecting the data marginals ($\Pstar[0] = \nu, \Pstar[\horizon] = \nu'$). 

In this paper, we will consider a general class of $\refpro$'s that includes most existing processes in the machine learning applications of \acp{SB}. Specifically, with some initial condition $\refsde[0]$, we will take $\refpro$ to be the measure of the linear \acdef{SDE}:
% To facilitate the numerical solutions of \eqref{eq:SB}, $\refpro$ is typically chosen so that conditional distributions given initial points admit a simple expression. To this end, we will consider the following general class of \acdefp{SDE} which incorporates, to our knowledge, all the existing processes in the literature: With some initial condition $\refsde[0]$, take $\refpro$ to be the measure of
\begin{equation}
\label{eq:linearsde}
\drefsde = \parens*{\drift \refsde  + \shift }\dd t + \volat \dWiener[\ctime] \defeq \driftf\dt + \volat \dWiener[\ctime].
\end{equation}
Here, $\drift: \R^+ \to \R$, $\shift : \R^+ \to \R^\vdim$, and $\volat: \R^+ \to \R^+$ are smooth functions. In this case, \acp{SB} can be seen as generalized dynamical \ac{OT} between two (not necessarily Gaussian) measures:
%it is well-known that \eqref{eq:SB} can be seen as a generalized Benamou-Brenier \ac{OT} for two arbitrary (not necessarily Gaussian) measures \citep{chen2016relation, gentil2017analogy}:
\begin{theorem}
\label{thm:GSBtoBB}
Consider the \acl{SB} problem with $\refsde$ as the reference process: 
\begin{equation}
\label{eq:YtSB}
%\tag{$\refsde$-SB}
\min_{ \substack{ \Pinit = \nu, \; \Pend = \nu'} } \KL{\Pmargin}{\refsde}.% \quad \sdev >0 \textup{ and \Wiener
\end{equation}
Then \eqref{eq:YtSB} is equivalent to
\begin{align}
\label{eq:dynamicGSB}
&\inf_{(\measure,\vect)} \mathbb{E}\Bigg[\int_0^\horizon  \frac{\norm{\vect}^2}{2\volatsq[\ctime]} + \frac{\volatsq[\ctime]}{8} \norm{\nabla \log \measure}^2  - \frac{1}{2} \inner{\driftf}{\nabla \log\measure} \dt\Bigg]
\end{align}where the infimum is taken all pairs $(\measure,\vect)$ such that $\measure[0] = \nu, \measure[\horizon] = \nu'$, $\measure$ absolutely continuous, and
\begin{align}
\label{eq:Coneq2}
\partial_t \measure = -\Div \parens*{  \measure\parens*{ \driftf + \vect}  }.
\end{align}
\end{theorem}
The proof of \cref{thm:GSBtoBB}, which we defer to \cref{app:GSBtoBB}, is a straightforward extension of the argument in \citep{leonard2013survey, chen2016relation, gentil2017analogy} which establishes the equivalence when $\refsde$ is a reversible \acl{BM}, \ie $\driftf \equiv0, \volat[\ctime] \equiv \sigma,$ and $\refsde[0]$ follows the Lebesgue measure.\footnote{The reversible \acl{BM} is a technical construct to simplify the computations. For our purpose, one can think of $\refsde[0]\sim \xi$ instead of the Lebesgue measure, and our results still hold verbatim.}

\subsection{The Gaussian Schr\"odinger Bridge Problem}
\label{sec:tere}

The central goal of our paper is to derive the closed-form solution of \acp{SB} when the marginal constraints are Gaussians $\xi \sim \Ncal(\mu, \Sigma),\ \xi' \sim \Ncal(\mu', \Sigma')$. Namely, we are interested in the following class of the \acp{SB}, termed \aclp{GSB}:% refers to the KL-minimization problem:
\begin{equation}
\label{eq:GSB}
\tag{GSB}
\min_{ \substack{ \Pinit = \xi, \; \Pend = \xi'} } \KL{\Pmargin}{\refsde}.% \quad \sdev >0 \textup{ and \Wiener
\end{equation}
To emphasize the dependence on the reference \ac{SDE}, we will sometimes call \eqref{eq:GSB} the $\refsde$-\ac{GSB}.

%In view of \cref{thm:GSBtoBB}, $\refsde$-\acp{GSB} are 
%When $\refsde = \variance \Wiener$, 
% $\mathbb{P}^\star_{0,1} = \pi^\star$ 
%
%General reference processes

%\subsection{Technical Challenges and Related Work} 
%\label{sec:tere}

\textbf{Technical challenges; related work.} In order to analyze \eqref{eq:GSB}, we first notice that the objective in \eqref{eq:dynamicGSB} becomes $\sigma^{-2}\exof*{\int_{0}^\horizon  \frac{1}{2} \norm{\vect}^2 + \frac{\sigma^4}{8} \norm*{\nabla\log\measure}^2 \dt}$ for $\sigma\Wiener$-\acp{GSB}. Up to a constant factor, this is simply \eqref{eq:BBGaussian}, so \cref{thm:GSBtoBB} reduces to the well-known fact that $\sigma\Wiener$-\acp{GSB} are a reformulation of the dynamic Gaussian \ac{OT} \citep{leonard2013survey, chen2016relation, gentil2017analogy}. 

At first sight, this might suggest that one can extend existing tools in Gaussian \ac{OT} to analyze \acp{GSB}.%\eqref{eq:BBGaussian} to \eqref{eq:dynamicGSB}.
~Unfortunately, the major difficulty of tackling \acp{GSB} is that these existing tools are fundamentally insufficient for the generalized objective \eqref{eq:dynamicGSB}. 
~To be more precise, there exist three prominent frameworks for studying Gaussian \ac{OT} problems:% \eqref{eq:BBGaussian}:
\begin{itemize}[leftmargin=.4cm,itemsep=.0cm,topsep=.0cm]
\item \textbf{Convex analysis:} An extremely fruitful observation in the field is that many Gaussian \ac{OT} instances can be reduced to a \emph{convex} program, for which one can import various convex techniques such as KKT or fixed-point arguments. This is the case for static Gaussian \ac{OT} \eqref{eq:RegW2Gaussian}, both when $\sigma = 0$ \citep{dowson1982frechet, olkin1982distance, bhatia2019bures} and $\sigma>0$ \citep{janati2020entropic}. Furthermore, in the case of $\sigma=0$, the solution to the dynamic formulation \eqref{eq:BBGaussian} can be recovered from the static one via a simple linear interpolation \citep{mccann1997convexity}.

\item \textbf{Ad hoc computations:} When $\sigma>0$ in \eqref{eq:BBGaussian}, the problem is no longer reducible to a convex program \citep{leonard2013survey, chen2021stochastic}. In this case, the only technique we are aware of is the ad hoc approach of \citep{mallasto2021entropy}, which manages to find a closed form for \eqref{eq:BBGaussian} (and thus $\sigma\Wiener$-\acp{GSB}) through a series of brute-force computations.

\item \textbf{Control theory:} On a related note, in a series of papers, \citet{chen2015optimal,chen2016relation,chen2018optimal} exploit the deep connection between $\sigma\Wiener$-\acp{GSB} and control theory to study the \emph{existence} and \emph{uniqueness} of the solutions. Although a variety of new optimality conditions are derived in these works, they are all expressed in terms differential equations with coupled initial conditions, and it is unclear whether solving these differential equations is an easier task than \eqref{eq:GSB} itself. In particular, no closed-form, even for $\sigma\Wiener$-\acp{GSB}, can be found therein.
\end{itemize}

By \cref{thm:GSBtoBB}, \acp{GSB} are more general than \eqref{eq:BBGaussian} and thus irreducible to convex programs, so there is no hope for the convex route. As for ad hoc computations, the time-dependent $\driftf$ and $\volat[\ctime]$ terms in \eqref{eq:dynamicGSB} present a serious obstruction for generalizing the approach of \citet{mallasto2021entropy} to $\refsde$-\acp{GSB} when $\driftf \neq 0$ or $\volat[\ctime]$ is not constant; this is exemplified by the convoluted expressions in our \cref{thm:GaussianSB}, which hopefully will convince the reader that they are beyond any ad hoc guess. Finally, the control-theoretic view has so far fallen short of producing closed-form solutions even for $\sigma\Wiener$-\acp{GSB}, so it is essentially irrelevant for our purpose. 

To conclude, in order to find an analytic expression for general \acp{GSB}, we will need drastically different techniques.

\para{Our approach}

To overcome the aforementioned challenges, in \cref{sec:mechanics}, we will first develop a principled framework for analyzing the closed-form expressions of $\sigma\Wiener$-\acp{GSB}, \ie \eqref{eq:BBGaussian}. Unlike the ad hoc approach of \citet{mallasto2021entropy} which is very specific to \aclp{BM}, our analysis reveals the general role played by the \emph{Lyapunov operator} (see \eqref{eq:lya}) on covariance matrices, thereby essentially reducing the solutions of \acp{GSB} to solving a matrix equation. This route is enabled via yet another equivalent formulation of \eqref{eq:BBGaussian}, namely the action minimization problem on the \emph{Bures-Wasserstein geometry}, which has recently emerged as a rich source for inspiring new computational methods \citep{chewi2020gradient, altschuler2021averaging, han2021riemannian}. In \cref{sec:results}, we show how the insight gained from our geometric framework in \cref{sec:mechanics} can be easily adapted to \acp{GSB} with general reference processes, which ultimately leads to the full resolution of \eqref{eq:GSB}.

%----------------------------------------------------------------------
%%% MECHANICS
%----------------------------------------------------------------------
\section{The Bures-Wasserstein Geometry of $\sigma\Wiener$-Gaussian Schr\"odinger Bridges}
\label{sec:mechanics}

This section illustrates the simple geometric intuition that underlies the somewhat technical proof of our main result (\cf \cref{thm:GaussianSB}). After briefly reviewing the action minimization problems on Euclidean spaces in \cref{sec:actionRd}, we present the main observation in \cref{sec:actionBW}: $\sigma\Wiener$-\acp{GSB} are but action minimization problems on the Bures-Wasserstein manifolds, which can be tackled by following a standard routine in physics.

\subsection{A Brief Review on Action Minimization Problems}% in Euclidean and Riemannian Settings}
%\vspace{-8pt}
\label{sec:actionRd}

Consider the following \emph{action minimization} problem with fixed endpoints $\point, \point' \in \R^\vdim$:
\begin{align}
\label{eq:Euclid_Lag}
\min_{x(0) = \point, x(\horizon) = \point'} \int_0^\horizon {\frac{1}{2} \norm{\dot{x}(t)}^2 - U(x(t)) } \dt,
\end{align}
where the minimum is taken over all piecewise smooth curves. 
%It is common to interpret the term $\frac{1}{2}\norm{\dot{x}(t)}^2$ as the \emph{kinetic} energy of $x(t)$, and $U(x(t))$ the \emph{potential} energy. 
~A celebrated result in physics asserts that the optimal curve for \eqref{eq:Euclid_Lag} satisfies the \emph{Euler-Lagrange} equation:
\begin{equation}
\label{eq:EL}
\ddot{x}(t) = -\nabla U(x(t)), \quad x(0) = x,\quad x(\horizon) = x'.
\end{equation}
In particular, when $U \equiv0$, \eqref{eq:EL} reduces to $\ddot{x} \equiv 0$, \ie $x(t)$ is a straight line connecting $\point$ and $\point'$.

More generally, one can consider \eqref{eq:Euclid_Lag} on any \emph{Riemannian manifold}, provided that the Euclidean norm $\norm{\cdot}$ in \eqref{eq:Euclid_Lag} is replaced by the corresponding Riemannian norm. In this case, the Euler-Lagrange equation \eqref{eq:EL} still holds, with $\ddot{x}$ and $\nabla U$ replaced with their Riemannian counterparts \citep{villani2009optimal}.

\subsection{$\sigma\Wiener$-\acp{GSB} as Action Minimization Problems}

\label{sec:actionBW}

% The Benamou-Brenier \ac{OT} problem \eqref{eq:BBGaussian} can be posed for any pair of measures and is thus not restricted to Gaussians. However, since Gaussians are uniquely determined by their means and covariances, specializing the Benamou-Brenier \ac{OT} to Gaussians 

We begin with the following simple observation. Based on the seminal work by \citet{otto2001geometry}, \citet{gentil2020dynamical} show that \acp{SB} between two arbitrary measures can be formally understood as an action minimization problem of the form \eqref{eq:Euclid_Lag} on an \emph{infinite}-dimensional manifold. Since we have restricted the measures in \eqref{eq:GSB} to be Gaussian, and since Gaussian measures are uniquely determined by their means and covariances, \citet{gentil2020dynamical} strongly suggests a \emph{finite}-dimensional geometric interpretation of $\sigma\Wiener$-\acp{GSB}. The main result in this section, \cref{thm:mechanics} below, makes this link precise.

The proper geometry we need is the \emph{Bures-Wasserstein manifold} \citep{takatsu2010wasserstein, bhatia2019bures} defined as follows. Consider the space of covariance matrices (\ie symmetric positive definite matrices) of dimension $\vdim$, which we denote by $\SPD$, and consider its natural {tangent space} as the space of symmetric matrices:
\begin{equation}
\mathcal{T}_{\ssstyle \Sigma} \SPD \defeq \setdef{U \in \R^{\vdim\times\vdim}}{ U^\top = U}.
\end{equation}
A notion that will play a pivotal role is the so-called \emph{Lyapunov operator}: For any $\Sigma \in \SPD$ and $U\in\mathcal{T}_{\ssstyle \Sigma}\SPD$, we define $\lyap[\Sigma][U]$ to be the symmetric solution to the equation
\begin{align}
\label{eq:lya}
%\tag{Lya}
A: \quad \Sigma A  + A \Sigma = U.
\end{align}
It is shown in \citet{takatsu2010wasserstein} that the Lyapunov operator defines a geometry on $\SPD$, known as the \emph{Bures-Wasserstein geometry}: For any two tangent vectors $U, V\in \tspace$, the operation
\begin{align}
\label{eq:BW-tensorMain}
\innerBW[U][V][\Sigma] \defeq %\tr\lyap[\Sigma][U]\Sigma\lyap[\Sigma][V] = 
\frac{1}{2}\tr \lyap[\Sigma][U]V%,  \quad \normBWsq[U][\Sigma] \coloneqq \innerBW[U][U][\Sigma]
\end{align}satisfies all the axioms of the Riemannian metric; additional background on the Bures-Wasserstein geometry can be found in \cref{app:reviewBW}. 

We are now ready to state the main result of the section. Let $\normBW[\cdot][\Sigma]$ be the induced norm of $\innerBW[\cdot][\cdot][\Sigma]$. Fix $\sdev >0$ and let $\Wiener$ be a {reversible} \acl{BM}. Consider the following special case of \eqref{eq:GSB}:% where $\refsde \subs \sdev \Wiener$:
\begin{equation}
\label{eq:GaussianSBWiener}
%\tag{$\Wiener$-SB$_{\ssstyle\Ncal}$}
\min_{ \Pinit = \Ncal(0, \Sigma),\; \Pend = \Ncal(0, \Sigma') } \KL{\Pmargin}{\sdev\Wiener}.
\end{equation}
%For the purpose of illustration, we assume that $\mu = \mu' = 0$.% the mean of both initial and final distributions to be zero.
Then we have:
\begin{restatable}{theorem}{mechanics}
\label{thm:mechanics}
The minimizer of \eqref{eq:GaussianSBWiener} (and hence \eqref{eq:BBGaussian}) coincides with the solution of the action minimization problem:
\begin{align}
\label{eq:LagBW}
%\tag{L}
\min_{\Sigcurve[0] = \Sigma, \Sigma_\horizon = \Sigma'} \int_{0}^\horizon \LagBW \dt
\end{align}where $\penergyBW \defeq - \frac{\sdev^4}{8} \tr \Sigcurveinv$ and the minimum is taken over all piecewise smooth curves in $\SPD$. In particular, the minimizer of \eqref{eq:GaussianSBWiener} solves the Euler-Lagrange equation in the Bures-Wasserstein geometry:
\begin{equation}
\label{eq:EL-BW}
\nabla_{\ssstyle\dSigcurve}\dSigcurve = -  \gradBW \penergyBW, \quad \Sigcurve[0] = \Sigma,\quad \Sigcurve[\horizon] = \Sigma',
\end{equation}where $\nabla_{\ssstyle\dSigcurve}\dSigcurve$ denotes the Riemannian acceleration and $\gradBW$ the Riemannian gradient in the Bures-Wasserstein sense.
\end{restatable}
%\vspace{-8pt}

\para{An important implication}As alluded to in \cref{sec:overview}, the solution curve to \eqref{eq:BBGaussian} or \eqref{eq:GaussianSBWiener} is not new; it is derived in \citet{mallasto2021entropy} via a strenuous and rather unenlightening calculation:
\begin{equation}
\label{eq:solMain}
\Sigmasol \defeq \ctimebar^2 \Sigma + \ctime^2 \Sigma' + \ctime\cdot\ctimebar\parens*{\C+\C^\top+ \sdev^2 \eye}.
\end{equation}Here, $\ctimebar \defeq 1-\ctime$ and $\C$ is defined in \eqref{eq:Cstar}. However, the interpretation of \eqref{eq:solMain} as the minimizer of \eqref{eq:LagBW} is new and suggests a principled avenue towards the closed-form solution of $\sdev\Wiener$-\acp{GSB}: solve the Euler-Lagrange equation \eqref{eq:EL-BW}. Inspecting the formulas for $\nabla_{\ssstyle\dSigcurve}\dSigcurve$ and $\gradBW \penergyBW$ (see \eqref{eq:gradBW-lyapinv} and \eqref{eq:coderBW}), one can further reduce \eqref{eq:EL-BW} to computing the Lyapunov operator $\lyap$, which presents the bottleneck in the proof of \cref{thm:mechanics} as there is, in general, no closed form for the matrix equation \eqref{eq:lya}. To this end, our main contribution is the following technical Lemma:
\begin{lemma}\label{lem:lyapMain}
Define the matrix $\tilde{\St}$ to be:
\begin{equation}
\label{eq:HtMain}
\Ht \defeq \ctime\Sigma' + \ctimebar \C - \ctimebar\Sigma - \ctime\C^\top + \frac{\sdev^2}{2}(\ctimebar-\ctime)\eye.
\end{equation}Then $\Ht^\top\Sigmasolinv$ is symmetric.
\end{lemma}
Armed with \cref{lem:lyapMain}, it is straightforward to verify that $\lyap = \Ht^\top\Sigmasolinv$, \ie $\Ht^\top \Sigmasolinv$ is symmetric and satisfies:
\begin{align}
   \Ht^\top\Sigmasolinv \cdot \Sigmasolinv + \Sigmasolinv \cdot\Sigmasolinv \Ht &= \Ht^\top +  \Ht= \dSigmasol
\end{align}which is more or less equivalent to the original Euler-Lagrange equation \eqref{eq:EL-BW}; we defer the details to \cref{app:proofmechanics}.

To conclude, in contrast to the purely technical approach of \citet{mallasto2021entropy}, our \cref{thm:mechanics} provides a geometric and conceptually clean solution for $\sdev\Wiener$-\acp{GSB}: Compute the Lyapunov operator $\lyap$ via verifying the symmetry of the matrix in \cref{lem:lyapMain}. \emph{It turns out that this technique can be readily extended to general \acp{GSB}}, and therefore serves as the foundation for the proof of our main result; see \cref{sec:results}.

\para{Remark}
It is interesting to note that the matrix $\Ht$ in \eqref{eq:HtMain} is itself \emph{not} symmetric. Other consequences of \cref{thm:mechanics} that might be of independent interest can be found in \cref{sec:interesting}. We also note that, when $\sdev = 0$, the solution to \eqref{eq:LagBW} is simply the Wasserstein geodesic between Gaussian measures, whose formula is well-known \citep{dowson1982frechet, takatsu2010wasserstein}. However, as explained in \cref{sec:overview}, the case of $\sdev > 0$ requires a completely different analysis since, unlike when $\sdev = 0$, it is not reducible to a convex program. This leads to the significantly more involved proofs of \cref{thm:mechanics} and of \eqref{eq:solMain} in \citet{mallasto2021entropy}.

%----------------------------------------------------------------------
%%% GAUSSIAN SBs
%----------------------------------------------------------------------
%\vspace{-8pt}
\section{Closed-Form Solutions of General Gaussian Schr\"odinger Bridges}
%\vspace{-8pt}
\label{sec:results}
We now present the closed-form solutions of general \acp{GSB}.

%----------------------------------------------------------------------
%%% LINEAR SDES
%----------------------------------------------------------------------
\subsection{Linear Stochastic Differential Equations}
\label{sec:linearsdes}

We need the following background knowledge on the linear \ac{SDE} $\refsde$. Let $\aggtime \defeq \exp\parens*{  \int_0^\ctime \drift[\ctimealt] \dd \ctimealt}$. Then the solution to \eqref{eq:linearsde} is \citep{platen2010numerical}:
\begin{align}
\label{eq:linearsdesol}
\refsde = \aggtime  \parens*{\refsde[0] +  \int_0^\ctime  \aggtimeinv[\ctimealt]\shift[\ctimealt]  \dd \ctimealt + \int_0^\ctime {\aggtimeinv[\ctimealt]}{ \volat[\ctimealt] } \dWiener }.
\end{align}
%where $\aggtime \defeq \exp\parens*{  \int_0^\ctime \drift[\ctimealt] \dd \ctimealt}$.

Another crucial fact in our analysis is that $\refsde$ is a \emph{Gaussian process given $\refsde[0]$}, and is thus characterized by the first two moments. Using the independent increments of $\Wiener$ and It\^o's isometry \citep{protter2005stochastic}, we compute:
\begin{align}
\label{eq:mYdef}
\exof*{  \refsde \given \refsde[0]  } &= \aggtime  \parens*{\refsde[0] +  \int_0^\ctime {\aggtimeinv[\ctimealt]}{ \shift[\ctimealt] } \dd \ctimealt  } \eqdef \mYcinit  
\end{align}and, for any $\ctime' \geq \ctime$, 
\begin{align}
\label{eq:covYdef}
&\exof*{  \parens*{ \refsde - \mYcinit  } \parens*{ \refsde[\ctime']- \mYcinit[\ctime']   }^\top \given \refsde[0] } \\
&\hspace{20mm}= \parens*{\aggtime \aggtime[\ctime'] \intdasq} \eye \eqdef \kernel \eye.
\nn
\end{align}

%----------------------------------------------------------------------
%%% CLOSED-FORM OF GAUSSIAN SB
%----------------------------------------------------------------------
\subsection{Main Result}%\vspace{-8pt}
\label{sec:GaussianSB-main}

{\setlength\doublerulesep{0.4pt}
\begin{table*}[!ht]
\vspace{-7mm}
\vskip 0.15in
%\begin{center}
\begin{small}
\begin{sc}
    \centering
    \addtolength{\leftskip} {-2cm}
    \addtolength{\rightskip}{-2cm}
\begin{adjustbox}{width=\textwidth}
\begin{tabular}{ccccccccccr}
\toprule[1.2pt]\midrule[0.2pt]
\makecell{\ac{SDE} with\\$\shift\equiv 0$} & Setting  & $\kernel$ & $\sigma^2_\star$  & $\ratio$ & $\ratioc$ & $\efftr$ & $\tshift$ \\
\midrule
\acs{BM}    & \makecell{$\scriptstyle\drift  \equiv 0$\\ $\scriptstyle\volat \equiv \vconst \in \R^+$}  & $\vconst^2\ctime$ & $\vconst^2$ & ${\ctime}$ & $1- {\ctime}$ & ${\ctime}$ & $0$  \\
\acs{VESDE} & \makecell{$\scriptstyle\drift  \equiv 0$\\ $\scriptstyle\volat = \sqrt{\dqv}$}  & $\qv$ & $\qv[\horizon]$ & $\frac{\qv}{\qve}$ & $1- \frac{\qv}{\qve}$ & $\frac{\qv}{\qve}$ & $0$  \\
\acs{VPSDE}    & \makecell{$\scriptstyle-2\drift = \volatsq[\ctime]$}  & $\scriptstyle\aggtime[\ctime']\parens*{\aggtimeinv[\ctime]-\aggtime} $ & $\scriptstyle\aggtimeinv[\horizon]-\aggtime[\horizon]$ & $\scriptstyle\frac{\aggtimeinv[\ctime]-\aggtime }{\aggtimeinv[\horizon]-\aggtime[\horizon]}$ & $\scriptstyle \aggtime[\horizon]\parens*{ \frac{\aggtime}{\aggtime[\horizon]} - \frac{\aggtimeinv[\ctime]-\aggtime }{\aggtimeinv[\horizon]-\aggtime[\horizon]} }$ & $\scriptstyle\frac{\aggtimeinv[\ctime]\parens*{\aggtimeinv[\ctime]-\aggtime }}{\aggtimeinv[\horizon]\parens*{\aggtimeinv[\horizon]-\aggtime[\horizon]}}$ & $0$  \\
sub\textendash\acs{VPSDE}    & \makecell{$\scriptstyle\frac{\volatsq[\ctime]}{-2\drift} = { 1- \aggtimeqr} $}  & $\scriptstyle \aggtime\aggtime[\ctime']\parens*{ \aggtimeinv[\ctime]-\aggtime}^2  $ & $\scriptstyle\aggtime[\horizon]\parens*{ \aggtimeinv[\horizon]-\aggtime[\horizon]}^2$ & $\scriptstyle \frac{\aggtime}{\aggtime[\horizon]}\cdot\parens*{\frac{ \aggtimeinv[\ctime]-\aggtime}{ \aggtimeinv[\horizon]-\aggtime[\horizon]}}^2$ & $\scriptstyle\aggtime\parens*{1- \parens*{\frac{ \aggtimeinv[\ctime]-\aggtime}{ \aggtimeinv[\horizon]-\aggtime[\horizon]}}^2}$ & $\scriptstyle\parens*{\frac{\aggtimeinv[\ctime]-\aggtime}{\aggtimeinv[\horizon] - \aggtime[\horizon]}}^2$ & $0$  \\
\midrule
\midrule
\makecell{\ac{SDE} with\\$\shift\not\equiv 0$} & Setting  & $\kernel$ & $\sigma^2_\star$  & $\ratio$ & $\ratioc$ & $\efftr$ & $\tshift$ \\
\midrule
\acs{OU}/Vasicek   & \makecell{$\scriptstyle\drift \equiv -\dconst\in\R$\\$\scriptstyle\shift \equiv \sconst\in\R^{\vdim}$\\$\scriptstyle\volat \equiv \vconst \in \R^+$}  & $\scriptstyle\frac{\vconst^2e^{-\dconst \ctime'} \sinh \dconst\ctime}{\dconst} $ & $\scriptstyle\frac{\vconst^2\sinh\dconst}{\dconst}$ & $\scriptstyle\frac{\sinh\dconst\ctime}{\sinh\dconst}$ & \makecell{$\scriptstyle\sinh\dconst\ctime\coth\dconst\ctime$\\$-\scriptstyle\sinh\dconst\ctime\coth\dconst$} & \makecell{$\scriptstyle e^{-\dconst(\horizon-\ctime)}$\\$\scriptstyle\cdot\frac{\sinh\dconst\ctime}{\sinh\dconst}$} & $\scriptstyle\frac{\sconst}{\dconst}\parens*{1-e^{-\dconst\ctime}} $  \\
$\shift$-\acs{BDT}   & \makecell{$\scriptstyle\drift \equiv 0$\\$\scriptstyle\volat \equiv \vconst \in \R^+$}  & $\vconst^2\ctime$ & $\vconst^2\horizon$ & ${\ctime}$ & $1- {\ctime}$ & ${\ctime}$ & $\scriptstyle\int_0^\ctime \shift[\ctimealt]\dd \ctimealt$  \\
\midrule[0.3pt]\bottomrule[1.2pt]
\end{tabular}
\end{adjustbox}
\caption{Examples of reference \acp{SDE} and the corresponding solutions of \acp{GSB}. All relevant functions in the Table are either introduced in \cref{sec:linearsdes} or \eqref{eq:functions}.}
\label[table]{tab:examples}
\end{sc}
\end{small}
%\end{center}
\vskip -0.1in
\end{table*}
}

We now present the main result of our paper. With the important application of diffusion-based models in mind, we will not only derive solution curves as in \eqref{eq:solMain} but also their \ac{SDE} representations.

Let $\xi= \Ncal(\m,\covar)$ and $\xi' = \Ncal(\malt,\covaralt)$ be two arbitrary Gaussian distributions in \eqref{eq:GSB}, and let $\D, \Cstar$ be as defined in \eqref{eq:Cstar}.
%Set $D_\sdev \defeq \parens{ 4\covar^{\frac{1}{2}} \covaralt \covar^{\frac{1}{2}} +  \sdev^4\eye  }^{\frac{1}{2}}$ and $C_\sdev \defeq \frac{1}{2}\parens{\covar^{\frac{1}{2}} D_\sdev \covar^{-\frac{1}{2}} - \sdev^2\eye}$.
%We will adopt the following notation from \citep{janati2020entropic}:
%\begin{align}
%\nn
%D_\sdev \defeq \parens*{ 4\covar^{\frac{1}{2}} \covaralt \covar^{\frac{1}{2}} +  \sdev^4\eye  }^{\frac{1}{2}}, \quad
%C_\sdev \defeq \frac{1}{2}\parens*{\covar^{\frac{1}{2}} D_\sdev \covar^{-\frac{1}{2}} - \sdev^2\eye},
%\end{align}
%where $\sdev>0$ is to be determined later.
%Let $ \ratio \defeq {\frac{t}{T}}, \ratioc \defeq 1-\ratio$, and define $P_\ctime \defeq \ratio \covaralt + \ratioc C$, $Q_t \defeq \ratioc\covar + \ratio C$. Let $\xi_t \sim \Ncal\parens{\mu_t,\Sigma_t}$ where
%\begin{align}
%\label{eq:Psolmean}
%\mu_t &= \ratioc\m + \ratio\malt, \\
%\label{eq:Psolcov}
%\Sigma_t &= \ratioc^2 \covar + \ratio^2 \covaralt + \ratio\ratioc\parens*{C + C^\top + \horizon\eye }.
%\end{align}
%Our main result in this section is:
\begin{restatable}{theorem}{GaussianSB}
\label{thm:GaussianSB}
Denote by $\Psol$ the solution to \aclp{GSB} \eqref{eq:GSB}. Set 
\begin{align}
\nn
&\ratio \defeq \frac{\kernel[\ctime][\horizon]  }{ \kernel[\horizon][\horizon]},\quad \ratioc \defeq \aggtime - \ratio \aggtime[\horizon],\quad \sdev_\star \defeq \sqrt{ \aggtimeinv[\horizon]\kernel[\horizon][\horizon]}, \\ 
\nn
&\tshift \defeq \aggtime\int_0^\ctime \aggtimeinv\shift[\ctimealt]\dd \ctimealt, \efftr \defeq \frac{\intdasq}{\intdasqT}, \\
\nn
&P_\ctime \defeq \dratio\parens*{\ratio \covaralt + \ratioc C_{\sdev_\star}}, \quad Q_\ctime \defeq -\dratioc\parens*{\ratioc\covar + \ratio C_{\sdev_\star}}, 
%\label{eq:PtQtSt}
\\ 
&S_\ctime \defeq P_\ctime -  Q_\ctime^\top + \bracks*{ \drift \kernel[\ctime][\ctime] \parens*{ 1- \efftr }  -  \volatsq[\ctime] \efftr  }\eye.
\label{eq:functions}
\end{align} 

Then the following holds:
\begin{enumerate}[leftmargin=.5cm,itemsep=.01cm,topsep=0cm]
\item The solution $\Psol$ is a Markov Gaussian process whose marginal variable $\Xsol \sim\Ncal\parens*{\meansol,\Sigmasol}$, where
\begin{align}
\label{eq:meansol}
\meansol &\defeq \ratioc \m + \ratio \malt + \tshift - \ratio \tshift[\horizon], \\
\label{eq:covsol}
\Sigmasol&\defeq \ratioc^2\covar + \ratio^2 \covaralt + \ratio\ratioc \parens*{ C_{\sdev_\star} + C_{\sdev_\star}^\top } +  \kernel[\ctime][\ctime]  \parens*{ 1- \efftr } \eye.
\end{align}%Then the marginal variable $\Xsol \sim \Psol$ follows $\Ncal\parens*{\meansol,\Sigmasol}$.
\item $\Xsol$ admits a closed-form representation as the \ac{SDE}:
\begin{align}
\label{eq:GSB-forward-sol}
\dXsol = \GSBf[\ctime][\Xsol] \dd \ctime + \volat\dWiener[\ctime]
\end{align}where 
\begin{align}
\label{eq:GSB-forward}
\GSBf &\defeq \St^\top\Sigmasolinv\parens*{\point - \meansol} + \dmeansol.
\end{align}Moreover, the matrix $S_\ctime^\top\Sigmasolinv$ is symmetric.
\end{enumerate}
\end{restatable}

As in \cref{thm:mechanics}, the key step in the proof of \cref{thm:GaussianSB} is to recognize the symmetry of the matrix $\St^\top \Sigmasolinv$ where $\St$, defined in \eqref{eq:functions}, simply becomes the $\Ht$ in \cref{lem:lyapMain} (up to an additive factor of $\frac{\sdev^2\bar{t}}{2}\eye$) for $\sdev\Wiener$-\acp{GSB}. Although this can be directly verified via generalizing \cref{lem:lyapMain}, the computation becomes quite tedious, so our proof of \cref{thm:GaussianSB} will follow a slightly different route. In any case, given the symmetry of $\St^\top \Sigmasolinv$, the proof simply boils down to a series of straightforward calculations; see \cref{app:GaussianSB}.

\para{Closed forms for conditional distributions}
In many practical applications such as generative modeling, a requirement to employ the \ac{SDE} representation of \acp{GSB} in \eqref{eq:GSB-forward-sol} is that its \emph{conditional distributions} given the initial points can be computed efficiently. As an immediate corollary of \cref{thm:GaussianSB}, we obtain the following closed-form expressions for these conditional distributions. % Importantly, this will allow us to perform a \emph{pretraining} procedure, described in detail in \cref{sec:methods}, for \emph{both} the forward and the backward drifts in the likelihood objectives \eqref{eq:SB-sde-forward}-\eqref{eq:SB-sde-backward}.%, which will prove extremely valuable in practice.
\begin{restatable}{corollary}{GSBMargins}
\label{cor:GSB-Margins}
Let $\Xsol \sim \Psol$ be the the solution to \eqref{eq:GSB}. Then the conditional distribution of $\Xsol$ given end points has a simple solution: $\Xsol \vert \Xsol[0] = x_0 \sim \Ncal\parens*{  \cmeansol{\ctime}{0}, \cSigmasol{\ctime}{0} }$, where
\begin{align}
%\nn
\cmeansol{\ctime}{0} 
&= \ratioc x_0 + \ratio \parens*{   \malt + C_{\sdev_\star}^\top \covar^{-1}(x_0 - \m) } + \tshift - \ratio \tshift[\horizon], %\\
\label{eq:cmean-marginal-0}
%&= \ratioc x_0 + \ratio \cmeansol{\horizon}{0} + \tshift - \ratio \tshift[\horizon], \\
\\
%\nn
\cSigmasol{\ctime}{0} &= \ratio^2 \parens*{ \covaralt - \Cstar^\top \covar^{-1}\Cstar } + \kernel[\ctime][\ctime]  \parens*{ 1- \efftr } \eye. %\\
\label{eq:csigma-marginal-0}
%&= \ratio^2 \cSigmasol{\horizon}{0} +  \kernel[\ctime][\ctime]  \parens*{ 1- \efftr } \eye.
\end{align}
Similarly, $\Xsol \vert \Xsol[\horizon] = x_{\horizon} \sim \Ncal\parens*{  \cmeansol{\ctime}{\horizon}, \cSigmasol{\ctime}{\horizon} }$, where
\begin{align}
%\nn
\cmeansol{\ctime}{\horizon} 
&= \ratio x_{\horizon} + \ratioc \parens*{   \m + C_{\sdev_\star} \covaralt^{-1}(x_{\horizon} - \malt) } + \tshift - \ratio \tshift[\horizon], %\\
\label{eq:cmean-marginal-T} \\
%&= \ratio x_{\horizon} + \ratioc \cmeansol{0}{\horizon} + \tshift - \ratio \tshift[\horizon], \\
%\nn
\cSigmasol{\ctime}{\horizon} &= \ratioc^2 \parens*{ \covar - \Cstar \covaralt^{-1}\Cstar^{\top} } + \kernel[\ctime][\ctime]  \parens*{ 1- \efftr } \eye. %\\
\label{eq:csigma-marginal-T}
%&= \ratioc^2 \cSigmasol{0}{\horizon} +  \kernel[\ctime][\ctime]  \parens*{ 1- \efftr } \eye.
\end{align}
\end{restatable}
%The proof is a straightforward combination of \eqref{eq:Xsol}, \cref{lem:Cond-Gaussians}, and \eqref{eq:meansol}\textendash\eqref{eq:covsol}.

\textbf{Examples of \acp{GSB}.}
Our framework captures most popular reference \acp{SDE} in the machine learning literature as well as other mathematical models in financial engineering; see \cref{tab:examples}.%, where one can simply plug in the corresponding entry into \eqref{eq:GSB-forward} to yield the desirable \ac{GSB}. 
~A non-exhaustive list includes:
\begin{itemize}[leftmargin=.5cm,itemsep=.01cm,topsep=0cm]
\item The basic \acdef{BM} and the \acdef{OU} processes, both widely adopted as the reference process for \ac{SB}-based models \citep{de2021simulating, de2021diffusion, lavenant2021towards, vargas2021solving, pmlr-v139-wang21l}. We also remark that, even though \eqref{eq:covsol} is known for \ac{BM} \citep{mallasto2021entropy}, what is crucial in these applications is the \ac{SDE} presentation \eqref{eq:GSB-forward-sol}, which is new even for BM.
\item The \acdefp{VESDE}, which underlies the training of \acli{SMLD} for diffusion-based generative modeling \citep{huang2021variational, song2019generative, song2020score}.
\item The \acdefp{VPSDE}, which can be seen as the continuous limit of \aclip{DDPM} \citep{ho2020denoising, sohl2015deep, song2020score}, another important class of algorithms for diffusion-based generative modeling.
\item The \emph{sub-\acp{VPSDE}} proposed by \citep{song2020score}, which are motivated by reducing the variance of \acp{VPSDE}.
\item Several important \acp{SDE} in financial engineering, such as the \emph{Vasicek model} (which generalizes \ac{OU} processes) and the \emph{constant volatility $\shift$-\acdef{BDT} model} \citep{platen2010numerical}. 
\end{itemize}

%----------------------------------------------------------------------
%%% METHOD
%----------------------------------------------------------------------
%\vspace{-8pt}
%\section{Dynamics Reconstruction via \acroalg}
%%\vspace{-8pt}
%\label{sec:methods}
%\input{content/methods.tex}

%----------------------------------------------------------------------
%%% EXPERIMENTS
%----------------------------------------------------------------------
\section{Empirical Evaluation}
\label{sec:experiments}
\begin{figure}[t]
     \centering
         \centering
         \includegraphics[width=\linewidth]{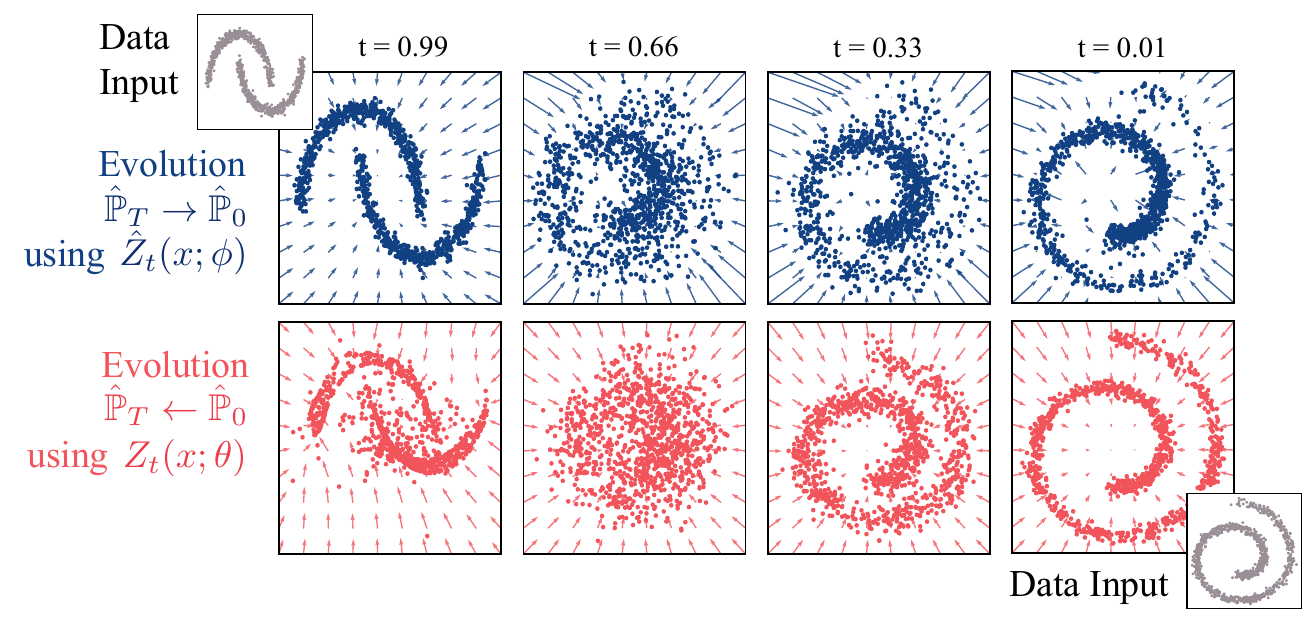}
         \caption{Illustration of the time-dependent drifts learned by \textsc{GSBflow} with VE SDE for two toy marginal distributions. \emph{Top.} Evolution of $\distend$ (moons) $\rightarrow \distinit$ (spiral) via backward policy {\color{blue} $\SBb$}. \emph{Bottom.} Evolution of $\distinit$ (spiral) $\rightarrow \distend$ (moons) via forward policy {\color{pink} $\SBf$}.}
        \label{fig:res_synthetic}
\end{figure}

\begin{table}[t]
    \caption{Evaluation of predictive performance w.r.t. the entropy-regularized Wasserstein distance $W_\varepsilon$ \citep{cuturi2013sinkhorn} of \textsc{GSBflow} and baselines on generating different single-cell datasets (using 3 runs).}
    \label{tab:exp_wasserstein_cells}
    \centering
\adjustbox{max width=.99\linewidth}{%
    \begin{tabular}{lcc}
    \toprule
         \textbf{Method} & \multicolumn{2}{c}{\textbf{Tasks}} \\
         & \multicolumn{2}{c}{\makecell{Wasserstein Loss $W_\varepsilon \downarrow$}} \\
         \cmidrule{2-3}
         & \makecell{\citet{moon2019visualizing}} & \makecell{\citet{schiebinger2019optimal}}  \\
    \midrule
      \citet{song2020score} \\
        \tabindent VESDE & $20.83 \pm 0.18$ & $40.81 \pm 0.42$ \\
        \tabindent sub-VPSDE & $\mathbf{19.96 \pm 0.58}$ & $48.15 \pm 3.38$ \\
      \textsc{GSBflow} (\emph{ours}) \\
         \tabindent VESDE & $25.18 \pm 0.10 $ & $\mathbf{27.85 \pm 0.68}$ \\
    \bottomrule
    \end{tabular}
}
\end{table}

\acused{DDPM}
\acused{SGM}

The purpose of our experiments is to demonstrate that, by leveraging moment information, \textsc{GSBflow} is significantly more stable compared to other \ac{SB}-based objectives, especially when moving beyond the \emph{generative} setting where $\distend$ is a simple Gaussian. Indeed, while performing competitively in the {generative} setting ($\Ncal_0 \rightarrow \distend$), our method \textit{outperforms} when modeling the evolution of two complex distributions ($\distinit \rightarrow \distend$), the most general and ambitious setting to estimate a bridge. This is demonstrated on synthetic data as well as a task from molecular biology concerned with modeling the dynamics of cellular systems, i.e., single-cell genomics \citep{macosko2015highly, frangieh2021multimodal, kulkarni2019beyond}.

\subsection{Synthetic Dynamics}

\label{sec:synthetic}
Before conducting the single-cell genomics experiments, we first test \textsc{GSBflow} on a synthetic setting. 
Our first task involves recovering the stochastic evolution of two-dimensional synthetic data containing two interleaving half circles ($\distend$) into a spiral ($\distinit$). 
\cref{fig:res_synthetic} shows the trajectories learned by \textsc{GSBflow} based on the \ac{VESDE} (see \cref{tab:examples} and \cref{app:vesde}).

While it is sufficient to parameterize only a single policy ({\color{blue} $\SBb$}) in generative modeling, the task of learning to evolve $\distinit$ into $\distend$ requires one to recover \emph{both} vector fields {\color{blue} $\SBb$} and {\color{pink} $\SBf$}.
As demonstrated in \cref{fig:res_synthetic}, \textsc{GSBflow} is able to successfully learn both policies {\color{pink} $\SBf$} and {\color{blue} $\SBb$} and reliably recovers the corresponding targets of the forward and backward evolution. While initializing the reference process through the closed-form SB between the Gaussian approximations of both synthetic datasets provides good results, the power of \textsc{GSBflow} becomes evident in more complex applications which we tackle next.

\begin{figure*}
     \centering
         \centering
         \includegraphics[width=\textwidth]{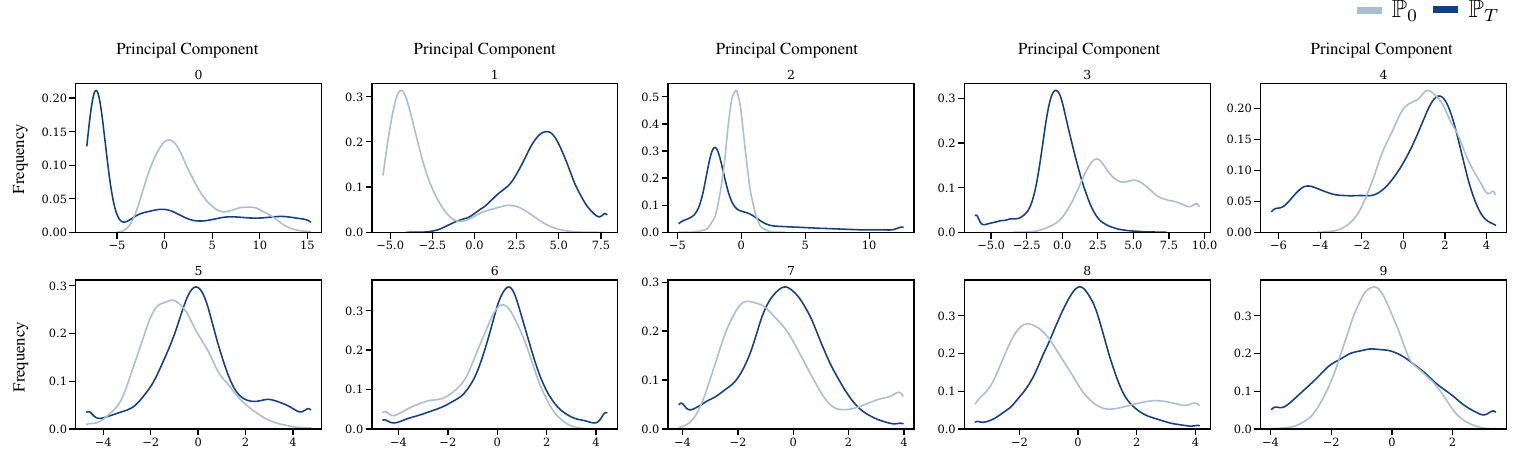}
         \caption{The expression levels of the first 10 principal components from the dataset by \citet{schiebinger2019optimal}.}
\label{fig:gaussianMain}
\end{figure*}

\subsection{Single-Cell Dynamics}

\looseness -1 Modern single-cell profiling technologies are able to provide rich feature representations (e.g., gene expression) of \textit{individual} cells at any development state. A crucial issue that arises with such profiling methods is their destructive nature: Measuring a cell requires destroying it and thus a cell cannot be measured twice. As a result, independent samples are collected at each snapshot, with no access to ground-truth single-cell trajectories throughout time, resulting in challenging, \emph{unaligned}, datasets.
Recovering cellular dynamics from such unaligned snapshots, i.e., $\distinit$ to $\distend$, has, however, extremely important scientific and biomedical relevance \citep{kulkarni2019beyond}. For example, it determines our understanding on how and why tumor cells evade cancer therapies \citep{frangieh2021multimodal} or unveils mechanisms of cell differentiation and development \citep{schiebinger2019optimal}. Following related work, in particular previous methods based on optimal transport \citep{schiebinger2019optimal, bunne2021learning, bunne2022supervised, tong2020trajectorynet}, the task is thus to learn the stochastic process that described the evolution of single cells from $\distinit$ to $\distend$.

\subsubsection{Experimental Setup}

\para{Single-cell genomics via \acp{SB}}
Let us consider the evolution of a gene, for which we can collect the empirical distributions $\distinit,\distend$ of its expression levels at the times $t=0,1$ \citep{schiebinger2019optimal,moon2019visualizing}. Our goal is to two-fold: 
\begin{enumerate}[itemsep=.0cm,topsep=0cm]
\item To solve the \textbf{generative modeling} problem, \ie to generate $\distinit$ or $\distend$ from a standard Gaussian noise, and
\item to \textbf{evolve} $\Pinit\to\Pend$ or $\Pend\to\Pinit$, \ie to recover a stochastic process $\Pmargin$ satisfying $\Pinit = \distinit, \Pend = \distend$.
\end{enumerate}

Although there are numerous algorithms for generative modeling, to our knowledge, the only framework that can simultaneously solve both tasks is the \ac{SB}-based scheme recently proposed in \citep{chen2021likelihood}. In order to apply this framework, one has to choose a prior process $\refsde$, which is taken by the authors to be the high-performing \ac{VESDE} and sub-\ac{VPSDE}. These \ac{SB}-based methods, as well as several standard generative modeling algorithms \citep{ho2020denoising, sohl2015deep, song2020score, huang2021variational, song2019generative, song2020score} for the first task, constitute strong baselines for our experiments.

\para{Our choice of $\refsde$; the \acroalg}

Instead of directly diving into the numerical solution of \acp{SB} as in \citet{chen2021likelihood}, we first empirically verify that the distributions $\distinit,\distend$ in single-cell genomics are typically close to \emph{non-standard} Gaussian distributions: See \cref{fig:gaussianMain} for the canonical dataset \citep{schiebinger2019optimal} and \cref{fig:gaussian} in \cref{app:empirical} for the same phenomenon on another standard benchmark \citep{moon2019visualizing}. 

Since the solutions of \acp{SB} are Lipschitz in terms of $\distinit,\distend$ \citep{carlier2022lipschitz}, a reasonable approximation to the original \ac{SB} objective is to replace $\distinit,\distend$ by Gaussians with matching moments. This results in a \ac{GSB} problem which can be solved in closed form by our \cref{thm:GaussianSB}. Intuitively, if we denote an existing prior process by $\refsde$ and the solution of its corresponding \ac{GSB} by $\Xsol$, then $\Xsol$ presents a more appealing prior process than $\refsde$ since it carries the moment information of $\distinit$ and $\distend$, whereas $\refsde$ is completely data-oblivious.

Motivated by these observations, we propose a simple modification of the framework in \citet{chen2021likelihood}: Replace the prior process $\refsde$ by its \ac{GSB} approximation and keep everything else the same. The resulting scheme, which we term the \acroalg, learns a pair of forward {\color{pink} $\SBf$} and backward parametrized drifts {\color{blue} $\SBb$} that progressively transport samples from $\distinit \rightarrow \distend$ and $\distend \rightarrow \distinit$, respectively. The full algorithm is presented in \cref{sec:methods} for completeness.

\subsubsection{Results}
\label{sec:cell}

\begin{figure*}
     \centering
     \includegraphics[width=\textwidth]{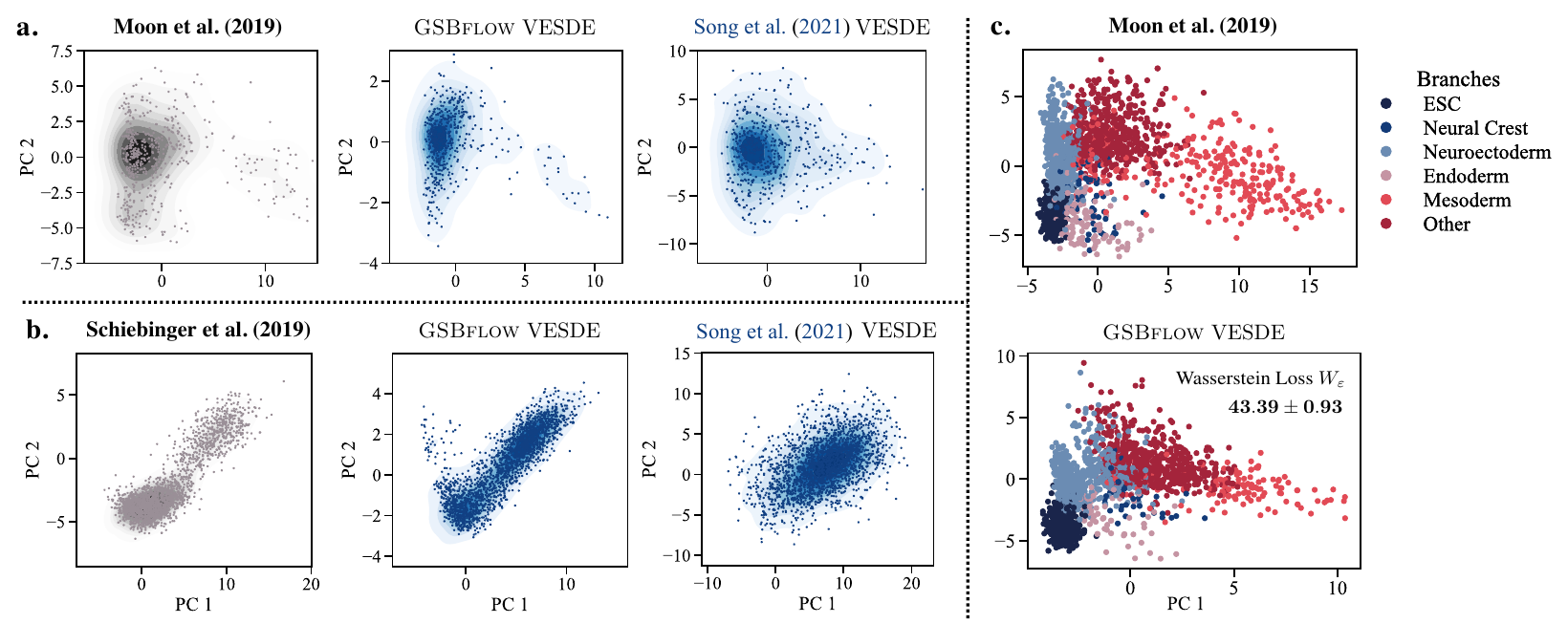}
    \caption{\textbf{a.}-\textbf{b.} Visual evaluation of the ability of our method to model the \textbf{generation} of data from \textbf{a.} \citet{moon2019visualizing} and \textbf{b.} \citet{schiebinger2019optimal}. Density plots are visualized in 2D PCA space and show generated data points using either \textsc{GSBflow} (our method) or the procedure in~\citet{song2020score}. \textbf{c.} Evaluation of \textsc{GSBflow}'s ability to model the entire \textbf{evolution} of a developmental process of \citet{moon2019visualizing}, visualized by the data and \textsc{GSBflow} predictions colored by the lineage branch class.}
    \label{fig:all_results}
\end{figure*}

\looseness -1 We investigate the ability of \textsc{GSBflow} to generate cell populations $\distend$ from noise $\Ncal_0$ ($\Ncal_0 \rightarrow \distend$, Fig.~\ref{fig:all_results}a, b) on the the canonical datasets \citep{moon2019visualizing, schiebinger2019optimal}; as well as to predict the dynamics of single-cell genomics ($\distinit \rightarrow \distend$, Fig.~\ref{fig:all_results}c) \citep{moon2019visualizing}, i.e., the inference of cell populations $\distend$ resulting from the developmental process of an initial cell population $\distinit$, with the goal of learning individual dynamics, identify ancestor and descendant cells. Details on datasets and experimental design can be found in \crefrange{app:datasets}{app:experiments}. % In addition, in order to test our hypothesis that the moment information carried by \textsc{GSBflow} leads to a better overall performance, we scale the datasets \citep{moon2019visualizing, schiebinger2019optimal} by 20 and repeat each training procedure.
% We consider the development of human \acp{ESC} grown as embryoid bodies into diverse cell lineages monitored by single-cell RNA sequencing methods \citep{moon2019visualizing} as well as reprogramming \acp{MEF} into \acp{iPSC} \citep{schiebinger2019optimal}.
The evaluation is conducted on the first 20 or 30 components of the PCA space of the >~1500 highly differentiable genes (see \crefrange{fig:moon_expl_variance}{fig:schiebinger_expl_variance}).

\looseness -1 We evaluate the quality of the generated cellular states through the entropy-regularized Wasserstein distance $W_\varepsilon$
% (\texttt{OTT}\footnote{\url{https://github.com/ott-jax/ott}})
(see \cref{tab:exp_wasserstein_cells}) and by visualizing the first two principal components (PC), see \cref{fig:all_results}a,~b.
\textsc{GSBflow} performs competitively on reconstructing embryoid body differentiation landscapes \citep{moon2019visualizing}, and outperforms score-based generative models baselines on the iPSC reprogramming task \citep{schiebinger2019optimal} as quantified by $W_\varepsilon$ between data and predictions.
Further, we analyze \textsc{GSBflow}'s ability to predict the temporal evolution of embryoid body differentiation \citep{moon2019visualizing}, where cells measured at day 1 to 3 serve as samples of $\distinit$, while $\distend$ is constructed from samples between day 12 to 27. As no ground truth trajectories are available in the data, we compare the predicted evolution to the data and compare how well the heterogeneity of lineage (\cref{fig:all_results}c, upper panel) or sublineage branches (\cref{fig:res_evo_subbranches}a) is captured.
\cref{fig:all_results}c (lower panel) and \cref{fig:res_evo_subbranches}b thereby closely resemble the data (see $W_\varepsilon$ in \cref{fig:all_results}c) and thus demonstrate \textsc{GSBflow}'s ability to learn cell differentiation into various lineages and to capture biological heterogeneity on a more macroscopic level.

%----------------------------------------------------------------------
%%% CONCLUSION
%----------------------------------------------------------------------
\vspace{-5pt}
\section{Conclusion and Future Work}
\vspace{-5pt}
\label{sec:conclusion}
\looseness -1 We derive closed-form solutions of \acp{GSB}, an important class of dynamic \ac{OT} problems. Our technique originates from a deep connection between Gaussian \ac{OT} and the Bures-Wasserstein geometry, which we generalize to the case of general \ac{SB} problems. Numerically, we demonstrate that our new closed forms inspire a simple modification of existing \ac{SB}-based numerical schemes, which can however lead to significantly improved performance.

\textbf{Limitation of our framework. }
\looseness -1 In a broader context, we hope our results can serve as the inspiration for more learning algorithms, much like how existing closed-form solutions of Gaussian \ac{OT} problems have contributed to the machine learning community. We thus acknowledge a severe limitation of our closed-form solutions: These formulas require matrix inversions, which might face scalability issues for high-dimensional data. In addition, existing matrix inversion algorithms are typically extremely sensitive to the condition number, and thus our formulas are not as useful for ill-conditioned data. Lifting these constraints to facilitate further applications, such as to image datasets, is an important future work.

\subsection*{Acknowledgments}
This research was supported by the European Research Council (ERC) under the European Union’s Horizon 2020 research and innovation program grant agreement no. 815943 and created as part of NCCR Catalysis (grant number 180544), a National Centre of Competence in Research funded by the Swiss National Science Foundation.
Ya-Ping Hsieh acknowledges funding through an ETH Foundations of Data Science (ETH-FDS) postdoctoral fellowship.

\bibliographystyle{abbrvnat}
\bibliography{references}

\begin{thebibliography}{62}
\providecommand{\natexlab}[1]{#1}
\providecommand{\url}[1]{\texttt{#1}}
\expandafter\ifx\csname urlstyle\endcsname\relax
  \providecommand{\doi}[1]{doi: #1}\else
  \providecommand{\doi}{doi: \begingroup \urlstyle{rm}\Url}\fi

\bibitem[Altschuler et~al.(2021)Altschuler, Chewi, Gerber, and
  Stromme]{altschuler2021averaging}
J.~Altschuler, S.~Chewi, P.~R. Gerber, and A.~Stromme.
\newblock {Averaging on the Bures-Wasserstein manifold: dimension-free
  convergence of gradient descent}.
\newblock \emph{Advances in Neural Information Processing Systems (NeurIPS)},
  34, 2021.

\bibitem[Ambrosio et~al.(2006)Ambrosio, Gigli, and
  Savar{\'e}]{ambrosio2006gradient}
L.~Ambrosio, N.~Gigli, and G.~Savar{\'e}.
\newblock \emph{{Gradient Flows in Metric Spaces and in the Space of
  Probability Measures}}.
\newblock Springer, 2006.

\bibitem[Bernton et~al.(2019)Bernton, Heng, Doucet, and Jacob]{bernton2019schr}
E.~Bernton, J.~Heng, A.~Doucet, and P.~E. Jacob.
\newblock {Schr\"odinger Bridge Samplers}.
\newblock In \emph{arXiv preprint arXiv:1912.13170}, 2019.

\bibitem[Bhatia et~al.(2019)Bhatia, Jain, and Lim]{bhatia2019bures}
R.~Bhatia, T.~Jain, and Y.~Lim.
\newblock {On the Bures--Wasserstein distance between positive definite
  matrices}.
\newblock \emph{Expositiones Mathematicae}, 37\penalty0 (2), 2019.

\bibitem[Bojilov and Galichon(2016)]{bojilov2016matching}
R.~Bojilov and A.~Galichon.
\newblock {Matching in Closed-Form: Equilibrium, Identification, and
  Comparative Statics}.
\newblock \emph{Economic Theory}, 61\penalty0 (4), 2016.

\bibitem[Bonneel et~al.(2015)Bonneel, Rabin, Peyr{\'e}, and
  Pfister]{bonneel2015sliced}
N.~Bonneel, J.~Rabin, G.~Peyr{\'e}, and H.~Pfister.
\newblock {Sliced and Radon Wasserstein Barycenters of Measures}.
\newblock \emph{Journal of Mathematical Imaging and Vision}, 51\penalty0 (1),
  2015.

\bibitem[Bunne et~al.(2021)Bunne, Stark, Gut, del Castillo, Lehmann, Pelkmans,
  Krause, and Ratsch]{bunne2021learning}
C.~Bunne, S.~G. Stark, G.~Gut, J.~S. del Castillo, K.-V. Lehmann, L.~Pelkmans,
  A.~Krause, and G.~Ratsch.
\newblock {Learning Single-Cell Perturbation Responses using Neural Optimal
  Transport}.
\newblock \emph{bioRxiv}, 2021.

\bibitem[Bunne et~al.(2022)Bunne, Krause, and Cuturi]{bunne2022supervised}
C.~Bunne, A.~Krause, and M.~Cuturi.
\newblock {Supervised Training of Conditional Monge Maps}.
\newblock In \emph{Advances in Neural Information Processing Systems
  (NeurIPS)}, 2022.

\bibitem[Carlier et~al.(2022)Carlier, Chizat, and
  Laborde]{carlier2022lipschitz}
G.~Carlier, L.~Chizat, and M.~Laborde.
\newblock {Lipschitz Continuity of the {S}chr\"odinger Map in Entropic Optimal
  Transport}.
\newblock \emph{arXiv preprint arXiv:2210.00225}, 2022.

\bibitem[Chen et~al.(2022)Chen, Liu, and Theodorou]{chen2021likelihood}
T.~Chen, G.-H. Liu, and E.~A. Theodorou.
\newblock {Likelihood Training of {S}chr\"{o}dinger Bridge using
  Forward-Backward {SDE}s Theory}.
\newblock In \emph{International Conference on Learning Representations
  (ICLR)}, 2022.

\bibitem[Chen et~al.(2015)Chen, Georgiou, and Pavon]{chen2015optimal}
Y.~Chen, T.~T. Georgiou, and M.~Pavon.
\newblock {Optimal Steering of a Linear Stochastic System to a Final
  Probability Distribution--Part III}.
\newblock \emph{IEEE Transactions on Automatic Control}, 61\penalty0 (5), 2015.

\bibitem[Chen et~al.(2016)Chen, Georgiou, and Pavon]{chen2016relation}
Y.~Chen, T.~T. Georgiou, and M.~Pavon.
\newblock On the relation between optimal transport and {S}chr{\"o}dinger
  bridges: A stochastic control viewpoint.
\newblock \emph{Journal of Optimization Theory and Applications}, 169\penalty0
  (2), 2016.

\bibitem[Chen et~al.(2019)Chen, Shi, and Zhang]{chen2018optimal}
Y.~Chen, Y.~Shi, and B.~Zhang.
\newblock {Optimal Control Via Neural Networks: A Convex Approach}.
\newblock In \emph{International Conference on Learning Representations
  (ICLR)}, 2019.

\bibitem[Chen et~al.(2021)Chen, Georgiou, and Pavon]{chen2021stochastic}
Y.~Chen, T.~T. Georgiou, and M.~Pavon.
\newblock {Stochastic Control Liaisons: Richard Sinkhorn Meets Gaspard Monge on
  a Schrödinger Bridge}.
\newblock \emph{SIAM Review}, 63\penalty0 (2), 2021.

\bibitem[Chewi et~al.(2020)Chewi, Maunu, Rigollet, and
  Stromme]{chewi2020gradient}
S.~Chewi, T.~Maunu, P.~Rigollet, and A.~J. Stromme.
\newblock {Gradient descent algorithms for Bures-Wasserstein barycenters}.
\newblock In \emph{Conference on Learning Theory (COLT)}. PMLR, 2020.

\bibitem[Cuturi(2013)]{cuturi2013sinkhorn}
M.~Cuturi.
\newblock {Sinkhorn Distances: Lightspeed Computation of Optimal Transport}.
\newblock In \emph{Advances in Neural Information Processing Systems
  (NeurIPS)}, volume~26, 2013.

\bibitem[De~Bortoli et~al.(2021{\natexlab{a}})De~Bortoli, Doucet, Heng, and
  Thornton]{de2021simulating}
V.~De~Bortoli, A.~Doucet, J.~Heng, and J.~Thornton.
\newblock {Simulating Diffusion Bridges with Score Matching}.
\newblock In \emph{arXiv preprint arXiv:2111.07243}, 2021{\natexlab{a}}.

\bibitem[De~Bortoli et~al.(2021{\natexlab{b}})De~Bortoli, Thornton, Heng, and
  Doucet]{de2021diffusion}
V.~De~Bortoli, J.~Thornton, J.~Heng, and A.~Doucet.
\newblock {Diffusion Schr\"odinger Bridge with Applications to Score-Based
  Generative Modeling}.
\newblock In \emph{Advances in Neural Information Processing Systems
  (NeurIPS)}, volume~35, 2021{\natexlab{b}}.

\bibitem[del Barrio and Loubes(2020)]{del2020statistical}
E.~del Barrio and J.-M. Loubes.
\newblock The statistical effect of entropic regularization in optimal
  transportation.
\newblock \emph{arXiv preprint arXiv:2006.05199}, 2020.

\bibitem[Dowson and Landau(1982)]{dowson1982frechet}
D.~Dowson and B.~Landau.
\newblock {The Fr\'echet Distance between Multivariate Normal Distributions}.
\newblock \emph{Journal of Multivariate Analysis}, 12\penalty0 (3), 1982.

\bibitem[Frangieh et~al.(2021)Frangieh, Melms, Thakore, Geiger-Schuller, Ho,
  Luoma, Cleary, Jerby-Arnon, Malu, Cuoco, et~al.]{frangieh2021multimodal}
C.~J. Frangieh, J.~C. Melms, P.~I. Thakore, K.~R. Geiger-Schuller, P.~Ho, A.~M.
  Luoma, B.~Cleary, L.~Jerby-Arnon, S.~Malu, M.~S. Cuoco, et~al.
\newblock {Multimodal pooled Perturb-CITE-seq screens in patient models define
  mechanisms of cancer immune evasion}.
\newblock \emph{Nature Genetics}, 53\penalty0 (3), 2021.

\bibitem[Gentil et~al.(2017)Gentil, L{\'e}onard, and Ripani]{gentil2017analogy}
I.~Gentil, C.~L{\'e}onard, and L.~Ripani.
\newblock About the analogy between optimal transport and minimal entropy.
\newblock In \emph{Annales de la Facult{\'e} des sciences de Toulouse:
  Math{\'e}matiques}, volume~26, 2017.

\bibitem[Gentil et~al.(2020)Gentil, L{\'e}onard, and
  Ripani]{gentil2020dynamical}
I.~Gentil, C.~L{\'e}onard, and L.~Ripani.
\newblock {Dynamical aspects of the generalized Schr{\"o}dinger problem via
  Otto calculus--A heuristic point of view}.
\newblock \emph{Revista Matem{\'a}tica Iberoamericana}, 36\penalty0 (4), 2020.

\bibitem[Han et~al.(2021)Han, Mishra, Jawanpuria, and Gao]{han2021riemannian}
A.~Han, B.~Mishra, P.~K. Jawanpuria, and J.~Gao.
\newblock {On Riemannian optimization over positive definite matrices with the
  Bures-Wasserstein geometry}.
\newblock \emph{Advances in Neural Information Processing Systems (NeurIPS)},
  34, 2021.

\bibitem[Ho et~al.(2020)Ho, Jain, and Abbeel]{ho2020denoising}
J.~Ho, A.~Jain, and P.~Abbeel.
\newblock {Denoising Diffusion Probabilistic Models}.
\newblock In \emph{Advances in Neural Information Processing Systems
  (NeurIPS)}, 2020.

\bibitem[Holdijk et~al.(2022)Holdijk, Du, Hooft, Jaini, Ensing, and
  Welling]{holdijk2022path}
L.~Holdijk, Y.~Du, F.~Hooft, P.~Jaini, B.~Ensing, and M.~Welling.
\newblock {Path Integral Stochastic Optimal Control for Sampling Transition
  Paths}.
\newblock \emph{arXiv preprint arXiv:2207.02149}, 2022.

\bibitem[Huang et~al.(2021{\natexlab{a}})Huang, Lim, and
  Courville]{huang2021variational}
C.-W. Huang, J.~H. Lim, and A.~Courville.
\newblock {A Variational Perspective on Diffusion-Based Generative Models and
  Score Matching}.
\newblock In \emph{Advances in Neural Information Processing Systems
  (NeurIPS)}, 2021{\natexlab{a}}.

\bibitem[Huang et~al.(2021{\natexlab{b}})Huang, Jiao, Kang, Liao, Liu, and
  Liu]{huang2021schrodinger}
J.~Huang, Y.~Jiao, L.~Kang, X.~Liao, J.~Liu, and Y.~Liu.
\newblock {Schr{\"o}dinger-F{\"o}llmer Sampler: Sampling without Ergodicity}.
\newblock \emph{arXiv preprint arXiv:2106.10880}, 2021{\natexlab{b}}.

\bibitem[Janati et~al.(2020)Janati, Muzellec, Peyr{\'e}, and
  Cuturi]{janati2020entropic}
H.~Janati, B.~Muzellec, G.~Peyr{\'e}, and M.~Cuturi.
\newblock {Entropic Optimal Transport between Unbalanced Gaussian Measures has
  a Closed Form}.
\newblock In \emph{Advances in Neural Information Processing Systems
  (NeurIPS)}, volume~33, 2020.

\bibitem[Kingma and Ba(2014)]{kingma2014adam}
D.~P. Kingma and J.~Ba.
\newblock {Adam: A Method for Stochastic Optimization}.
\newblock In \emph{International Conference on Learning Representations
  (ICLR)}, 2014.

\bibitem[Kulkarni et~al.(2019)Kulkarni, Anderson, Merullo, and
  Konopka]{kulkarni2019beyond}
A.~Kulkarni, A.~G. Anderson, D.~P. Merullo, and G.~Konopka.
\newblock Beyond bulk: a review of single cell transcriptomics methodologies
  and applications.
\newblock \emph{Current Opinion in Biotechnology}, 58:\penalty0 129--136, 2019.

\bibitem[Lavenant et~al.(2021)Lavenant, Zhang, Kim, and
  Schiebinger]{lavenant2021towards}
H.~Lavenant, S.~Zhang, Y.-H. Kim, and G.~Schiebinger.
\newblock Towards a mathematical theory of trajectory inference.
\newblock \emph{arXiv preprint arXiv:2102.09204}, 2021.

\bibitem[L{\'e}onard(2013)]{leonard2013survey}
C.~L{\'e}onard.
\newblock A survey of the {S}chr\"{o}dinger problem and some of its connections
  with optimal transport.
\newblock \emph{arXiv preprint arXiv:1308.0215}, 2013.

\bibitem[Liu et~al.(2022)Liu, Chen, So, and Theodorou]{liu2022deep}
G.-H. Liu, T.~Chen, O.~So, and E.~A. Theodorou.
\newblock {Deep Generalized Schr\"odinger Bridge}.
\newblock In \emph{Advances in Neural Information Processing Systems
  (NeurIPS)}, 2022.

\bibitem[Luecken and Theis(2019)]{luecken2019current}
M.~D. Luecken and F.~J. Theis.
\newblock {Current best practices in single-cell RNA-seqanalysis: a tutorial}.
\newblock \emph{Molecular Systems Biology}, 15\penalty0 (6), 2019.

\bibitem[Macosko et~al.(2015)Macosko, Basu, Satija, Nemesh, Shekhar, Goldman,
  Tirosh, Bialas, Kamitaki, Martersteck, et~al.]{macosko2015highly}
E.~Z. Macosko, A.~Basu, R.~Satija, J.~Nemesh, K.~Shekhar, M.~Goldman,
  I.~Tirosh, A.~R. Bialas, N.~Kamitaki, E.~M. Martersteck, et~al.
\newblock Highly parallel genome-wide expression profiling of individual cells
  using nanoliter droplets.
\newblock \emph{Cell}, 161\penalty0 (5):\penalty0 1202--1214, 2015.

\bibitem[Mallasto et~al.(2021)Mallasto, Gerolin, and Minh]{mallasto2021entropy}
A.~Mallasto, A.~Gerolin, and H.~Q. Minh.
\newblock Entropy-regularized 2-{W}asserstein distance between {G}aussian
  measures.
\newblock \emph{Information Geometry}, pages 1--35, 2021.

\bibitem[Mansuy and Yor(2008)]{mansuy2008aspects}
R.~Mansuy and M.~Yor.
\newblock \emph{{Aspects of Brownian motion}}.
\newblock Springer Science \& Business Media, 2008.

\bibitem[Martin and Evans(1975)]{martin1975}
G.~R. Martin and M.~J. Evans.
\newblock {Differentiation of Clonal Lines of Teratocarcinoma Cells: Formation
  of Embryoid Bodies In Vitro}.
\newblock \emph{Proceedings of the National Academy of Sciences}, 72\penalty0
  (4), 1975.

\bibitem[McCann(1997)]{mccann1997convexity}
R.~J. McCann.
\newblock A convexity principle for interacting gases.
\newblock \emph{Advances in Mathematics}, 128\penalty0 (1), 1997.

\bibitem[Moon et~al.(2019)Moon, van Dijk, Wang, Gigante, Burkhardt, Chen, Yim,
  van~den Elzen, Hirn, Coifman, et~al.]{moon2019visualizing}
K.~R. Moon, D.~van Dijk, Z.~Wang, S.~Gigante, D.~B. Burkhardt, W.~S. Chen,
  K.~Yim, A.~van~den Elzen, M.~J. Hirn, R.~R. Coifman, et~al.
\newblock {Visualizing structure and transitions in high-dimensional biological
  data}.
\newblock \emph{Nature Biotechnology}, 37\penalty0 (12), 2019.

\bibitem[Olkin and Pukelsheim(1982)]{olkin1982distance}
I.~Olkin and F.~Pukelsheim.
\newblock The distance between two random vectors with given dispersion
  matrices.
\newblock \emph{Linear Algebra and its Applications}, 48, 1982.

\bibitem[Otto(2001)]{otto2001geometry}
F.~Otto.
\newblock The geometry of dissipative evolution equations: the porous medium
  equation.
\newblock \emph{Taylor \& Francis}, 2001.

\bibitem[Peyré and Cuturi(2019)]{peyre2019computational}
G.~Peyré and M.~Cuturi.
\newblock {Computational Optimal Transport}.
\newblock \emph{Foundations and Trends in Machine Learning}, 11\penalty0 (5-6),
  2019.

\bibitem[Platen and Bruti-Liberati(2010)]{platen2010numerical}
E.~Platen and N.~Bruti-Liberati.
\newblock \emph{{Numerical Solution of Stochastic Differential Equations with
  Jumps in Finance}}, volume~64.
\newblock Springer Science \& Business Media, 2010.

\bibitem[Protter(2005)]{protter2005stochastic}
P.~E. Protter.
\newblock {Stochastic Differential Equations}.
\newblock In \emph{{Stochastic Integration and Differential Equations}}, pages
  249--361. Springer, 2005.

\bibitem[Rabin et~al.(2011)Rabin, Peyr{\'e}, Delon, and
  Bernot]{rabin2011wasserstein}
J.~Rabin, G.~Peyr{\'e}, J.~Delon, and M.~Bernot.
\newblock {Wasserstein Barycenter and Its Application to Texture Mixing}.
\newblock In \emph{International Conference on Scale Space and Variational
  Methods in Computer Vision}. Springer, 2011.

\bibitem[Schiebinger et~al.(2019)Schiebinger, Shu, Tabaka, Cleary, Subramanian,
  Solomon, Gould, Liu, Lin, Berube, et~al.]{schiebinger2019optimal}
G.~Schiebinger, J.~Shu, M.~Tabaka, B.~Cleary, V.~Subramanian, A.~Solomon,
  J.~Gould, S.~Liu, S.~Lin, P.~Berube, et~al.
\newblock {Optimal-Transport Analysis of Single-Cell Gene Expression Identifies
  Developmental Trajectories in Reprogramming}.
\newblock \emph{Cell}, 176\penalty0 (4), 2019.

\bibitem[Shamblott et~al.(2009)Shamblott, Kerr, Axelman, Littlefield, Clark,
  Patterson, Addis, Kraszewski, Kent, and Gearhart]{shamblott2009derivation}
M.~J. Shamblott, C.~L. Kerr, J.~Axelman, J.~W. Littlefield, G.~O. Clark, E.~S.
  Patterson, R.~C. Addis, J.~N. Kraszewski, K.~C. Kent, and J.~D. Gearhart.
\newblock {Derivation and Differentiation of Human Embryonic Germ Cells}.
\newblock In \emph{Essentials of Stem Cell Biology}. Elsevier, 2009.

\bibitem[Sohl-Dickstein et~al.(2015)Sohl-Dickstein, Weiss, Maheswaranathan, and
  Ganguli]{sohl2015deep}
J.~Sohl-Dickstein, E.~Weiss, N.~Maheswaranathan, and S.~Ganguli.
\newblock {Deep Unsupervised Learning using Nonequilibrium Thermodynamics}.
\newblock In \emph{International Conference on Machine Learning (ICML)}, 2015.

\bibitem[Song and Ermon(2019)]{song2019generative}
Y.~Song and S.~Ermon.
\newblock {Generative Modeling by Estimating Gradients of the Data
  Distribution}.
\newblock In \emph{Advances in Neural Information Processing Systems
  (NeurIPS)}, 2019.

\bibitem[Song et~al.(2021)Song, Sohl-Dickstein, Kingma, Kumar, Ermon, and
  Poole]{song2020score}
Y.~Song, J.~Sohl-Dickstein, D.~P. Kingma, A.~Kumar, S.~Ermon, and B.~Poole.
\newblock {Score-Based Generative Modeling through Stochastic Differential
  Equations}.
\newblock In \emph{International Conference on Learning Representations
  (ICLR)}, volume~9, 2021.

\bibitem[Takatsu(2010)]{takatsu2010wasserstein}
A.~Takatsu.
\newblock {On Wasserstein geometry of Gaussian measures}.
\newblock In \emph{{Probabilistic Approach to Geometry}}. Mathematical Society
  of Japan, 2010.

\bibitem[Tong et~al.(2020)Tong, Huang, Wolf, Van~Dijk, and
  Krishnaswamy]{tong2020trajectorynet}
A.~Tong, J.~Huang, G.~Wolf, D.~Van~Dijk, and S.~Krishnaswamy.
\newblock {TrajectoryNet: A Dynamic Optimal Transport Network for Modeling
  Cellular Dynamics}.
\newblock In \emph{International Conference on Machine Learning (ICML)}, 2020.

\bibitem[user26872(2012)]{126767}
user26872.
\newblock Reference for {M}ultidimensional {G}aussian {I}ntegral.
\newblock Mathematics Stack Exchange, 2012.
\newblock URL \url{https://math.stackexchange.com/q/126767}.

\bibitem[Vargas et~al.(2021)Vargas, Thodoroff, Lawrence, and
  Lamacraft]{vargas2021solving}
F.~Vargas, P.~Thodoroff, N.~D. Lawrence, and A.~Lamacraft.
\newblock {Solving {S}chr\"odinger Bridges via Maximum Likelihood}.
\newblock \emph{Entropy}, 23\penalty0 (9), 2021.

\bibitem[Vayer et~al.(2019)Vayer, Flamary, Tavenard, Chapel, and
  Courty]{vayer2019sliced}
T.~Vayer, R.~Flamary, R.~Tavenard, L.~Chapel, and N.~Courty.
\newblock {Sliced Gromov-Wasserstein}.
\newblock In \emph{Advances in Neural Information Processing Systems
  (NeurIPS)}, volume~32, 2019.

\bibitem[Villani(2009)]{villani2009optimal}
C.~Villani.
\newblock \emph{Optimal transport: old and new}, volume 338.
\newblock Springer, 2009.

\bibitem[Wang et~al.(2021)Wang, Jiao, Xu, Wang, and Yang]{pmlr-v139-wang21l}
G.~Wang, Y.~Jiao, Q.~Xu, Y.~Wang, and C.~Yang.
\newblock {Deep Generative Learning via {S}chr\"odinger Bridge}.
\newblock In \emph{International Conference on Machine Learning (ICML)}, 2021.

\bibitem[Wolf et~al.(2018)Wolf, Angerer, and Theis]{wolf2018scanpy}
F.~A. Wolf, P.~Angerer, and F.~J. Theis.
\newblock {SCANPY: large-scale single-cell gene expression data analysis}.
\newblock \emph{Genome Biology}, 19\penalty0 (1), 2018.

\bibitem[Zee(2010)]{zee2010quantum}
A.~Zee.
\newblock \emph{{Quantum Field Theory in a Nutshell}}, volume~7.
\newblock Princeton University Press, 2010.

\bibitem[Zheng et~al.(2017)Zheng, Terry, Belgrader, Ryvkin, Bent, Wilson,
  Ziraldo, Wheeler, McDermott, Zhu, et~al.]{zheng2017massively}
G.~X. Zheng, J.~M. Terry, P.~Belgrader, P.~Ryvkin, Z.~W. Bent, R.~Wilson, S.~B.
  Ziraldo, T.~D. Wheeler, G.~P. McDermott, J.~Zhu, et~al.
\newblock Massively parallel digital transcriptional profiling of single cells.
\newblock \emph{Nature Communications}, 8\penalty0 (1), 2017.

\end{thebibliography}

\newpage
\appendix
\clearpage
\onecolumn
\numberwithin{equation}{section}		% for numbering  in the appendix
\numberwithin{lemma}{section}		% for numbering  in the appendix
\numberwithin{proposition}{section}		% for numbering  in the appendix
\numberwithin{theorem}{section}		% for numbering in the appendix

%----------------------------------------------------------------------
%%% PROOF OF BURES-WASSERSTEIN MECHANICS
%----------------------------------------------------------------------
\section{Proof of \cref{thm:GSBtoBB}}
\label{app:GSBtoBB}

It is known that, for \acp{SB}, the optimal solution can be searched within the class of stochastic processes \citep{leonard2013survey}
\begin{equation}
\label{eq:holddd}
X_t \sim \Pmargin: \quad\dd X_t =  (\driftf(X_t) + \vectalt(X_t) )\dt + \volat \dWiener[\ctime].
\end{equation}
The Fokker-Planck equation for the \ac{SDE} \eqref{eq:holddd} is
\begin{align}
\label{eq:FK}
\del_t \measure = - \Div \parens*{  \measure(  \driftf +\vectalt  ) } + \frac{\volatsq[\ctime]}{2}\Laplace \measure.
\end{align}
A simple application of the Girsanov's theorem then shows, up to a constant, 
\begin{align}
\label{eq:KLobj}
\KL{ \Pmargin }{ \refsde} = \exof*{ \int_0^\horizon\frac{\norm{\vectalt}^2}{2\volatsq[\ctime]}   \dt}.
\end{align}
Using a change of variable $\vect = \vectalt - \frac{\volatsq[\ctime]}{2}\nabla\log\measure$, we see that \eqref{eq:FK} is equivalent to
\begin{equation}
\del_t \measure = - \Div \parens*{  \measure(  \driftf +\vect  ) }.
\end{equation}
On the other hand, since $\norm{\vectalt}^2 = \norm{\vect}^2 + \frac{\volat^4}{4}\norm*{\nabla\log\measure}^2 + 2 \inner*{\vect}{ \frac{\volatsq[\ctime]}{2}\nabla\log\measure}$, the integrand in the objective of \eqref{eq:KLobj} becomes 
\begin{align}
 \mathbb{E}\Bigg[\int_0^\horizon  \frac{\norm{\vect}^2}{2\volatsq[\ctime]} + \frac{\volatsq[\ctime]}{8} \norm{\nabla \log \measure}^2  + \frac{1}{2} \inner{\vect}{\nabla \log\measure} \dt\Bigg].
\end{align}

Letting $H(\measure) \defeq \int \measure\log\measure$ be the entropy, we have
\begin{align*}
H(\measure[\horizon]) - H(\measure[0]) &= \int_0^\horizon \del_t H(\measure) \dt \\
&= \int_0^\horizon  \int   (1+ \log\measure) \del_t\measure \drm \point \dt  \\
&= \int_0^\horizon  \int   (1+ \log\measure) \cdot \parens*{ - \Div \parens*{ \measure(\driftf+\vect) }  } \drm \point \dt  \quad\quad \text{by \eqref{eq:FK}}\\
&= \int_0^\horizon  \int   \measure \inner{\nabla\log\measure}{ \driftf+\vect} \drm \point \dt 
\end{align*}
by integration by parts for the divergence operator. Therefore,
\begin{align}
\exof*{\int_0^\horizon  \inner{\nabla\log\measure}{ \vect}  \dt } = H(\measure[\horizon]) - H(\measure[0])-\exof*{\int_0^\horizon  \inner{\nabla\log\measure}{ \driftf}  \dt }
\end{align} 
which concludes the proof.\hfill$\qed$

%----------------------------------------------------------------------
%%% PROOF OF BURES-WASSERSTEIN MECHANICS
%----------------------------------------------------------------------
\section{The Bures-Wasserstein Geometry of Gaussian Schr\"odinger Bridges}
\label{app:mechanics}
\subsection{Review of Bures-Wasserstein Geometry}

\label{app:reviewBW}

Recall that the \emph{metric tensor} $\innerBW[\cdot][\cdot][\Sigma]$ in the \emph{Bures-Wasserstein geometry} \citep{takatsu2010wasserstein} is defined in terms of the Lyapunov operator: 
\begin{align}
\label{eq:BW-tensor}
\forall\ U, V\in \tspace, \quad \innerBW[U][V][\Sigma] \defeq \tr\lyap[\Sigma][U]\Sigma\lyap[\Sigma][V] = \frac{1}{2}\tr \lyap[\Sigma][U]V.
\end{align}The corresponding Bures-Wasserstein norm is induced via $\normBWsq[U][\Sigma] \defeq \innerBW[U][U][\Sigma]$.
Another important operator is the Bures-Wasserstein \emph{gradient}: For any function $F \from \SPD \to \R$, 
\begin{equation}
\label{eq:gradBW-def}
\tspace\ni \gradBW F(\Sigma) \defeq 2 \parens*{ \nabla F(\Sigma)\Sigma + \Sigma \nabla F(\Sigma)^\top }
\end{equation}where $\nabla$ is the usual Euclidean gradient of $F$, viewed as a function from $\R^{\vdim\times\vdim}$ to $\R$. Note that
\begin{align}
\lyap[\Sigma][\gradBW F(\Sigma)] &= 2 \lyap[\Sigma][\nabla F(\Sigma)\Sigma + \Sigma\nabla F(\Sigma)] \\
&= 2 \nabla F(\Sigma)
\end{align}by definition of the Lyapunov operator. In other words,
\begin{align}
\label{eq:gradBW-lyapinv}
\gradBW F(\Sigma) = \lyapinv.
\end{align}

Lastly, we recall the Bures-Wasserstein \emph{acceleration} of a curve $\Sigcurve: [0,\horizon] \to \SPD$, which we denote by $\nabla_{\ssstyle\dSigcurve}\dSigcurve$:\footnote{More formally, $\nabla_{\ssstyle\dSigcurve}\dSigcurve$ is the Bures-Wasserstein covariant derivative of $\dSigcurve$ in the direction of $\dSigcurve$.}
%To this end, let $X$ be any \emph{vector field} extending $\dSigcurve$ to the whole space.\footnote{It is well-known that such extensions always exist, and the Riemannian acceleration does not depend on which extension we choose \citep{villani2009optimal}.} Let $\dder$ denote the vector field whose value at
%point $\Sigma$ is the derivative at $\Sigma$ of $X$ in the direction $X(\Sigma)$:
%\begin{align}
%\dder (\Sigma) \defeq \lim_{h\to 0 } \frac{  X(\Sigma + h X(\Sigma)) - X(\Sigma)  }{h}.
%\end{align}Then we have
\begin{align}
\label{eq:coderBW}
\coder = \ddSigcurve -  \parens*{  \lyap \dSigcurve + \dSigcurve\lyap } + \parens*{ \Sigcurve \parens*{\lyap}^2 + \parens*{\lyap}^2 \Sigcurve}.
\end{align}

\subsection{Proof of \cref{thm:mechanics}}
\label{app:proofmechanics}

For convenience, we restate \cref{thm:mechanics} in full below: 
\mechanics*

%Without loss of generality, we assume $\horizon = 1$ (the general case follows by a simple rescaling argument).
%\para{Proof outline} 
%We will prove \cref{thm:mechanics} in two steps: 
%
%\begin{enumerate}
%\item We show that the following curve $\Sigcurve$ verifies \eqref{eq:EL-BW}, the Euler-Lagrange equation in the Bures-Wasserstein geometry:
%\begin{equation}
%\label{eq:sol}
%\Sigmasol \defeq \ctimebar^2 \Sigma + \ctime^2 \Sigma' + \ctime\cdot\ctimebar\parens*{\C+\C^\top+ \sdev^2 \eye}.
%\end{equation}Here, $\ctimebar \defeq 1-\ctime$ and $\C$ is defined in \eqref{eq:Cstar}. %with $\Sigcurve[0] \subs \Sigma$ and $\Sigcurve[\horizon] \subs \Sigma'$. 
%~In particular, the curve in \eqref{eq:sol} is the optimal solution to the action minimization problem \eqref{eq:LagBW}.
%
%\item We show that the minimizer of \eqref{eq:GaussianSBWiener} coincides with \eqref{eq:LagBW}. This step heavily relies on a previous result in \citep{chen2016relation, gentil2017analogy}.
%\end{enumerate}

%by showing  It can be readily verified that the boundary conditions in \eqref{eq:EL-BW} hold for the curve in \eqref{eq:sol}.%$\Sigcurve[0] = \Sigma$ and $\Sigcurve[\horizon] = \Sigma'$.

The proof consists of verifying the Euler-Lagrange equation \eqref{eq:EL-BW} for the curve \eqref{eq:solMain}.

\subsubsection{Verifying the Euler-Lagrange Equation \eqref{eq:EL-BW}}
\label{sec:verifyEL}
We begin by noting that the boundary conditions in \eqref{eq:EL-BW} hold for the curve in \eqref{eq:solMain}.

We now compute the two sides of \eqref{eq:EL-BW} separately:

\textbf{The \acl{RHS} of \eqref{eq:EL-BW}: $-\gradBW\penergyBW$.}
~Since $\nabla \penergyBW = -\nabla \parens*{\tr \frac{\sdev^4}{8} \Sigcurveinv} =  \frac{\sdev^4}{8} \Sigcurveinv\cdot \Sigcurveinv$, we see from \eqref{eq:gradBW-def} that the negative Bures-Wasserstein gradient of $\penergyBW$ is
\begin{align}
\nn
-\gradBW \penergyBW &= -2\parens*{\frac{\sdev^4}{8} \Sigcurveinv\cdot \Sigcurveinv \cdot \Sigcurve + \Sigcurve\cdot \frac{\sdev^4}{8} \Sigcurveinv\cdot \Sigcurveinv } \\
\label{eq:gradBWU}
&= -\frac{\sdev^4}{2} \Sigcurveinv.
\end{align}

%is the symmetric solution to the matrix equation:
%\begin{equation}
%A: \quad X\Sigcurve +  X = \Sigcurve
%\end{equation}
%Recalling the 

\textbf{The \acl{LHS} of \eqref{eq:EL-BW}: $\coder$.}
~Computing $\coder$ is significantly trickier than $-\gradBW\penergyBW$. The central piece of the proof is the following technical lemma:%solution to the following Lyapunov equation:
\begin{lemma}\label{lem:lyap}
Define the matrix $\Ht$ to be:
\begin{equation}
\label{eq:Ht}
\Ht \defeq \ctime\Sigma' + \ctimebar \C - \ctimebar\Sigma - \ctime\C^\top + \frac{\sdev^2}{2}(\ctimebar-\ctime)\eye.
\end{equation}Then $\lyap = \Ht^\top \Sigcurveinv$. In other words, $\Ht^\top \Sigcurveinv$ is symmetric and solves the Lyapunov equation:
\begin{equation}
  A: \quad  A\Sigcurve + \Sigcurve A= \dSigcurve.
\end{equation}
Moreover, $\Ht$ satisfies the following identity:
\begin{equation}
\label{eq:Ht-id}
\dHt - \Sigcurveinv \Ht^2 =- \frac{\sdev^4}{4} \Sigcurveinv.
\end{equation}

%
%In particular, $\Ht^\top \Sigcurve$ is symmetric.
%
%
%Let $\Ht$ be given as in \eqref{eq:Ht}. A straightforward but tedious computation (which we defer to the appendix) shows that $\Ht^\top\Sigmasolinv = \Sigmasolinv \Ht$ and $\Ht + \Ht^\top = \dSigmasol$. As a result, we have
%\begin{align*}
%\Ht^\top \Sigmasolinv \cdot \Sigmasol + \Sigmasol \cdot \Sigmasolinv\Ht = \Ht + \Ht^\top = \dSigmasol,
%\end{align*}or, in other words, $\Ht^\top \Sigmasolinv = \Sigmasolinv \Ht = \lyap$. Furthermore, since $\Sigmasol \mg 0$, $\Sigmasolinv \Ht$ is the unique solution to the Lyapunov equation $\dSigmasol = \Sigmasol X + X \Sigmasol$. In other words, we have shown
%\begin{equation}
%\lyap = \Ht^\top\Sigmasolinv = \Sigmasolinv\Ht.
%\end{equation}
\end{lemma}
%\para{Remark}It is interesting to note that the matrix $\Ht$ in \eqref{eq:Ht} is itself \emph{not} symmetric.

Before commencing the proof of \cref{lem:lyap}, let us show how it readily leads us to \eqref{eq:EL-BW}.% follows from a straightforward calculation from it. 

Recall the definition of $\coder$ in \eqref{eq:coderBW}. First, note that, by \eqref{eq:solMain} and \eqref{eq:Ht},
\begin{align}
\label{eq:ddot}
\frac{1}{2}\ddSigcurve &=   \Sigma+\Sigma' -  \parens*{  \C + \C^\top+ \sdev^2\eye }   \\
&=  \dHt.
\end{align}
On the other hand, \cref{lem:lyap} entails that
\begin{align}
\nn
\Sigcurve \parens*{\lyap}^2 + \parens*{\lyap}^2 \Sigcurve &= \Sigcurve \lyap \cdot \lyap + \lyap \cdot \lyap\Sigcurve \\
\nn
 &= \Sigcurve \Sigcurveinv \Ht \cdot \lyap + \lyap \cdot \Ht^\top \Sigcurveinv \Sigcurve \\
 \label{eq:holdBW}
 &= \Ht \lyap + \lyap \Ht^\top.
\end{align}
By noting, again from \cref{lem:lyap},
\begin{align}
\nn
\dSigcurve &= \Ht^\top \Sigcurveinv \cdot \Sigcurve + \Sigcurve \cdot \Sigcurveinv\Ht \\
\label{eq:HtHtt}
&= \Ht + \Ht^\top,
\end{align}
we thus get
\begin{align}\nn
 \Sigcurve \parens*{\lyap}^2 + \parens*{\lyap}^2 \Sigcurve- \parens*{  \lyap \dSigcurve + \dSigcurve\lyap } 
 &= \parens*{\Ht - \dSigcurve} \lyap + \lyap \parens*{\Ht^\top - \dSigcurve} \\
 \nn
 &= - \parens*{ \Ht^\top\lyap + \lyap \Ht  }
\end{align}by \eqref{eq:HtHtt}. But $\Ht^\top\lyap  = \Ht^\top \cdot \Sigcurveinv \Ht =\Sigcurveinv \Ht^2$ by symmetry of $\Ht^\top \Sigcurveinv$ and, similarly, we have $\lyap\Ht =  \Sigcurveinv \Ht^2$. As a result, \eqref{eq:coderBW} reduces to
\begin{align}
\label{eq:holdBW1}
\coder &= 2\dHt - 2 \Sigcurveinv\Ht^2.
\end{align}In lieu of \eqref{eq:EL-BW}, \eqref{eq:gradBWU}, and \eqref{eq:holdBW1}, the proof of \eqref{eq:LagBW} can thus be reduced to showing
\begin{align}
2\dHt -  2\Sigcurveinv\Ht^2 = -\frac{\sdev^4}{2} \Sigcurveinv
\end{align}which is exactly \eqref{eq:Ht-id}.%, thereby completing the proof.

%It remains to prove \cref{lem:lyap}.

\begin{proof}[Proof of \cref{lem:lyap}] 
We now prove \cref{lem:lyap}. We begin by proving some useful identities that will inspire our proof for the general \acp{GSB} in \cref{sec:results}.

\textbf{Useful identities.}
~First, note that the definition of $\C$ immediately implies $\C \Sigma = \Sigma \C^\top$. In addition, we have
\begin{align}
\nn
\C^{-1} \Sigma &= 2\parens*{ \Sigma^{\frac{1}{2}} \D  \Sigma^{-\frac{1}{2}} - \sdev^2\eye }^{-1} \Sigma \\
\nn
&= 2 \parens*{ \Sigma^{-\frac{1}{2}} \D  \Sigma^{-\frac{1}{2}} - \sdev^2\Sigma^{-1} }^{-1} \\
\label{eq:Csym}
&= \Sigma \C^{-\top}.
\end{align}
Recall from \citep{janati2020entropic} that $\C$ solves the following matrix equation:
\begin{align}
\label{eq:C}
\C^2 + \sdev^2\C =  \Sigma\Sigma'.
\end{align}
We therefore have
\begin{align*}
\C &= \C^{-1} \Sigma\Sigma' - \sdev^2\eye, \\
\C^\top &= \Sigma'\Sigma\C^{-\top} - \sdev^2\eye,
\end{align*}which, together with \eqref{eq:Csym}, implies
\begin{align}
\nn
\C^\top \Sigma' &= \Sigma'\Sigma \C^{-\top}\Sigma'-\sdev^2\Sigma'  \\
\nn
&=\Sigma' \C^{-1}\Sigma \Sigma'-\sdev^2\Sigma'   \\
\label{eq:Ctsym}
&= \Sigma' \C.
\end{align}

Now, set $\Ht = \Pt - \Qt^\top + \frac{\sdev^2}{2}(\ctimebar-\ctime)\eye$ where
\begin{align}
\Pt \defeq \ctime \Sigma' + \ctimebar \C, \quad \Qt \defeq \ctimebar \Sigma + \ctime \C.
\end{align}
Note that, by \eqref{eq:Ctsym}, 
\begin{align}
\nn
\Sigma'\Pt^{-1}  &= \parens*{ \Pt\Sigma'^{-1} }^{-1} \\
\nn
&= \parens*{\ctime\eye + \ctimebar  \C \Sigma'^{-1} }^{-1} \\
\nn
&=  \parens*{\ctime\eye + \ctimebar \Sigma'^{-1} \C^\top  }^{-1} \\
\nn
&= \parens*{ \Sigma'^{-1} \Pt^\top  }^{-1} \\
\label{eq:Ptsym}
&= \Pt^{-\top}\Sigma'.
\end{align}A similar calculation leading to \eqref{eq:Ptsym} shows
\begin{align}
\label{eq:Qtsym}
\Qt^{-1} \Sigma = \Sigma\Qt^{-\top}.
\end{align}

\textbf{Proof of symmetry of $\Ht^\top\Sigcurveinv$.}
~We get, by \eqref{eq:C} and \eqref{eq:Ctsym},

\begin{align}
\nn
\Pt^2 + \sdev^2\ctimebar\Pt &= t^2 \Sigma'^2 + \ctimebar^2\C^2 + \ctime\ctimebar\parens*{ \Sigma'\C + \C\Sigma' } + \sdev^2 \ctime\ctimebar \Sigma' + \sdev^2\ctimebar^2 \C \\
\nn
&= t^2 \Sigma'^2 + \ctimebar^2 \parens*{ \C^2 + \sdev^2\C } + \ctime\ctimebar\parens*{ \C^\top\Sigma' + \C\Sigma' } + \sdev^2 \ctime\ctimebar \Sigma'\\
\label{eq:Pteq}
&= t^2 \Sigma'^2 + \ctimebar^2 \Sigma\Sigma' + \ctime\ctimebar\parens*{ \C^\top + \C + \sdev^2\eye}\Sigma' = \Sigcurve\Sigma'.
\end{align}

It then follows from \eqref{eq:Pteq} that

\begin{align}
\label{eq:Pt2}
\Pt &=  \Sigcurve\Sigma'\Pt^{-1} - \sdev^2 \ctimebar\eye, \\
\label{eq:Ptt2}
\Pt^\top &=  \Pt^{-\top} \Sigma'\Sigcurve - \sdev^2\ctimebar\eye.
\end{align}

As a result, we get, by \eqref{eq:Ptsym} and \eqref{eq:Pt2}-\eqref{eq:Ptt2},

\begin{align}
\nn
\Sigcurveinv \Pt &= \Sigma'\Pt^{-1} - \sdev^2\ctimebar\Sigcurveinv \\
\nn
&= \Pt^{-\top}\Sigma' - \sdev^2\ctimebar\Sigcurveinv \\
\label{eq:PtSig}
&= \Pt^\top\Sigcurve^{-1}.
\end{align}

In exactly the same vein, we have

\begin{align}
\label{eq:Qteq}
\Qt^2 + \sdev^2 \ctime\Qt = \Sigma\Sigcurve
\end{align}

as well as

\begin{align}
\label{eq:QtSig}
\Sigcurveinv\Qt^\top = \Qt \Sigcurveinv.
\end{align}The symmetry of $\Ht^\top\Sigcurveinv$ is then an immediate consequence of \eqref{eq:PtSig} and \eqref{eq:QtSig}. In addition, we have
\begin{align}
\nn
\dSigcurve &= 2\ctime\Sigma' - 2 \ctimebar\Sigma + (\ctimebar-\ctime)\parens*{   \C +\C^\top  + \sdev^2\eye } \\
\label{eq:HtHtt2}
&=  \Ht + \Ht^\top.
\end{align}
Combining the symmetry of $\Ht^\top\Sigcurveinv$ and \eqref{eq:HtHtt2}, we see that
\begin{align}
\nn
\Ht^\top\Sigcurveinv \cdot \Sigcurve + \Sigcurve \cdot \Sigcurveinv \Ht = \Ht+\Ht^\top = \dSigcurve,
\end{align}\ie $\lyap = \Ht^\top\Sigcurveinv$.

\para{Proof of \eqref{eq:Ht-id}}

We next compute
\begin{align}
\nn
\Pt \Qt^\top &= \parens*{\ctime \Sigma' + \ctimebar \C}\parens*{\ctimebar \Sigma + \ctime \C^\top} \\
\nn
&= \ctime\ctimebar\Sigma'\Sigma + \ctime^2\Sigma'\C^\top + \ctimebar^2\C\Sigma + \ctime\ctimebar\C\C^\top \\
\nn
&= \ctimebar^2\Sigma \C^\top + \ctime^2\Sigma'\C^\top+ \ctime\ctimebar \parens*{ \C^{\top 2}+ \sdev^2\C^\top }+ \ctime\ctimebar  \C\C^\top
\\
\label{eq:PtQtt}
&= \Sigcurve\C^\top
\end{align}
where we have used \eqref{eq:Csym} in the third equality of \eqref{eq:PtQtt}. A similar computation further shows 
\begin{align}
\label{eq:QttPt}
\Qt^\top\Pt = \Sigcurve\C.
\end{align}

We thus get, by combining \eqref{eq:Pteq} \eqref{eq:Qteq}
\begin{align}
\nn
\Ht^2 &= \Pt^2 -\Pt \Qt^\top + \frac{\sdev^2}{2}(\ctimebar-\ctime) \Pt -\Qt^\top \Pt + \Qt^{\top2} - \frac{\sdev^2}{2}(\ctimebar-\ctime)\Qt^\top +  \frac{\sdev^2}{2}(\ctimebar-\ctime)\Pt -  \frac{\sdev^2}{2}(\ctimebar-\ctime)\Qt^\top + \frac{\sdev^4}{4}(\ctimebar-\ctime)^2\eye \\
\nn
&= \Pt^2 + \sdev^2(\ctimebar-\ctime)\Pt + \Qt^{\top2} - \sdev^2(\ctimebar-\ctime)\Qt^\top - \parens*{ \Pt\Qt^\top + \Qt^\top \Pt }  + \frac{\sdev^4}{4}(\ctimebar-\ctime)^2\eye\\
\nn
&= \Sigcurve\Sigma' - \sdev^2\ctime\Pt + \Sigcurve \Sigma - \sdev^2 \ctimebar\Qt^\top - \parens*{  \Sigcurve \C^\top + \Sigcurve \C }  + \frac{\sdev^4}{4}(\ctimebar-\ctime)^2\eye - \sdev^2 \Sigcurve + \sdev^2\Sigcurve\\
\nn
&= \Sigcurve\parens*{  \Sigma + \Sigma' - \parens*{\C + \C^\top + \sdev^2\eye }} + \sdev^2 \parens*{\Sigcurve -  \ctime\Pt - \ctimebar\Qt^\top }+ \frac{\sdev^4}{4}(\ctimebar-\ctime)^2\eye  \\
\nn 
&= \Sigcurve \dHt + \sdev^2 \cdot\ctime\ctimebar\sdev^2 \eye +  \frac{\sdev^4}{4}(\ctimebar-\ctime)^2\eye \\
\label{eq:holdBW3}
&= \Sigcurve \dHt + \frac{\sdev^4}{4}\eye
\end{align}where the third equality follows from \eqref{eq:Pteq}, \eqref{eq:Qteq}, and \eqref{eq:PtQtt}-\eqref{eq:QttPt}, and the fifth equality follows from \eqref{eq:Ht}. Multiplying both sides of \eqref{eq:holdBW3} by $\Sigcurveinv$ from the right yields the desired \eqref{eq:Ht-id}. 
%which concludes the proof of \cref{lem:lyap}.
\end{proof}

\subsubsection{Equivalence between \eqref{eq:GaussianSBWiener} and \eqref{eq:LagBW}}

We first note that, by \eqref{eq:BW-tensor} and \cref{lem:lyap},
\begin{align}
\nn
\kenergyBW &= \frac{1}{2}\tr \lyap \Sigmasol \lyap \\
\nn
&= \frac{1}{2}\tr \Ht^\top \Sigmasolinv \cdot \Sigmasol \cdot \Sigmasolinv \Ht \\
&= \frac{1}{2} \tr \Ht^\top\Sigmasolinv\Ht,
\end{align}and therefore the integrand in \eqref{eq:LagBW} is equal to
\begin{align}
\label{eq:holdBW2}
\tr \parens*{   \frac{1}{2} \Ht^\top\Sigmasolinv\Ht + \frac{\sdev^4}{8} \Sigcurveinv}.
\end{align}

To proveed, we will need another formulation of \eqref{eq:GaussianSBWiener}, which is \citep{chen2016relation, gentil2017analogy} specialized to our case: 
\begin{lemma}\label{lem:EntOT}
Let $\Ncal_0 \defeq \Ncal(0,\Sigma)$ and $\Ncal_\horizon \defeq \Ncal(0,\Sigma')$. Then \eqref{eq:GaussianSBWiener} is equivalent to
\begin{align}
\label{eq:LagMeasure}
\min_{\ssstyle\substack{{\measure[0] = \Ninit, \measure[\horizon] = \Nend}}} \int_{0}^\horizon  \exof*{\frac{1}{2} \norm{\vecsol}^2 + \frac{\sdev^4}{8} \norm{ \nabla \log \measure }^2} \dt
\end{align}where the minimization is taken over all pairs $(\measure, \vecsol)$ such that $\Phi_\ctime:\R^\vdim\to\R$ are differentiable functions and the continuity equation holds:
\begin{equation}
\label{eq:ConeqApp}
\pt\measure = - \Div(\measure \vecsol).
\end{equation}
\end{lemma}
%The equivalence between \eqref{eq:LagMeasure} and \eqref{eq:GaussianSB} is well-known \citep{chen2016relation,gentil2017analogy}.

We will also need the \emph{Jacobi formula}: Let $A(\ctime) \from \R^+ \to \R^{\vdim\times\vdim} $ be a differentiable matrix-valued function. Then
\begin{align}
\label{eq:jacobi}
\ddt \det A(\ctime) = \det A(\ctime) \cdot \tr A^{-1}(\ctime) \cdot \ddt A(\ctime).
\end{align}

We are now ready to finish the proof of \cref{thm:mechanics}. By \citet{leonard2013survey}, the optimal curve for \eqref{eq:LagMeasure} is Gaussian with zero mean. We denote by $\Sigmasol$ the covariance of the solution at time $\ctime$. By \eqref{eq:jacobi}, we have
\begin{align}
\nn
\pt\measure(\point) &= \pt \parens*{ \nconst (\det\Sigmasol)^{-\frac{1}{2}} \exp\parens*{ -\frac{1}{2}\point^\top \Sigmasolinv\point   }}  \\
\nn
&=   \nconst \parens*{-\frac{1}{2} (\det\Sigmasol)^{-\frac{3}{2}}}  
\cdot\det\Sigmasol\cdot \tr \Sigmasolinv \dSigmasol \exp\parens*{ -\frac{1}{2}\point^\top \Sigmasolinv\point   } \\
\nn
&\hspace{20mm}+ \nconst (\det\Sigmasol)^{-\frac{1}{2}} \exp\parens*{ -\frac{1}{2}\point^\top \Sigmasolinv\point   } \cdot \parens*{\frac{1}{2} \point^\top \Sigmasolinv \dSigmasol \Sigmasolinv \point} \\
\label{eq:pt-m}
&= \measure(\point) \cdot \parens*{ \frac{1}{2} \point^\top \Sigmasolinv \dSigmasol \Sigmasolinv \point -\frac{1}{2}\tr \Sigmasolinv \dSigmasol }.
\end{align}

On the other hand, by the chain rule for the divergence, we have
\begin{align}
\label{eq:div-mv}
\Div(\measure \vecsol)  &=   \inner{\nabla\measure}{\vecsol}  + \measure \Laplace\potsol.
%&= \measure \parens*{ -\Sigmasolinv\point }.
\end{align}
Since $\nabla \measure = \measure \parens*{ -\Sigmasolinv\point }$, the continuity equation \eqref{eq:ConeqApp} together with \eqref{eq:pt-m}-\eqref{eq:div-mv} implies that $\Sigmasol$ must satisfy
\begin{align}
\Laplace\potsol &= \frac{1}{2} \tr \Sigmasolinv \dSigmasol, \\
\inner{\Sigmasolinv\point}{\vecsol(\point)}  &= \frac{1}{2} \inner{\Sigmasolinv\point}{\dSigmasol\Sigmasolinv\point}, \quad \forall \point \in \R^\vdim.
\end{align}
In other words, the optimal vector field is of the form $\vecsol(\point) = \Ht^\top\Sigmasolinv\point$ for some matrix $\Ht$ such that 
\begin{align}
\tr\Ht^\top \Sigmasolinv &= \frac{1}{2}\tr\dSigmasol\Sigmasolinv, \\
\tr \Sigmasolinv \Ht^\top\Sigmasolinv \point\point^\top &=\frac{1}{2} \tr \Sigmasolinv \dSigmasol \Sigmasolinv \point\point^\top, \quad \forall \point\in\R^\vdim.
\end{align}
Therefore, we see that
\begin{align}
\nn
\exof*{\norm{\vecsol}^2} &= \exof*{\tr \Ht^\top\Sigmasolinv\point \point^\top\Sigmasolinv \Ht} \\
\nn
&= \tr \Ht^\top\Sigmasolinv \exof{\point\point^\top} \Sigmasolinv \Ht \\
&= \tr \Ht^\top\Sigmasolinv \Ht.
\end{align}
Furthermore, we have
\begin{align}
\nn
\exof*{ \norm{\nabla \log \measure}^2} &= \exof*{\tr \Sigmasolinv \point\point^\top \Sigmasolinv } \\
&= \tr \Sigmasolinv.
\end{align}
Finally, since the optimal vector field $\vecsol$ is a gradient field, we must have $\Ht^\top \Sigmasolinv = \Sigmasolinv\Ht$. Combing all the above, we see that \eqref{eq:LagMeasure} is equivalent to
\begin{align}
\label{eq:LagMeasureEquiv}
\min_{\ssstyle\substack{{\Sigcurve[0] = \Sigma, \Sigcurve[\horizon] = \Sigma'} \\ { \Ht^\top\Sigcurveinv =  \Sigcurveinv\Ht }}} \int_{0}^\horizon  \tr\ \parens*{\frac{1}{2}\Ht^\top\Sigcurveinv \Ht  + \frac{\sdev^4}{8} \Sigcurveinv} \dt
\end{align}which, in view of \eqref{eq:holdBW2}, is exactly the same as \eqref{eq:LagBW}.

\subsection{Some Interesting Consequences of \cref{thm:mechanics}}
\label{sec:interesting}

Here, we collect some interesting corollaries of \cref{thm:mechanics}, although they will not be used in the rest of the paper.

\subsubsection{Conservation of Hamiltonian}
The first result concerns the \emph{Hamiltonian formulation} of the action minimization problem \eqref{eq:LagBW}. 
 
\NewDocumentCommand{\codiv}{ O{\dSigmasol} O{\dSigmasol} }{ \nabla_{\ssstyle #1}{#2} }
\newmacro{\hambase}{ \mathcal{H} }
\newcommand{\ham}[1][\Sigmasol]{ \hambase\parens*{#1} }

\begin{corollary}[Conservation of Hamiltonian]
%\begin{enumerate}
%\item $\codiv = - \gradBW \penergyBW = -\frac{\sdev^4}{2} \Sigmasolinv$.
Define the \textbf{Hamiltonian} associated with \eqref{eq:LagBW} to be
\begin{align}
\nn
\ham &\defeq \kenergyBW + \penergyBW \\
\label{eq:HamBW}
\tag{H}
&=  \tr  \parens*{\frac{1}{2} \Ht^\top\Sigmasolinv\Ht - \frac{\sdev^4}{8} \Sigmasolinv}.
\end{align}Then the Hamiltonian is conserved along $\Sigmasol$: 
\begin{align}
\dot{\hambase} \equiv 0,  \text{ or, equivalently, }  \ham = \tr \parens*{\Sigma + \Sigma' - \D} \textup{ for all $\ctime$.}
\end{align}
%\end{enumerate}
\end{corollary}

The fact that the Hamiltonian, commonly interpreted as the \emph{total energy}, is conserved is a well-known fact in physics \citep{villani2009optimal} and directly follows from \cref{thm:mechanics}.

\subsubsection{Connection to Fisher Information}

The ``potential energy'' term $\penergyBW$ in \eqref{eq:LagBW} has an interesting origin: It is, up to a constant, the \emph{entropy production rate}, \ie the \emph{Fisher information}.

\begin{lemma}\label{lem:Fisher}
Let $\measurebase \sim \Ncal(0,\covbase)$, and let $\entropy$ be the (negative) Shannon entropy of $\measurebase$. Then 
%\begin{equation}
%\frac{1}{2}\fisherBW = \frac{\sdev^4}{8} \exof*{\norm{ \nabla \log \measurebase }^2} =  \frac{\sdev^4}{8} \tr \covbase^{-1}.
%\end{equation}
\begin{equation}
\label{eq:BW-U-Fisher}
\penergyBW = \frac{1}{2}\fisherBW %\frac{\sdev^4}{8} \tr \covbase^{-1} =  \frac{\sdev^4}{8} \tr \covbase^{-1}.
\end{equation}where 
\begin{equation}
\label{eq:BW-Fisher}
\fisherBW \defeq \frac{\sdev^4}{4} \normBWsq[\gradBW \entropy][\covbase].
\end{equation}
\end{lemma}
\begin{proof}
%A straightforward computation shows that $\exof*{\norm{ \nabla \log \measurebase }^2} = \covbase^{-1}$. On the other hand, 
Recall that $\nabla \entropy= \nabla\parens*{ -\frac{1}{2} \log\det \covbase - \frac{\vdim}{2} \log 2\pi e }= -\frac{1}{2}  \covbase^{-1}$. Therefore, by \eqref{eq:BW-tensor} and \eqref{eq:gradBW-lyapinv},
\begin{align*}
\fisherBW &= \frac{\sdev^4}{4} \normBWsq[\gradBW \entropy][\covbase] \\
&= \frac{\sdev^4}{4}\innerBW[\gradBW \entropy][\gradBW \entropy][\covbase] \\
&= \frac{\sdev^4}{4} \tr \lyap[\covbase][\gradBW \entropy]\covbase\lyap[\covbase][\gradBW \entropy] \\
&= \frac{\sdev^4}{4} \tr \lyap[\covbase][\lyapinv[\covbase][2\nabla \entropy]]\covbase\lyap[\covbase][\lyapinv[\covbase][2\nabla \entropy]] \\
&= \frac{\sdev^4}{4} (-\covbase^{-1})\covbase(-\covbase^{-1})=\frac{\sdev^4}{4}\covbase^{-1}. \qedhere
\end{align*}
\end{proof}

An infinite-dimensional version of \cref{lem:Fisher} for non-Gaussian measures is proved in \citet{chen2016relation, gentil2017analogy}; the connection to the Bures-Wasserstein geometry here seems to be new.

The specific form of the potential energy in \eqref{eq:BW-Fisher} has been shown to be intimately related to the \emph{gradient flow} of entropy:
\begin{align}
\dSigcurve = - \gradBW \entropy.
\end{align}

We refer the interested readers to \citep{gentil2020dynamical} for details.

\subsubsection{Solution of the Schr\"odinger Systems}

Another way of solving a system of the form \eqref{eq:LagMeasure} is via the so-called forward \emph{Schr\"odinger system} \citep{chen2021stochastic, leonard2013survey}:
\begin{align}
\label{eq:SchSys}
\left\{\begin{array}{lr}
\del_t \mu_t +\Div\parens*{  \mu_t\nabla \pot } = \frac{\sdev^2}{2}  \Laplace\mu_t \\
       \del_t \pot + \frac{\norm{ \nabla\pot}^2}{2} + \frac{\sdev^2}{2} \Laplace \pot = 0
        \end{array}\right..
\end{align}

By the various identities we prove in \cref{sec:verifyEL}, one can easily show that the solution to \eqref{eq:SchSys} is given by
\begin{align}
\pot(\point) = - \frac{\sdev^2}{4} \log\det\Sigmasol + \frac{\sdev^4}{4}  \int_{0}^t \tr \Sigmasolinv \dt + \frac{1}{2} \inner{\point}{\parens*{\Ht^\top- \frac{\sdev^2}{2}\eye}\Sigmasolinv \point} + \const
\end{align}

This is in fact the same solution of the \emph{fluid mechanical} problem
\begin{align}
\min_{\ssstyle\substack{{\measure[0] = \Ninit, \measure[\horizon] = \Nend} \\ {\pt\measure +\Div(\measure \vecsol) =\Laplace \measure }}} \int_{0}^\horizon  \exof*{\frac{1}{2} \norm{\vecsol}^2 } \dt
\end{align}which is yet another equivalent formulation of \eqref{eq:GaussianSBWiener}.

There is also a backward Schr\"odinger system:
\begin{align}
\left\{\begin{array}{lr}
-\del_t \mu_t + \Div\parens*{  \mu_t\nabla \rpot } =\frac{\sdev^2}{2}  \Laplace\mu_t  \\
      - \del_t \rpot + \frac{\norm{ \nabla\rpot}^2}{2} +\frac{\sdev^2}{2} \Laplace \rpot = 0
        \end{array}\right.,
\end{align}
whose solution is given by
\begin{align}
\rpot(\point) = - \frac{\sdev^2}{4} \log\det\Sigmasol - \frac{\sdev^4}{4}  \int_{0}^t \tr \Sigmasolinv \dt - \frac{1}{2} \inner{\point}{ \parens*{\Ht^\top+\frac{\sdev^2}{2}\eye}\Sigmasolinv \point} + \const
\end{align}

Notice that
\begin{align}
\pot + \rpot = \sdev^2 \log \measure
\end{align}which is a well-known feature of the solutions to the forward and backward Schr\"odinger systems \citep{chen2021stochastic,leonard2013survey}.

\section{Proof of the Closed-Form Solutions for Gaussian Schr\"odinger Bridges}
\label{app:GaussianSB}

\subsection{Preliminaries for the Proof of \cref{thm:GaussianSB}}
\label{app:PrelimproofGSB}

We need a technical lemma that is intimately related to the ``central identity of quantum field theory'' \citep{zee2010quantum}; the version below is adopted from \citep{126767}, wherein the readers can find an easy proof.
\begin{lemma}[The central identity of Quantum Field Theory]
\label[lemma]{lem:centralid}
The following identity holds for all matrix $\mat \mg 0$ and all sufficiently regular analytic function $v$ (\eg polynomials or $v\in \mathcal{C}^\infty(\RR^\vdim)$ with compact support):

\begin{equation}
\label{eq:QFT-iden}
(2\pi)^{-\frac{d}{2}} (\det \mat)^{\frac{1}{2}} \intR v(\point) \exp\parens*{ -\frac{1}{2} \point^\top \mat \point } \dd \point = \left.\exp\parens*{ \frac{1}{2} \del_\point^\top \mat^{-1} \del_\point}  v(\point)\right\vert_{\point = 0}
\end{equation}

where $\exp\parens*{ \frac{1}{2} \del_\point^\top \mat^{-1} \del_\point}$ is understood as a power series in the differential operators.
\end{lemma}
Lastly, we recall the elementary
\begin{lemma}[Conditional Gaussians are Gaussian]
\label[lemma]{lem:Cond-Gaussians}
Let $(Y_0, Y_1) \sim \Ncal \parens*{  \begin{bmatrix}
\mu_0 \\
\mu_1
\end{bmatrix},  \begin{bmatrix}
\Sigma_{00}& \Sigma_{01}\\
\Sigma_{10}&\Sigma_{11}
\end{bmatrix}}$. Then $Y_0 \vert Y_1 = y \sim \Ncal( \check{\mu}, \check{\Sigma})$ where
\begin{align}
\nn
\check{\mu} &= \mu_0 + \Sigma_{01}\Sigma_{11}^{-1}(y- \mu_1), \\
\check{\Sigma} &= \Sigma_{00} - \Sigma_{01}\Sigma^{-1}_{11}\Sigma_{10}.
\end{align}
\end{lemma}

\subsection{The Proof}
\label{app:proofGSB}

We are now ready for the proof. For convenience, we restate \cref{thm:GaussianSB} below:
\GaussianSB*

As the proof is quite complicated, we first outline the main steps below:
\begin{enumerate}[leftmargin=.5cm,itemsep=.01cm,topsep=0cm]
\item Leveraging existing results \citep{bojilov2016matching, del2020statistical, janati2020entropic,mallasto2021entropy}, we first solve an appropriately chosen \emph{static} \ac{GSB} determined by the reference process $\refpro$.% with $\sdev^2 \subs \horizon\cdot\scalingsq[\horizon]$ seen as the ``effective elapsed time''.
\item It can be shown from the disintegration formula \citep{leonard2013survey}, the solution of the static \acp{GSB} \eqref{eq:StaticGaussianSB-sol}, and properties of \eqref{eq:linearsdesol} that $\Psol$ is a Markov Gaussian process with mean \eqref{eq:meansol} and covariance \eqref{eq:covsol}.

\item Invoking the \emph{generator theory} \citep{protter2005stochastic}, to prove \eqref{eq:GSB-forward-sol}, it suffices to show that $\Xsol$ satisfies, for any sufficiently regular test function $\testfbase : \RR^+ \times \RR^\vdim \rightarrow \RR$,

\begin{align}
\label{eq:semi-gen}
\lim_{h\to 0} \frac{ \exof*{ \testf[\ctime+h][\Xsol[\ctime+h]] \given \Xsol = \point } }{h}
    = \generator \testf,
\end{align}
where 
\begin{align}
\label{eq:gen-def}
\generator \testf  &\defeq \frac{\del}{\del \ctime} \testf + \frac{ \volatsq[\ctime] }{2} \Laplace\testf  + \inner*{\nabla \testf}{ \GSBf } 
\end{align}
is the generator for the process \eqref{eq:GSB-forward-sol}.

\item Since the marginal/joint/conditional distributions of a Gaussian process are still Gaussian, the expectation in \eqref{eq:semi-gen} requires to express Gaussian integrals as differential operators. To this end, the appropriate tool is the ``central identity in quantum field theory'' \citep{zee2010quantum}.%; see \cref{lem:centralid}.

\item Proof concludes by matching terms in \eqref{eq:semi-gen} and \eqref{eq:gen-def}.\qedhere
\end{enumerate}

\begin{proof}[Proof of \cref{thm:GaussianSB}]
\textbf{From now on, we will invoke the notations in \eqref{eq:functions} without explicit mentions.}

\textbf{The static Gaussian \ac{SB}.}
~We begin by solving the \emph{static} Gaussian \ac{SB}

\begin{align}
\label{eq:staticGSB}
\min_{ \Psta}\KL{\Psta}{\Qsta} 
\end{align}

over all $\Psta$ having marginals $\Ncal\parens*{  \m, \covar }$ and $\Ncal\parens*{  \malt, \covaralt }$.

Recall that, conditioned on $\refsde[0]$, $\refsde \sim \refpro$ is a Gaussian process with mean \eqref{eq:mYdef} and covariance \eqref{eq:covYdef}. Thus, if we only consider the endpoint marginal distributions $(\refsde[0], \refsde[\horizon])$, it is easy to derive the transition probability:

\begin{align}
\refprobase\parens*{   \refsde[\horizon] = y_\horizon \middle| \refsde[0] = y_0 } &= \Nconst  \det \parens*{\kernel[\horizon][\horizon] \eye }^{-\frac{1}{2}} \exp\parens*{  -\frac{1}{2}  \parens*{y_\horizon - \mYcinit[\horizon]}^\top  \parens*{\kernel[\horizon][\horizon] \eye }^{-1}\parens*{y_\horizon - \mYcinit[\horizon]} } \\
&= \Nconst  \det \parens*{\kernel[\horizon][\horizon] \eye }^{-\frac{1}{2}} \exp\parens*{  -\frac{1}{2\kernel[\horizon][\horizon]}  \norm*{  y_\horizon - \aggtime[\horizon] y_0  - \tshift[\horizon]   }^2}.
\end{align}

Therefore, abusing the notation by continually writing $\Psta$ as the relative density of $\Psta$ with respect to the Lebesgue measure, we get

\begin{align}
\label{eq:KLsta}
\KL{\Psta}{\Qsta} &= \int_{\R^\vdim \times \R^\vdim}  \log \frac{\dd \Psta } {\dd \Qsta} \dd \Psta \\
\label{eq:hold}
&=   \const +  \frac{1}{2 \kernel[\horizon][\horizon]} \int_{\R^\vdim \times \R^\vdim}  \norm*{  y' - \aggtime[\horizon] y   - \aggtime[\horizon]\tshift[\horizon]    }^2  \dd \Psta(y,y') +  \intRR \log  \Psta  \dd \Psta.
\end{align}

If $\Psta$ is a joint distribution with marginals $Y \sim \Ncal\parens*{ \m, \covar}$ and $Y'\sim\Ncal\parens*{ \malt, \covaralt}$, then the change of variable $\tilde{Y} = \aggtime[\horizon]Y + \tshift[\horizon]$ gives rise to a joint distribution $\tPsta$ having marginals $\tilde{Y} \sim \Ncal\parens*{ \tm, \tcovar}$ and $Y'\sim\Ncal\parens*{ \malt, \covaralt}$, where

\begin{align}
\tm &= \aggtime[\horizon]\m + \tshift[\horizon], \\
\tcovar &= \aggtimesq[\horizon] \covar.
\end{align}

Obviously, there is a one-to-one correspondence between $\Psta$ and $\tPsta$.

The first integral in \eqref{eq:hold} is equal to $ \exof*{  \norm*{  Y'- \tilde{Y} }^2 }$. On the other hand, we always have 

\begin{equation}
\nn
\intRR \log  \tPsta  \dd \tPsta =\intRR \log  \Psta  \dd \Psta +\const
\end{equation}

Therefore, minimizing \eqref{eq:KLsta} over $\Psta$ is equivalent to

\begin{align}
\label{eq:tPstasol}
\min_{ \tPsta} \KL{ \tPsta }{ \Qsta}\equiv \min_{ \tPsta}\; \intRR  \frac{ \norm*{ y - y'}_2^2}{2} \dd \tPsta(y, y') +  \kernel[\horizon][\horizon]\intRR \log  \tPsta  \dd \tPsta.
\end{align}

By \eqref{eq:StaticGaussianSB-sol}, the solution to \eqref{eq:tPstasol} is given by the joint Gaussian 

\begin{align}
\tPstasol \sim \Ncal \parens*{  \begin{bmatrix}
\tm \\
\malt
\end{bmatrix},  \begin{bmatrix}
\tcovar & \tilde{C}_{\tilde{\sdev}    } \\
\tilde{C}_{ \tilde{\sdev}}^\top & \covaralt
\end{bmatrix}}
\end{align}

where $\tilde{\sdev} = \sqrt{ \kernel[\horizon][\horizon]}$ and

\begin{align}
\label{eq:hold1}
\tilde{C}_{\tilde{\sdev}} &=  \frac{1}{2}\parens*{\tcovar^{\frac{1}{2}} \tilde{D}_{\tilde{\sdev}} \tcovar^{-\frac{1}{2}} - \tilde{\sdev}^2\eye}, \\
\label{eq:hold2}
\tilde{D}_{\tilde{\sdev}} &= \parens*{ 4\tcovar^{\frac{1}{2}} \covaralt \tcovar^{\frac{1}{2}} +  \tilde{\sdev}^4\eye  }^{\frac{1}{2}}.
\end{align}
The optimal static Gaussian \ac{SB} $\Pstasol$ is then given by the inverse transform $Y = \aggtimeinv[\horizon] \parens*{\tilde{Y} - \tshift[\horizon]}$, \ie
\begin{align}
\label{eq:Pstasol}
\Pstasol \sim \Ncal \parens*{  \begin{bmatrix}
\m \\
\malt
\end{bmatrix},  \begin{bmatrix}
\covar &  \aggtimeinv[\horizon] \tilde{C}_{\tilde{\sdev}    } \\
\aggtimeinv[\horizon]\tilde{C}_{ \tilde{\sdev}}^\top & \covaralt
\end{bmatrix}}.
\end{align}Rearranging terms and using \eqref{eq:hold1} and \eqref{eq:hold2}, we get 
\begin{align}
\aggtimeinv[\horizon] \tilde{C}_{\tilde{\sdev}} = C_{\sdev_\star}
\end{align}where $\sdev_\star = \frac{\kernel[\horizon][\horizon]}{ \aggtime[\horizon] }$.

\para{The $\refprobase$\textendash bridges}

For future use, we will need the distribution of $\refsde$ conditioned on $\refsde[0]$ and $\refsde[\horizon]$. When $\refsde \equiv \Wiener$, the distribution is called the \emph{Brownian bridge}, which is in itself an important subject in mathematics and financial engineering \citep{mansuy2008aspects}. We thus term the conditional distribution of $\refsde$ the \emph{$\refprobase$\textendash Bridges}.

From \eqref{eq:mYdef} and \eqref{eq:covYdef}, one can infer that, given $\refsde[0]$, the joint distribution of $\parens*{\refsde,\refsde[\horizon]}$ is 

\begin{align}
\refsde,\refsde[\horizon] \vert \refsde[0] \sim  \Ncal \parens*{  \begin{bmatrix}
\mYcinit[\ctime] \\
\mYcinit[\horizon]
\end{bmatrix},  \begin{bmatrix}
\kernel[\ctime][\ctime] \eye &  \kernel[\ctime][\horizon]\eye \\
\kernel[\ctime][\horizon]\eye & \kernel[\horizon][\horizon]\eye
\end{bmatrix} }.
\end{align}

Therefore, \cref{lem:Cond-Gaussians} applied implies that, conditioned on $\refsde[0]$ and $\refsde[\horizon]$, $\refsde$ is Gaussian with mean

\begin{align}
\nn
\exof{  \refsde \given \refsde[0], \refsde[\horizon] } &= \mYcinit + \frac{\kernel[\ctime][\horizon]}{ \kernel[\horizon][\horizon] }\parens*{ \refsde[\horizon] - \mYcinit[\horizon] } \\
\nn
&= \aggtime \refsde[0] + \tshift + \frac{\kernel[\ctime][\horizon]}{ \kernel[\horizon][\horizon] }\parens*{ \refsde[\horizon] - \aggtime[\horizon]\refsde[0] - \tshift[\horizon] } \\
\nn
&= \parens*{  \aggtime -  \frac{\kernel[\ctime][\horizon]}{ \kernel[\horizon][\horizon] } \aggtime[\horizon] } \refsde[0] +  \frac{\kernel[\ctime][\horizon]}{ \kernel[\horizon][\horizon] } \refsde[\horizon] + \tshift -  \frac{\kernel[\ctime][\horizon]}{ \kernel[\horizon][\horizon] } \tshift[\horizon]  \\
\label{eq:QBridgem}
&= \ratioc \refsde[0] + \ratio \refsde[\horizon] + \tshift - \aggtime \tshift[\horizon]
\end{align}

and covariance process (for any $\ctime' \geq \ctime$)

\begin{align}
\label{eq:QBridgecov}
\exof*{ \parens*{ \refsde - \exof{  \refsde \given \refsde[0], \refsde[\horizon] } }\parens*{ \refsde[\ctime'] - \exof{  \refsde[\ctime'] \given \refsde[0], \refsde[\horizon] } }^\top \given \refsde[0], \refsde[\horizon]  } 
&= \parens*{   \kernel - \frac{\kernel[\ctime][\horizon] \kernel[\ctime'][\horizon]}{ \kernel[\horizon][\horizon]} } \eye.
\end{align}

Since a Gaussian process is uniquely determined by its mean and covariance processes, we have, for some Gaussian process $\xi_t$ independent of $\refsde$ having zero mean and covariance process \eqref{eq:QBridgecov},

\begin{equation}
\label{eq:QBridge}
\refsde \vert \refsde[0], \refsde[\horizon] \eqlaw \ratioc \refsde[0] + \ratio \refsde[\horizon] + \tshift - \aggtime \tshift[\horizon] + \xi_t.
\end{equation}

\para{From $\refprobase$\textendash bridges to $\meansol$ and $\Sigmasol$}
The disintegration formula of $\KL{\cdot}{\cdot}$ \citep{leonard2013survey} implies that the solution to \eqref{eq:GSB} is given by first generating $(\Xsol[0], \Xsol[\horizon]) \sim \Pstasol$ for $\Pstasol$ in \eqref{eq:Pstasol}, and then connecting $\Xsol[0]$ and $\Xsol[\horizon]$ using the $\refprobase$\textendash bridges \eqref{eq:QBridge}. Namely,

\begin{align}
\label{eq:Xsol}
\Xsol \eqlaw \ratioc \Xsol[0] + \ratio \Xsol[\horizon] + \tshift - \ratio \tshift + \xi_\ctime
\end{align}

from which \eqref{eq:meansol} and \eqref{eq:covsol} follow by a straightforward calculation. Furthermore, in view of \eqref{eq:Pstasol} and \eqref{eq:Xsol}, $\Xsol$ is obviously a Gaussian process. Finally, since $\refpro$ is a Markov process, \cite[Theorem 2.12]{leonard2013survey} implies that $\Psol$ is also Markov. This concludes the first half of \cref{thm:GaussianSB}.

\para{The \ac{SDE} representation of $\Xsol$}
The main idea of proving \eqref{eq:GSB-forward} is to compute

\begin{equation}
\label{eq:generatorsol}
\lim_{ h\to 0}  \frac{  \exof*{ \testf[\ctime+h][\Xsol[\ctime+h]] \given \Xsol = x }  - \testf  }{h}
\end{equation}

and equate \eqref{eq:generatorsol} with the \emph{generator} of \eqref{eq:GSB-forward-sol}, which is \citep{protter2005stochastic}

\begin{equation}
\label{eq:gensde}
\generator \testf \defeq  \frac{\del}{\del \ctime} \testf +  \frac{\volatsq[\ctime]}{2} \Laplace \testf +  \inner*{  \nabla \testf }{ \GSBf }.
\end{equation}

Since $\Xsol$ is a Gaussian process, we may derive the conditional expectation in \eqref{eq:generatorsol} using \cref{lem:Cond-Gaussians}. However, since eventually we will divide everything by $h$ and drive $h\to 0$, we can ignore any term that is $o(h)$ during the computation. This simple observation will prove to be extremely useful in the sequel.

We first compute the first-order approximation of $\Sigmasol$. In view of \eqref{eq:covsol}, and since $\ratio \kernel[\ctime][\horizon] = \kernel[\ctime][\ctime]\efftr$ and $\dratio\kernel[\ctime][\horizon]=\ratio \frac{\del}{\del \ctime} \kernel[\ctime][\horizon]$, we have

\begin{align}
\nn
\dSigmasol &= 2 \dratioc\ratioc \covar + 2 \dratio\ratio \covaralt + \parens*{  \dratio\ratioc + \ratio\dratioc } \parens*{ \Cs + \Cs^\top} + \parens*{  \frac{\del}{\del \ctime} \kernel[\ctime][\ctime] - \dratio \kernel[\ctime][\horizon] - \ratio \frac{\del}{\del \ctime} \kernel[\ctime][\horizon]  }\eye \\
\nn
&= \dratio \parens*{   \ratio \covaralt + \ratioc \Cs + \ratio \covaralt + \ratioc \Cs^\top  } + \dratioc \parens*{   \ratioc \covar + \ratio \Cs + \ratioc \covar + \ratio \Cs^\top  } + \parens*{  \frac{\del}{\del \ctime} \kernel[\ctime][\ctime] -2 \dratio \kernel[\ctime][\horizon]  }\eye\\
\label{eq:dSigmasol}
&= \parens*{ \Pt + \Pt^\top } - \parens*{ \Qt + \Qt^\top } + \parens*{  \frac{\del}{\del \ctime} \kernel[\ctime][\ctime] -2 \dratio \kernel[\ctime][\horizon]  }\eye.
\end{align}

Next, let $K_{\ctime,\ctime + h}$ denote the covariance process of $\Xsol$. We can estimate $K_{\ctime,\ctime + h}$ up to first order by computing:
\begin{align}
\nn
K_{\ctime, \ctime + h} &\defeq \exof*{  \parens*{\Xsol-\meansol} \parens*{\Xsol[\ctime+h]-\meansol[\ctime+h]}^\top   } \\
\nn
&=  \ratioc\ratioc[\ctime+h] \covar + \ratio \ratio[\ctime+h] \covaralt + \ratioc \ratio[\ctime+h] \Cs + \ratio\ratioc[\ctime+h]\Cs^\top 
+ \parens*{ \kernel[\ctime][\ctime+h] - \ratio[\ctime+h] \kernel[\ctime][\horizon] }\eye \\
\nn
&= \Sigmasol + \ratioc(\ratioc[\ctime+h]-\ratioc) \covar + \ratio(\ratio[\ctime+h] - \ratio) \covaralt + \ratioc(\ratio[\ctime+h] - \ratio) \Cs + \ratio( \ratioc[\ctime+h] -\ratioc) \Cs^\top \\
\nn
&\hspace{70mm}+ \parens*{ \kernel[\ctime][\ctime+h] - \kernel[\ctime][\ctime] - \ratio[\ctime+h] \kernel[\ctime][\horizon] + \ratio\kernel[\ctime][\horizon] }\eye \\
\nn
&= \Sigmasol + \frac{\ratio[\ctime+h] - \ratio}{\dratio} \Pt - \frac{\ratioc[\ctime+h] - \ratioc}{\dratioc} \Qt^\top + \parens*{ \kernel[\ctime][\ctime+h] - \kernel[\ctime][\ctime] - \ratio[\ctime+h] \kernel[\ctime][\horizon] + \ratio\kernel[\ctime][\horizon] }\eye \\
\label{eq:hold5}
&= \Sigmasol + h \braces*{  \Pt - \Qt^\top +  \bracks*{ \parens*{\frac{\del}{\del \ctime'}\kernelbase }(\ctime, \ctime) - \dratio \kernel[\ctime][\horizon] } \eye } + o(h),
\end{align}

where $\parens*{\frac{\del}{\del \ctime'}\kernelbase }(\ctime,\ctime') \defeq \lim_{h\to 0} \frac{ \kernel[\ctime][\ctime'+h] -\kernel[\ctime][\ctime']}{h} $ denotes the derivative of the function $\kernel[\ctime][\cdot]$. Using \eqref{eq:covYdef} and $\daggtime=\drift \aggtime$, we have

\begin{align}
\parens*{\frac{\del}{\del \ctime'}\kernelbase }(\ctime, \ctime) &= \frac{\del}{\del \ctime'} \parens*{  \aggtime[\ctime]\aggtime[\ctime'] \intdasq }\Bigg\vert_{\ctime' = \ctime} \\
\nn
&= \daggtime\aggtime\intdasq \\
\label{eq:6}
&= \drift \kernel[\ctime][\ctime].
\end{align} 

On the other hand, we have

\begin{align}
\nn
\dratio &= \frac{1}{\kernel[\horizon][\horizon]} \frac{\del}{\del \ctime} \parens*{  \aggtime \aggtime[\horizon] \intdasq  } \\
\nn
&= \frac{1}{\kernel[\horizon][\horizon]} \parens*{  \drift \kernel[\ctime][\horizon] + \aggtimeinv[\ctime] \aggtime[\horizon]  \volatsq[\ctime] } \\
\label{eq:7}
&= \drift \ratio + \frac{ \aggtime[\horizon] \volatsq[\ctime]}{ \aggtime \kernel[\horizon][\horizon] }. 
\end{align}

Combining \eqref{eq:6} and \eqref{eq:7}, using the fact that $\ratio \kernel[\ctime][\horizon] = \kernel[\ctime][\ctime]\efftr$ and $\frac{ \aggtime[\horizon] \kernel[\ctime][\horizon] }{ \aggtime \kernel[\horizon][\horizon] } = \efftr$, we may further write \eqref{eq:hold5} as

\begin{align}
\nn
K_{\ctime, \ctime + h} &= \Sigmasol + h \braces*{  \Pt - \Qt^\top + \bracks*{\drift\kernel[\ctime][\ctime] \parens*{1- \efftr} - \volatsq[\ctime] \efftr} \eye } + o(h)  \\
\label{eq:Kapprox}
&= \Sigmasol + h \St + o(h).
\end{align}

%\begin{align}
%\parens*{\frac{\del}{\del \ctime'}\kernelbase }(\ctime, \ctime) - \dratio \kernel[\ctime][\horizon] &= \drift\parens*{  \kernel[\ctime][\ctime] - \frac{\kernel[\ctime][\horizon]}{ \kernel[\horizon][\horizon] } } - 
%\end{align}

We are now ready to derive \eqref{eq:GSB-forward-sol}. By \cref{lem:Cond-Gaussians}, the random variable $\Xsol[\ctime+h]$ conditioned on $\Xsol = \point$ follows $\Ncal\parens*{  \muc, \Sigmac }$ where, by \eqref{eq:Kapprox},

\begin{align}
\nn
\muc &= \meansol[\ctime+h] + \Kth^\top \Sigmasolinv\parens*{ \point - \meansol  } \\
\nn
&= \meansol + h \dmeansol + \parens*{ \eye + h \St^\top \Sigmasolinv } (\point - \meansol) + o(h) \hspace{20mm} \\
\label{eq:mucapprox}
&= \point + h \parens*{  \St^\top \Sigmasolinv\parens*{ \point - \meansol  } + \dmeansol } + o(h),
\end{align}

and, by \eqref{eq:dSigmasol} and \eqref{eq:hold5},

\begin{align}
\nn
\Sigmac &= \Sigmasol[\ctime+h] - \Kth^\top \Sigmasolinv \Kth \\
\nn
&= \Sigmasol + h\dSigmasol - \parens*{   \Sigmasol + h \St^\top + h\St } + o(h) \\
\nn
&= h \Bigg[  \Pt + \Pt^\top - \Qt - \Qt^\top +  \parens*{  \frac{\del}{\del \ctime} \kernel[\ctime][\ctime] -2 \dratio \kernel[\ctime][\horizon]  }\eye  \\
\nn
&\hspace{15mm}- \parens*{ \Pt - \Qt^\top + \bracks*{\parens*{ \frac{\del}{\del \ctime'}\kernelbase }(\ctime, \ctime) - \dratio \kernel[\ctime][\horizon] } \eye  }^\top - \parens*{ \Pt - \Qt^\top + \bracks*{\parens*{\frac{\del}{\del \ctime'}\kernelbase }(\ctime, \ctime) - \dratio \kernel[\ctime][\horizon] } \eye  }  \Bigg] + o(h) \\
\label{eq:2}
&=  h\parens*{\frac{\del}{\del \ctime} \kernel[\ctime][\ctime]- 2 \parens*{\frac{\del}{\del \ctime'}\kernelbase }(\ctime, \ctime)}\eye + o(h).
\end{align}

However, by \eqref{eq:covYdef}, we have

\begin{align}
\nn
\frac{\del}{\del \ctime} \kernel[\ctime][\ctime] &= \frac{\del}{\del \ctime} \parens*{  \aggtimesq[\ctime] \intdasq } \\
\nn
&= 2 \daggtime\aggtime\intdasq + \volatsq[\ctime], \\
\parens*{\frac{\del}{\del \ctime'}\kernelbase }(\ctime, \ctime) &= \frac{\del}{\del \ctime'} \parens*{  \aggtime[\ctime]\aggtime[\ctime'] \intdasq }\Bigg\vert_{\ctime' = \ctime} \\
&= \daggtime\aggtime\intdasq, 
\end{align} 

from which \eqref{eq:2} simplifies to

\begin{equation}
\label{eq:Sigmacapprox}
\Sigmac = h \volatsq[\ctime] \eye + o(h).
\end{equation}

We can now compute $\exof*{ \testf[\ctime+h][\Xsol[\ctime+h]] \given \Xsol = \point  }$ as follows:

\begin{align}
\nn
\exof*{ \testf[\ctime+h][\Xsol[\ctime+h]] \given \Xsol = \point  } &=  \Nconst  \parens*{\det \Sigmac}^{-\frac{1}{2}} \intR \testf[\ctime+h][\pointalt] \exp\parens*{-\frac{1}{2} \parens*{\pointalt - \muc}^\top \Sigmacinv\parens*{\pointalt - \muc}} \dd \pointalt \\
\label{eq:3}
&= \Nconst\parens*{\det \Sigmac}^{-\frac{1}{2}} \intR \testf[\ctime+h][\pointalt + \muc]  \exp\parens*{-\frac{1}{2} \pointalt^\top \Sigmacinv\pointalt } \dd \pointalt.
\end{align}

Invoking \cref{lem:centralid}, we see that \eqref{eq:3} can be evaluated as

\begin{align}
\label{eq:4}
\exof*{ \testf[\ctime+h][\Xsol[\ctime+h]] \given \Xsol = \point  } &=  \left.\exp\parens*{ \frac{1}{2} \del_{\pointalt}^\top \Sigmac \del_{\pointalt}}  \testf[\ctime+h][\pointalt + \muc]\right\vert_{\pointalt = 0}.
\end{align}

Since $\Sigmac = h\volatsq[\ctime]\eye + o(h) $ by \eqref{eq:Sigmacapprox}, expanding the power series $\exp\parens*{ \frac{1}{2} \del_{\pointalt}^\top \Sigmac \del_{\pointalt}} $ and ignoring every $o(h)$ terms, \eqref{eq:4} becomes

\begin{align}
\nn
\exof*{ \testf[\ctime+h][\Xsol[\ctime+h]] \given \Xsol = \point  } &=  \left. \parens*{\testf[\ctime+h][\pointalt + \muc] + \frac{h\volatsq[\ctime]}{2}\Laplace \testf[\ctime+h][\pointalt + \muc]}\right\vert_{\pointalt = 0} + o(h) \\
\label{eq:5}
&= \testf[\ctime+h][ \muc] + \frac{h\volatsq[\ctime]}{2}\Laplace \testf[\ctime+h][ \muc] + o(h).
\end{align}

Recalling from \eqref{eq:mucapprox} that $\muc = \point + h \parens*{  \St^\top \Sigmasolinv\parens*{ \point - \meansol  } + \dmeansol } + o(h)$, the Taylor expansion in the $\point$ variable for $\testf$ shows that 

\begin{align}
\nn
\exof*{ \testf[\ctime+h][\Xsol[\ctime+h]] \given \Xsol = \point  } &= \testf[\ctime+h][\point] + h\parens*{ \frac{\volatsq[\ctime]}{2}\Laplace \testf[\ctime+h][ \point] + \inner*{ \nabla \testf[\ctime+h][\point]}{ \St^\top \Sigmasolinv\parens*{ \point - \meansol  } + \dmeansol }} + o(h)
\end{align}

whence 

\begin{align}
\nn
\lim_{ h\to 0}  \frac{  \exof*{ \testf[\ctime+h][\Xsol[\ctime+h]] \given \Xsol = x }  - \testf  }{h} = \frac{\del}{\del \ctime} \testf +  \frac{\volatsq[\ctime]}{2} \Laplace \testf +  \inner*{  \nabla \testf }{ \St^\top \Sigmasolinv\parens*{ \point - \meansol  } + \dmeansol  }.
\end{align}

This is exactly \eqref{eq:gensde} with $\GSBf \subs \St^\top \Sigmasolinv\parens*{ \point - \meansol  } + \dmeansol$, which concludes the proof for \eqref{eq:GSB-forward-sol} and \eqref{eq:GSB-forward}.

Finally, by \cite[(4.2)]{leonard2013survey}, the optimal drift $\GSBf$ is a \emph{gradient field}: 

\begin{equation}
\GSBf = \nabla \psi(\ctime,\point)
\end{equation}

for some function $\psi: \R^+ \times \R^\vdim \to \R$, implying that $\St^\top\Sigmasolinv$ must be symmetric.
\end{proof}

%----------------------------------------------------------------------
%%% ADDITIONAL EXPERIMENTIONS
%----------------------------------------------------------------------
\section{Additional Details for \cref{sec:experiments}}
\label{app:appendix}

\subsection{Further Empirical Validation of Gaussian Approximation}
\label{app:empirical}

A central thesis of our paper is that Gaussian approximation provides a reasonable initialization for the objectives in \eqref{eq:loss-backward} and \eqref{eq:loss-forward}. The purpose of the current section is to empirically validate this Gaussian hypothesis via examining the \emph{marginal distributions} of the two real datasets we considered.

\begin{figure}
     \centering
     \begin{subfigure}[b]{0.78\textwidth}
         \centering
         \includegraphics[width=\textwidth]{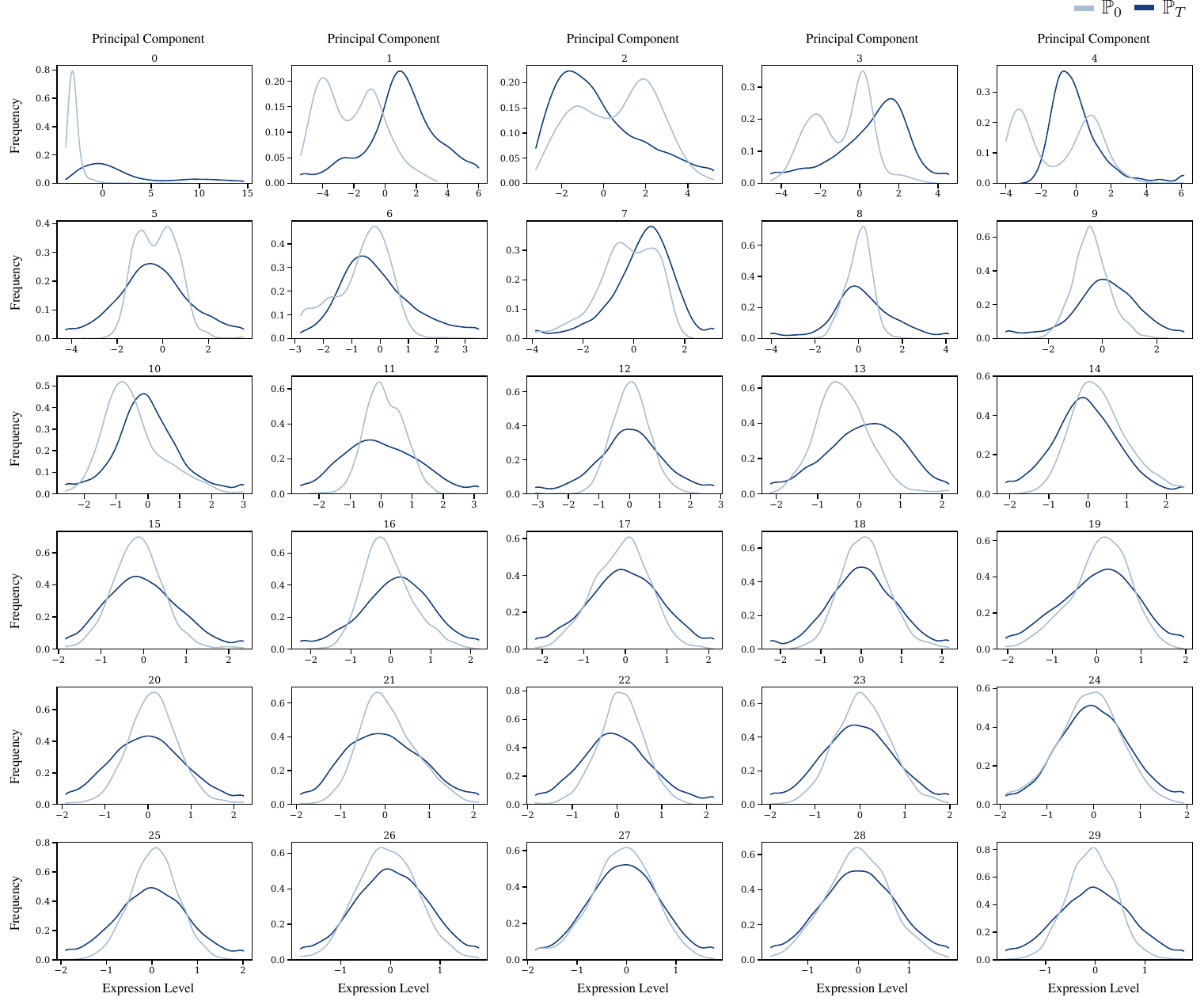}
         \caption{The first 30 principal components of the dataset from \citep{moon2019visualizing}.}
     \end{subfigure}
     \begin{subfigure}[b]{0.78\textwidth}
         \centering
         \includegraphics[width=\textwidth]{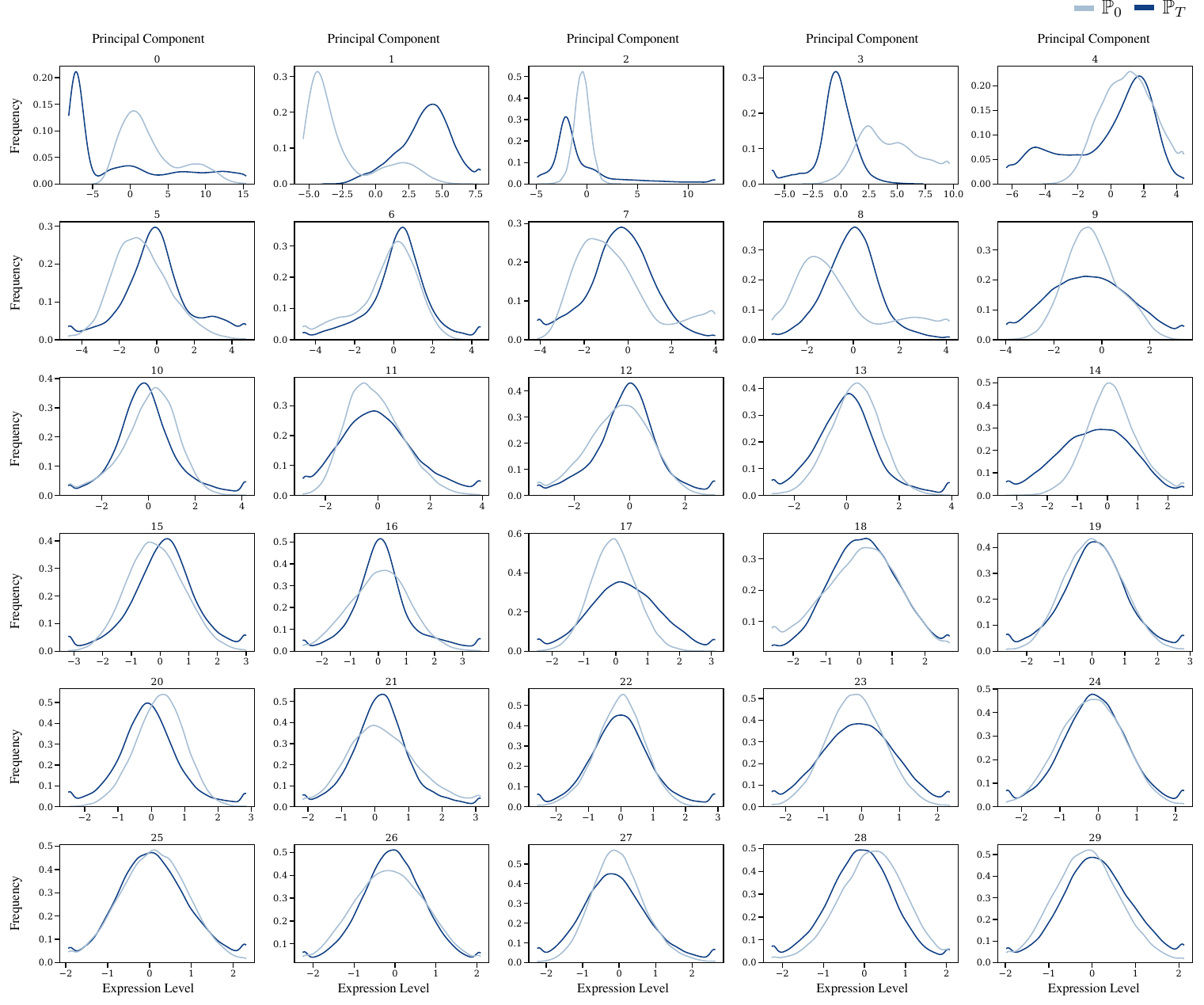}
         \caption{The first 30 principal components of the dataset from \citep{schiebinger2019optimal}.}
     \end{subfigure}
%     \begin{subfigure}[b]{0.8\textwidth}
%         \centering
%%         \includegraphics[width=\textwidth]{figures/fig_marginals_moon_genes.pdf}
%%         \caption{PHATE embedding hued by predicted lineage branch.}
%     \end{subfigure}
\caption{Empirical validation of the Gaussian approximation for the considered datasets.}
\label{fig:gaussian}
\end{figure}

We summarize our finding in \cref{fig:gaussian}, from which we can see that many marginal distributions (\eg the 8-29th principal components of both datasets) can be reasonably approximated by Gaussians \emph{provided that we take their mean and variances into account}. This is precisely the key feature of \acroalg, and thus provides empirical evidence favoring our approach.

%----------------------------------------------------------------------
%%% METHOD
%----------------------------------------------------------------------
%\vspace{-8pt}
\subsection{Dynamics Reconstruction via \acroalg}
%\vspace{-8pt}
\label{sec:methods}
%\textbf{Review of \ac{SB}.}
%In principle, one can take the reference process $\refpro$ in \eqref{eq:SB-updated} to be any \ac{SDE} of the form:% with drift and scale,
%\begin{align}
%\label{eq:general-refsde}
%\drefsde  = \tdrift  \dt + \volat \dWiener[\ctime]\equiv \tdriftbase \dt + \volatbase \dWiener[\ctime]
%\end{align}
%with some initial condition $\refsdebase_0$. In this case, %When $\refpro = \sdev\Wiener$, \eqref{eq:generalSB} reduces to the classical \eqref{eq:SB}.
%An important fact is that $\refpro$ enters the optimality condition of \eqref{eq:generalSB} only through the \emph{conditional} distributions of the form $\left.\refsde \right\vert \refsde[0], \refsde[\horizon]$ \citep{leonard2013survey}. %, we will ignore the initial condition for $\refsde$.
\subsubsection{Background on Schr\"odinger Bridges}
\label{sec:background}

Recall the general \ac{SB} problem \eqref{eq:YtSB}. It turns out that the solution to \eqref{eq:YtSB} is itself given by two coupled \acp{SDE} of the form \citep{leonard2013survey}
\begin{subequations}
\begin{eqnarray}
\label{eq:SB-sde-forward}
\dsde &= \parens*{\tdriftbase_t +  \volatbase_t \SBfbase_\ctime} \dt + \volatbase_t \dWiener[\ctime],\; \sde[0] \sim \distinit, \\%\; & \sde[0] \sim \Pinit, \\
\label{eq:SB-sde-backward}
\dsde &= \parens{\tdriftbase_t -  \volatbase_t \SBbbase_\ctime} \dt + \volatbase_t \dWiener[\ctime],\; \sde[\horizon] \sim \distend,%\; & \sde[\horizon] \sim \Pend.
\end{eqnarray}
\end{subequations}
where $\SBfbase_\ctime, \SBbbase_\ctime \from \R^\vdim \to \R^\vdim$ are two time-indexed smooth \emph{vector fields} called the optimal forward and backward drift, respectively, and \eqref{eq:SB-sde-backward} runs backward in time (\ie from $\horizon \to 0$). %and satisfies many interesting properties \citep{nelson1967dynamical,leonard2013survey,chen2021stochastic}.
~If we parametrize the forward drift by $\SBf$ and the backward drift by $\SBb$ with some parameters $\paramf,\paramb$, then the negative likelihood function for $\paramf$ and $\paramb$ can be expressed as \citep{chen2021likelihood}
\begin{subequations}
\begin{align}
\lossb =  \int_0^\horizon \mathbb{E}_{\eqref{eq:SB-sde-forward}}\Big[ \frac{1}{2}\norm{\SBbbase^{\paramb}_\ctime}^2 &+\volatbase \Div \SBbbase^{\paramb}_\ctime+ \inner{\SBfbase^{\paramf}_\ctime}{\SBbbase^{\paramb}_\ctime} \dt \Big\vert \sde[0] = \point_0 \Big] ,  \\
\lossf =  \int_0^\horizon \mathbb{E}_{\eqref{eq:SB-sde-backward}}\Big[ \frac{1}{2}\norm{\SBfbase^{\paramf}_\ctime}^2 &+\volatbase \Div \SBfbase^{\paramf}_\ctime + \inner{\SBbbase^{\paramb}_\ctime}{\SBfbase^{\paramf}_\ctime} \dt \Big\vert \sde[\horizon] = \point_{\horizon} \Big].
\end{align}
\end{subequations}
%where $\Div$ denotes the divergence operator w.r.t. the $\point$ variable: For any $v\from \R^\vdim \to \R^\vdim$, $\Div v(\point) \defeq \sum_{\coord=1}^{\vdim} \ddc v_\coord(\point)$.

\subsubsection{The \textsc{GSBflow} Algorithm}

Building on the closed-form solutions in \cref{sec:results}, we present an end-to-end learning paradigm that takes two marginal distributions $\distinit$, $\distend$ to output the reconstruction of the underlying stochastic dynamics $\Pmargin$. Because our framework relies on \acp{GSB}, we call our algorithm the \textbf{\acroalg}.

\textbf{Step 1: Moment estimates and \ac{GSB} initialization.} We first compute the means $\mu_0, \mu_\horizon$ and covariances $\Sigma_0, \Sigma_\horizon$ of the input distributions, and plug them into \eqref{eq:GSB-forward} and \eqref{eq:cmean-marginal-0}-\eqref{eq:csigma-marginal-T}. Note that these computations are done only \emph{once} for every dataset, and can be reused for all subsequent training.

\textbf{Step 2: Forward and backward pretraining.} Denoting by $\refapprox$ the measure of $\GSBfbase \dt + \volatbase_t \dWiener[\ctime]$ in \eqref{eq:GSB-forward}, we propose to minimize the objective
\begin{equation}
\label{eq:GSBflow}
% \tag{\acroalg}
\min_{ \Pinit = \distinit,\; \Pend = \distend } \KL{\Pmargin}{ \refapprox}.
\end{equation}
Following the framework of \citet{chen2021likelihood}, we see that the optimal solution to \eqref{eq:GSBflow} is given by two \acp{SDE} of the form:
%Following \eqref{eq:SB-sde-forward}-\eqref{eq:SB-sde-backward}, we see that the optimal solution to \eqref{eq:GSBflow} is given by two \acp{SDE} of the form:
\begin{subequations}
\begin{eqnarray}
\label{eq:GSB-sde-forward}
\dsde &= \parens*{\GSBfbase +  \volatbase_t \SBfbase_\ctime} \dt + \volatbase_t \dWiener[\ctime],\; & \sde[0] \sim \distinit, \\
\label{eq:GSB-sde-backward}
\dsde &= \parens*{\GSBfbase -  \volatbase_t \SBbbase_\ctime} \dt + \volatbase_t \dWiener[\ctime],\; & \sde[\horizon] \sim \distend,
\end{eqnarray}
\end{subequations}
where \eqref{eq:GSB-sde-backward} runs backward in time. After parameterizing $\SBfbase_\ctime$ and  $\SBbbase_\ctime$ by two neural networks $\SBf,\SBb$ with parameters $\paramf,\paramb$, the corresponding negative likelihood in \cref{sec:background} becomes

\begingroup
\allowdisplaybreaks
\begin{subequations}
\begin{align}
%\label{eq:SB-loss-forward}
\label{eq:loss-backward}
\lossb =  \int_0^\horizon \mathbb{E}_{\eqref{eq:GSB-sde-forward}}\Big[ \frac{1}{2}\norm{\SBbbase^{\paramb}_\ctime}^2 &+\volatbase \Div \SBbbase^{\paramb}_\ctime + \inner{\SBfbase^{\paramf}_\ctime}{\SBbbase^{\paramb}_\ctime} \dt \Big\vert \sde[0] = \point_0 \Big] ,  \\
\label{eq:loss-forward}
\lossf =  \int_0^\horizon \mathbb{E}_{\eqref{eq:GSB-sde-backward}}\Big[ \frac{1}{2}\norm{\SBfbase^{\paramf}_\ctime}^2 &+\volatbase \Div \SBfbase^{\paramf}_\ctime + \inner{\SBbbase^{\paramb}_\ctime}{\SBfbase^{\paramf}_\ctime} \dt \Big\vert \sde[\horizon] = \point_{\horizon} \Big].
\end{align}
\end{subequations}
\endgroup

\looseness-1 Following existing work on training \ac{SB}-based objectives \citep{chen2021likelihood,de2021diffusion, vargas2021solving}, we propose to initialize $\tilde{\paramf}_{0}, \tilde{\paramb}_{0}$ such that $\SBf[\ctime][\point][\tilde{\paramf}_{0}], \SBb[\ctime][\point][\tilde{\paramb}_{0}] \equiv 0$, which can be easily achieved by zeroing out the last layer of the corresponding neural networks. In this case, estimating the conditional expectations in both \eqref{eq:loss-backward}-\eqref{eq:loss-forward} reduces to simulating \eqref{eq:GSB-forward} \emph{conditioned} on the given start or end data points. Thanks to our closed-form expressions, this can be easily achieved by drawing Gaussian variables with mean and covariance prescribed in  \eqref{eq:GSB-sde-forward}-\eqref{eq:GSB-sde-backward}. The pretraining procedure is summarized in \cref{alg:pretraining}.

\newacro{IPF}{iterative proportional fitting}

\textbf{Step 3: Alternating minimization.} After the pretraining phase, we switch to minimizing \eqref{eq:loss-backward}-\eqref{eq:loss-forward} with general drifts in \eqref{eq:GSB-sde-forward}-\eqref{eq:GSB-sde-backward}. We carry out this step in an alternating fashion: Since the bottleneck of our framework is to simulate the trajectories of \acp{SDE}, we perform several gradient updates for one parameter before drawing another batch of samples. See \cref{alg:GSBflow} for a summary, and \cref{fig:overview} for an illustration.

\begin{minipage}[t]{0.46\textwidth}
\begin{algorithm}[H]
   \caption{Forward and Backward Pretraining}
   \label{alg:pretraining}
\begin{algorithmic}
   \STATE {\bfseries Input:} Marginal distributions $\distinit, \distend$, initial parameters $\tilde{\paramf}_{0}, \tilde{\paramb}_{0}$ such that $\SBf[\ctime][\cdot\ ][\tilde{\paramf}_0] = \SBb[\ctime][\cdot\ ][\tilde{\paramb}_0] \equiv 0$, iteration counts $\pretriterf, \pretriterb$, learning rates $\lrf,\lrb$ %\dots  data $x_i$, size $m$
   \STATE {\bfseries Output:} Pretrained parameters $\paramf_0,\paramb_0$
%   \STATE {\bfseries Construct \eqref{eq:GaussianSB}:} Compute the means and covariances of $\distinit, \distend$.
%   \REPEAT
   \STATE Initialize $\paramf_0 \subs \tilde{\paramf}_0$, $\paramb_0 \subs \tilde{\paramb}_0$.
   \FOR{$k=1$ {\bfseries to} $\pretriterb$}   
   \STATE Sample $\sde$ from \eqref{eq:cmean-marginal-0}-\eqref{eq:csigma-marginal-0} with $\point_0 \sim \distinit$
   \STATE Compute $\lossb$ via \eqref{eq:loss-backward}
   \STATE Update $\paramb_0 \subs \paramb_0 - \lrb\nabla \lossb[\point_0][\paramb_0]$
   \ENDFOR
   \FOR{$k=1$ {\bfseries to} $\pretriterf$}   
   \STATE Sample $\sde$ from \eqref{eq:cmean-marginal-T}-\eqref{eq:csigma-marginal-T} with $\point_\horizon \sim \distend$
   \STATE Compute $\lossf$ via \eqref{eq:loss-forward}
   \STATE Update $\paramf_0 \subs \paramf_0 -\lrf \nabla \lossf[\point_\horizon][\paramf_0]$
   \ENDFOR
%   \UNTIL{$noChange$ is $true$}
\end{algorithmic}
\end{algorithm}
\end{minipage}
\hfill
\begin{minipage}[t]{0.46\textwidth}
\begin{algorithm}[H]
   \caption{\acroalg}
   \label{alg:GSBflow}
\begin{algorithmic}
   \STATE {\bfseries Input:} Marginal distributions $\distinit, \distend$, pretrained parameters $\paramf_{0}, \paramb_{0}$, caching frequency $\caching$, iteration counts $\inneriter, \outeriter$, learning rates $\lrf,\lrb$ %\dots  data $x_i$, size $m$
   \STATE {\bfseries Output:} Optimal forward and backward drifts $\SBfbase_\ctime(\cdot), \SBbbase_\ctime(\cdot)$ for \eqref{eq:GSBflow}
%   \STATE {\bfseries Construct \eqref{eq:GaussianSB}:} Compute the means and covariances of $\distinit, \distend$.
%   \REPEAT
\smallskip
   \STATE Initialize $\paramf \subs \paramf_0$, $\paramb \subs \paramb_0$.
   \FOR{$k=1$ {\bfseries to} $\outeriter$} 
   \FOR{$j=1$ {\bfseries to} $\inneriter$}   
   \IF{$j \mod \caching = 0$}
   \STATE Simulate \eqref{eq:GSB-sde-forward} with $\point_0 \sim \distinit$
   \ENDIF
   \STATE Compute $\lossb$ via \eqref{eq:loss-backward}
   \STATE Update $\paramb \subs \paramb - \lrb\nabla \lossb$
   \ENDFOR
   \FOR{$j=1$ {\bfseries to} $\inneriter$}   
   \IF{$j \mod \caching = 0$}
   \STATE Simulate \eqref{eq:GSB-sde-backward} with $\point_\horizon \sim \distend$
   \ENDIF
   \STATE Compute $\lossf$ via \eqref{eq:loss-forward}
   \STATE Update $\paramf \subs \paramf - \lrf\nabla \lossf$
   \ENDFOR
   \ENDFOR
%   \UNTIL{$noChange$ is $true$}
\end{algorithmic}
\end{algorithm}
\end{minipage}

\subsection{Further Experimental Results} 
\label{app:more_results}

%\begin{figure}
% \centering
%
%\end{figure}

\begin{figure}[!htb]
    \centering
     \begin{subfigure}[b]{0.4\textwidth}
         \centering
         \includegraphics[width=\textwidth]{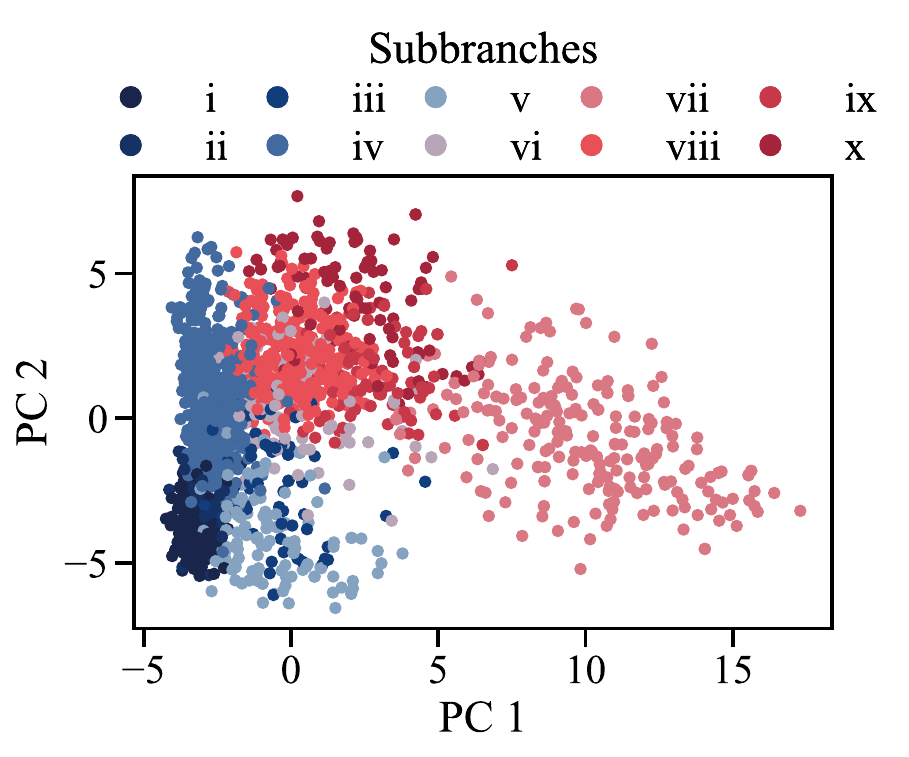}
         \caption{Data}
     \end{subfigure}
     \begin{subfigure}[b]{0.4\textwidth}
         \centering
         \includegraphics[width=\textwidth]{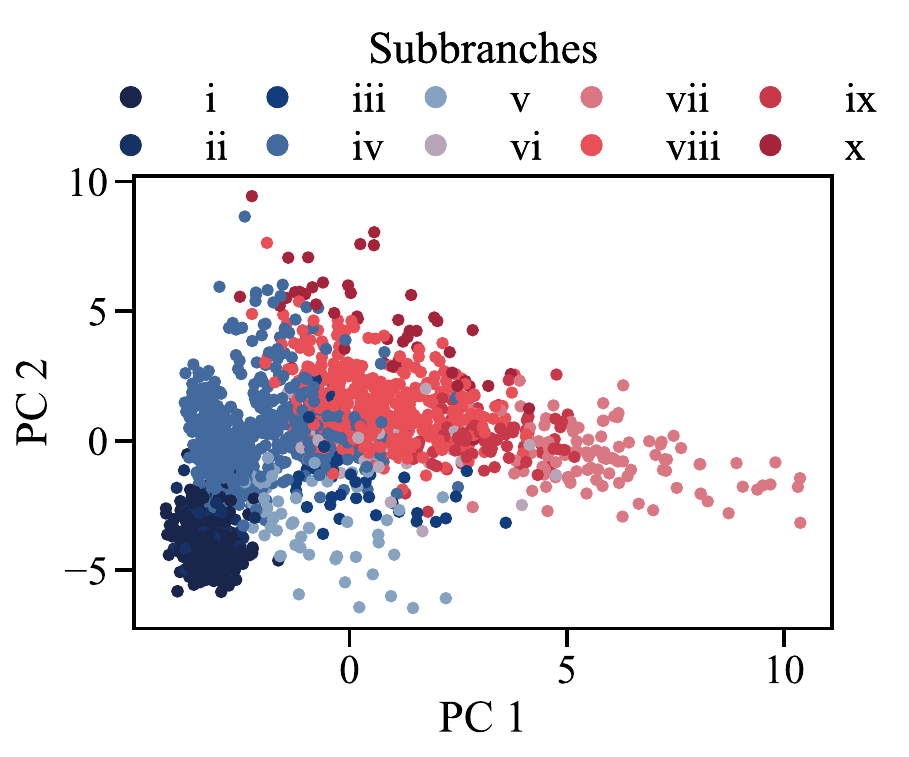}
         \caption{\textsc{GSBflow}}
     \end{subfigure}
    \caption{PCA embedding of the (a) \citep{moon2019visualizing} data and (b) the \textsc{GSBflow} predictions colored by the lineage subbranch class.}
    \label{fig:res_evo_subbranches}
\end{figure}

\cref{fig:res_evo_subbranches} shows the \acroalg's prediction of cell lineages on the \citet{moon2019visualizing} dataset but with annotated predictions to cellular sublineages. The result agrees with our observations on the \citet{moon2019visualizing} dataset present in the main text, where \acroalg\ succeeds in learning the cell's differentiation into various cell lineages and capturing biological heterogeneity on a more macroscopic level.

\subsection{Single-Cell Datasets} 

\label{app:datasets}
We evaluate \textsc{GSBflow} on multiple datasets. This includes synthetic population dynamics, whose results are described in \cref{sec:synthetic}, as well as dynamics of single-cells of a human developmental process, which we cover in \cref{sec:cell}. In the following, we describe the data collection and data preprocessing steps.

%\label{app:cell_dataset}
In biology, developmental processes involve complex tasks such as tissue and organ development, body axis formation, cell division, and cell differentiation, \ie the development of stem cells into functional cell types.
One prototype of such a process is the differentiation of \acdefp{ESC} into distinct lineages such as hematopoietic, cardiac, neural, pancreatic, hepatocytic and germ.
It is possible to approximate this development \textit{in vitro} via embryoid bodies (EBs) \citep{martin1975}, three-dimensional aggregates of pluripotent stem cells, including ESCs \citep{shamblott2009derivation}.
Recently, an scRNA-seq analysis is conducted by \citet{moon2019visualizing} in order to unveil the developmental trajectories, as well as cellular and molecular identities through which early
lineage precursors emerge from human ESCs.
The dataset of \citet{moon2019visualizing} can be found online via \href{https://data.mendeley.com/datasets/v6n743h5ng}{Mendeley Data (V6N743H5NG)}.%\footnote{Dataset available via \url{https://data.mendeley.com/datasets/v6n743h5ng}.}.

In the second task we consider, the natural developmental process is inverted and instead we study the reprogramming of  somatic cells into \acp{iPSC}. In a recent study, \citet{schiebinger2019optimal} unveiled this process in depth by measuring 315,000 single-cell RNA sequencing (scRNA-seq) profiles, collected at half-day intervals across 18 days.

In the following, we provide details for the the preprocessing of the raw scRNA-seq data as well as the lineage branch analysis extracting the functional cell types emerging in this developmental process.
This dataset can be found via \href{https://www.ncbi.nlm.nih.gov/geo/query/acc.cgi?acc=GSE122662}{\texttt{GEO: GSE122662}}.

\subsubsection{Data Preprocessing}
To preprocess the data, we follow the analysis of \citet{moon2019visualizing} as well as \citet{luecken2019current}. For the analysis, we invoke the Python package \texttt{scanpy} \citep{wolf2018scanpy}.

\begin{figure}[!htb]
    \centering
    \begin{minipage}{.49\textwidth}
        \centering
        \includegraphics[width=.6\linewidth]{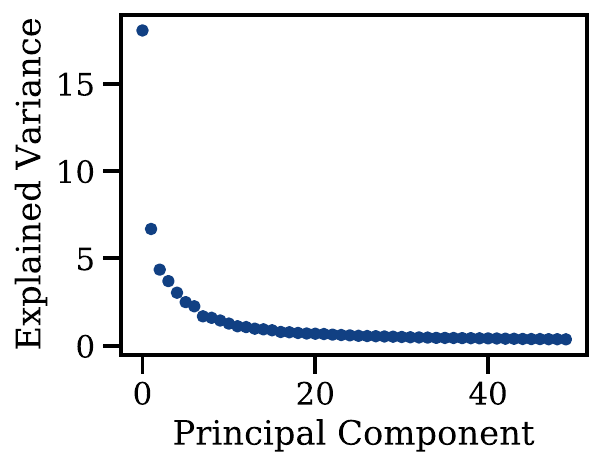}
        \caption{Proportion of explained variance per PC of the embryoid body scRNA-seq data after preprocessing \citep{moon2019visualizing}.}
        \label{fig:moon_expl_variance}
    \end{minipage}\hfill
    \begin{minipage}{0.49\textwidth}
        \centering
        \includegraphics[width=.6\linewidth]{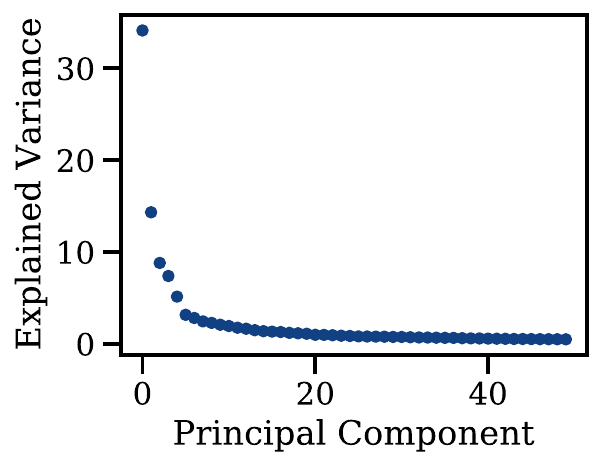}
        \caption{Proportion of explained variance per PC of MEFs reprogramming into iPSCs after preprocessing \citep{schiebinger2019optimal}.}
        \label{fig:schiebinger_expl_variance}
    \end{minipage}
\end{figure}

\citet{moon2019visualizing} take measures of approximately 31,000 cells over a 27-day differentiation time course, comprising gene expression matrices and barcodes, i.e., DNA tags for identifying reads originating from the same cell. Subsequently, the measured cells are filtered in a quality control stage, their gene expression levels normalized and further processed in a feature selection step, where only highly-differentiated genes are selected.
The resulting data is then visualized using a standard PCA as well as the dimensionality reduction method of PHATE \citep{moon2019visualizing} in order to extract biological labels.

The quality control of the data is based on the number of counts per barcode (count depth), the number of genes per barcode, and the fraction of counts from mitochondrial genes per barcode. In our experiments, we only keep cells with at least 4000 and at most 10000 counts, as well as more than 550 expressed genes and less than $20\%$ of mitochondrial counts, as a high fraction is indicative of cells whose cytoplasmic mRNA has leaked out through a broken membrane \citep{luecken2019current}.
For the subsequent analysis, we further screen out genes which are expressed in less than 10 genes.
After quality control procedure described above, the dataset consists of 15150 cells and 17945 genes.
We then normalize each cell by total counts over all genes and logarithmize the data matrix. We extract 4000 highly variable genes (HVG) the 10X genomics preprocessing software \texttt{Cell Ranger} \citep{zheng2017massively} to further reduce the dimensionality of the dataset and include only the most informative genes.
Given the resulting data matrix with 15150 cells and 4000 genes,% across 5 different time points, 
we compute a corresponding low-dimensional embedding using PCA. Figure~\ref{fig:moon_expl_variance} shows the proportion of explained variance of each principal component (PC). We use the first 20 or 30 PCs for predicting population dynamics using \acroalg.% and the forward method.
This is in alignment with previous analysis of developmental trajectories
%, which use 5 \citep{tong2020trajectorynet} and 30 PCs \citep{schiebinger2019}, respectively.
~which uses 30 PCs \citep{schiebinger2019optimal}. The same analysis on executed on the \acp{MEF} reprogramming dataset by \citet{schiebinger2019optimal}.

\subsubsection{Lineage Branch Analysis of the Embryoid Body scRNA-Seq Data} \label{app:lineage_analysis}

\begin{figure}
     \centering
     \begin{subfigure}[b]{0.8\textwidth}
         \centering
         \includegraphics[width=\textwidth]{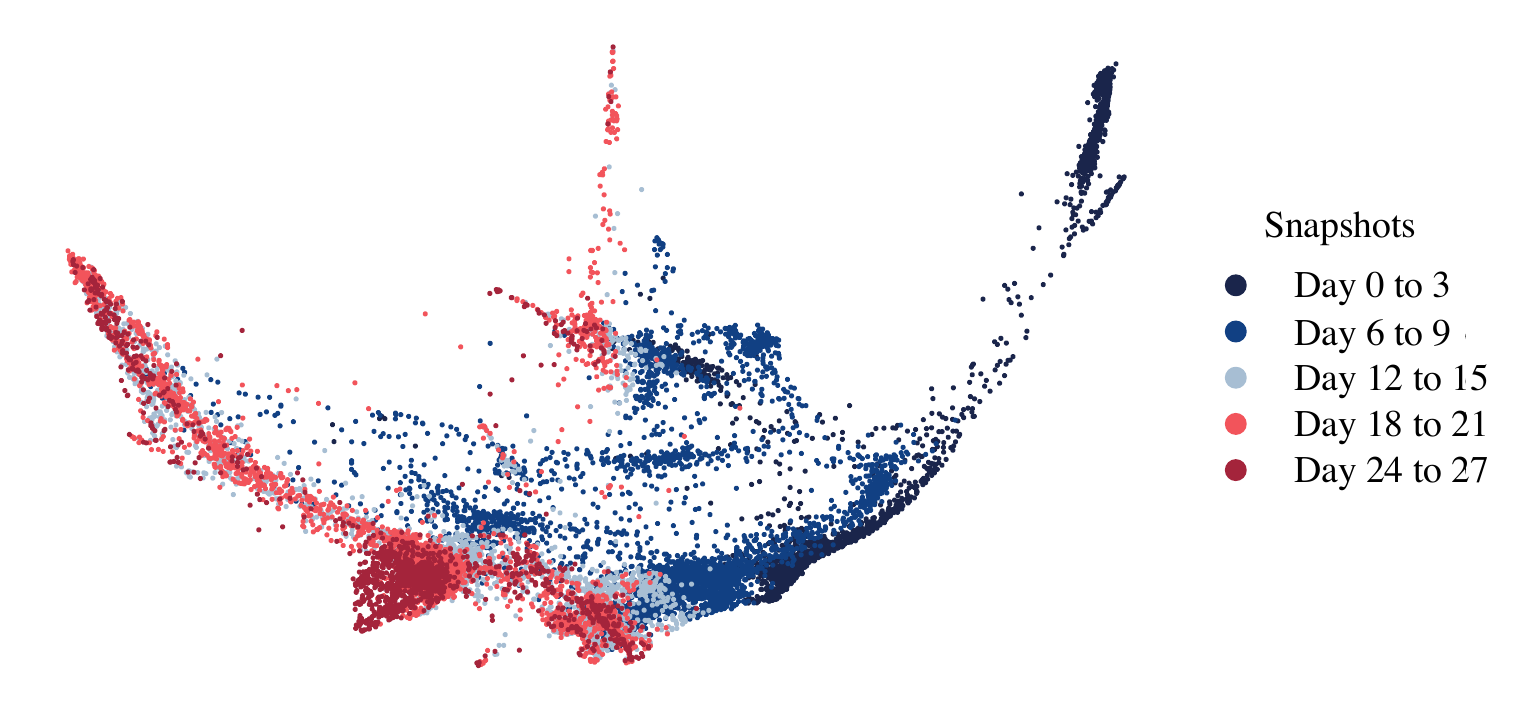}
         \caption{PHATE embedding colored by time of snapshot.}
     \end{subfigure}
     \begin{subfigure}[b]{0.8 \textwidth}
         \centering
         \includegraphics[width=\textwidth]{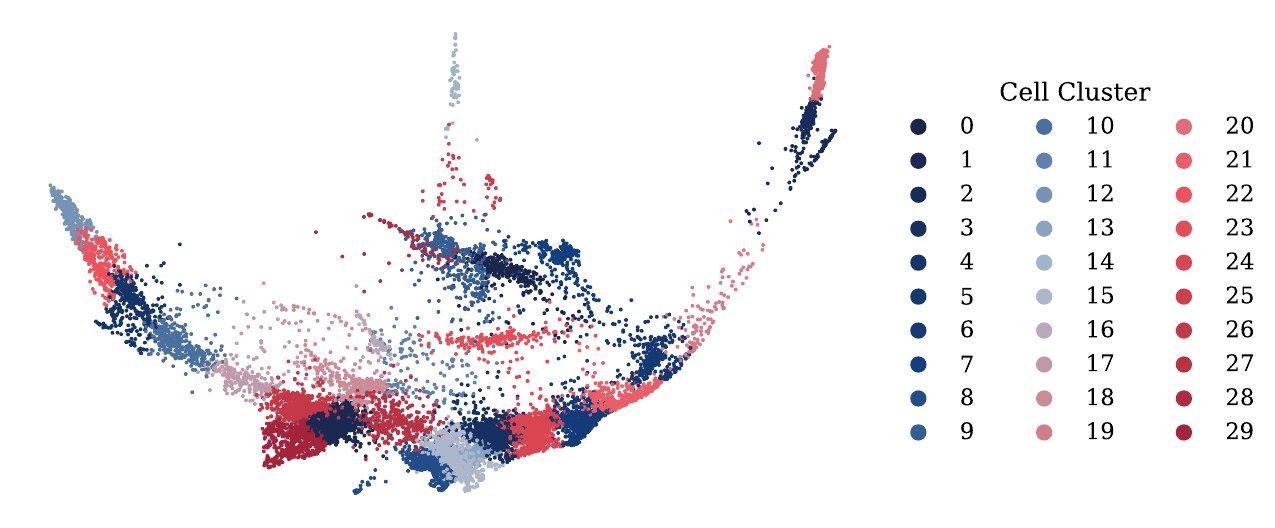}
         \caption{PHATE embedding hued by k-Means clustering ($k=30$).}
     \end{subfigure}
     \begin{subfigure}[b]{0.8\textwidth}
         \centering
         \includegraphics[width=\textwidth]{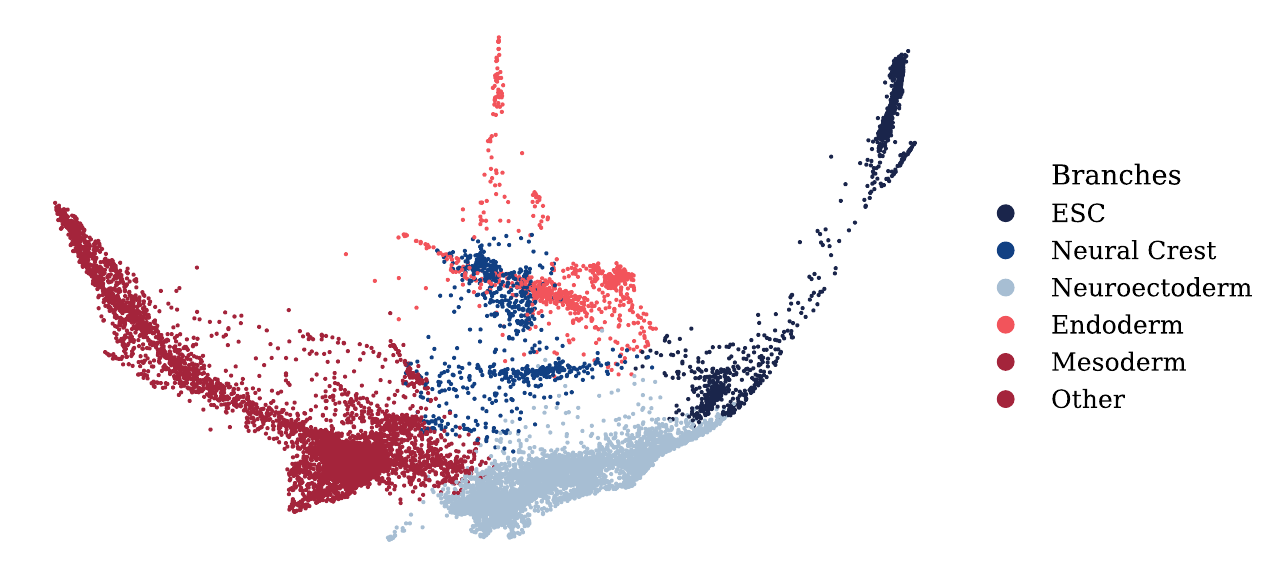}
         \caption{PHATE embedding hued by predicted lineage branch.}
     \end{subfigure}
\caption{Analysis of embryoid body scRNA-seq data based on PHATE embedding \citep{moon2019visualizing}. Lineage branches are determined based on contiguous k-means clusters.}
\label{fig:moon_analysis}
\end{figure}

In order to annotate the developmental process and detect lineage branches originating from the differentiation of \acp{ESC}, we again follow the analysis of \citet{moon2019visualizing}.
With a 10-dimensional PHATE embedding of the embryoid body scRNA-seq data (see the first two PHATE components in Fig.~\ref{fig:moon_analysis}a), we segment the dataset into 30 clusters using k-means.
%PHATE is a non-linear dimensionality reduction method capturing a denoised representation of both local and global structure of a dataset \citep{moon2019visualizing}.
We then assign the resulting cluster to a lineage subbranch (\textit{i} - \textit{x}), using the following assignment of subbranch to cluster identification (see Fig.~\ref{fig:moon_analysis}b):
\begin{enumerate}[noitemsep, topsep=0pt, label=\roman*., , labelsep=*]
\begin{multicols}{3}
\multicollinenumbers
    \item 2, 20
    \item 5, 19
    \item 9, 11, 23
    \item 3, 6, 8, 13, 15, 21, 24 \vspace{1cm} 
    \item 0, 7, 14, 25, 28
    \item 16, 18, 27
    \item 4, 10, 12, 17, 22
    \item 1
    \item 26
    \item 29.
\end{multicols}
\end{enumerate}
Then, subbranches are summarized to lineage branches using the assignment in \cite[Suppl. Note 4]{moon2019visualizing}:
\begin{itemize}[noitemsep, topsep=0pt, labelsep=*]
\begin{multicols}{2}
\centering
\multicollinenumbers
    \item[ESC.] i, ii
    \item[Neural Crest.] iii
    \item[Neuroectoderm.] iv
    \item[Endoderm.] v
    \item[Mesoderm.] vi, vii
    \item[Other.] viii, ix, x.
\end{multicols}
\end{itemize}
The resulting lineage branch annotation of the embryoid body scRNA-seq data can be found in Figure~\ref{fig:moon_analysis}c.

\subsection{Experimental Details} 
\label{app:experiments}
In the following, we describe model components, as well as provide details on networks architectures and hyperparameters used.

\subsubsection{The Reference Processes: \acp{VESDE} and sub\textendash\acp{VPSDE}}
\label{app:vesde}

We adopt the \ac{VESDE} \citep{song2020score} as the reference process of the \ac{GSB}

\begin{equation}
    \label{eq:vesde}
    \drefsde = \volat \dWiener,
\end{equation}

where

\begin{equation}
\nonumber
\volat = \sdev_{\min} \parens*{\frac{\sdev_{\max}}{\sdev_{\min}}}^\ctime\sqrt{2\log \frac{\sdev_{\max}}{\sdev_{\min}}}.
\end{equation}

Here, $\sdev_{\min},\sdev_{\max} \in \R^+$ are two hyperparameters that we sweep for each dataset. The corresponding $\qv$ in \cref{tab:examples} is $\sdev_{\min}^2 \parens*{\frac{\sdev_{\max}}{\sdev_{\min}}}^{2\ctime}$, from which one can easily compute all the other functions, and hence \eqref{eq:GSB-forward} as well as \eqref{eq:cmean-marginal-0}-\eqref{eq:csigma-marginal-T}.

\newmacro{\bmin}{\beta_{\min}}
\newmacro{\bmax}{\beta_{\max}}
For existing \ac{SB}-based methods, we additionally implement the sub\textendash\ac{VPSDE} \citep{song2020score}, which is determined by a linear function $\beta_\ctime \defeq \bmin + \ctime(\bmax-\bmin)$, where as in \ac{VESDE}, $\bmin,\bmax \in \R^+$ are two hyperparameters that we sweep for each dataset. The reference \ac{SDE} in \eqref{eq:linearsde} is given by 

\begin{align*}
    \drift &= -\frac{1}{2} \beta_\ctime, \\
    \volat &= \sqrt{\beta_\ctime \parens*{1- e^{ -2\int_0^\ctime \beta_{\ctimealt}\dd \ctimealt }}}.
\end{align*}

\subsubsection{Network Architectures} \label{app:architecure}
Forward and backward policies $\SBf,\SBb$ are time-indexed functions parameterizing the optimal forward and backward drift. Throughout the experiments, we parameterize both $\SBf,\SBb$ via multi-layer perceptrons (MLP). The network architectures change with the complexity of the task. We use four hidden layers of size 128 with sigmoid linear units (SiLU, Swish) as activation function for synthetic data and modeling embryoid body development \citep{moon2019visualizing}. In the MEF reprogramming task \citep{schiebinger2019optimal} we apply a MLP with five hidden layers.

\subsubsection{Hyperparameters and Training} \label{app:hyperparam}
For experiments on synthetic data as well as on the task of \citep{schiebinger2019optimal}, we train $\SBf,\SBb$ with batch size 1024. For the task of \citet{moon2019visualizing}, we set the batch size to 512. For all experiments, we use the Adam optimizer \citep{kingma2014adam} with learning rate $\text{lr} = 0.0002$ ($\beta_1 = 0.5$, $\beta_2 = 0.9$).
Further, we use exponential moving average (EMA) with the decay rate of 0.99.

The hyperparameters of each SDE class, i.e., VE SDE and sub-VP SDE are chosen based on the underlying dataset. On synthetic data, we set $\sigma_{\max} = 2$ in \eqref{eq:vesde}, for the single-cell tasks we use $\sigma_{\max} = 10$. For sub-VP SDE we set $\beta_{\max} = 4$ in synthetic data setting, and $\beta_{\max} = 20$ in single-cell experiments.
In all experiments, we assume a continuous time variable $t$.

\subsubsection{Additional Evaluation of Single-Cell Experiments}

Besides evaluating how well \textsc{GSBflow} resembles the spatio-temporal dynamics, we analyze its ability to capture biological heterogeneity.
Serving as an \emph{in vitro} model of early embryogenesis, embryoid bodies differentiation captures the development of \acp{ESC} into mesoderm, endoderm, neuroectoderm, neural crest and others.
Using an initial $k$-means clustering ($k=30$) and following \citet[Fig. 6, Suppl. Note 4]{moon2019visualizing}, we compute lineage branch classes (\cref{fig:moon_analysis}c) for all cells in a 10-dimensional embedding space using PHATE, a non-linear dimensionality reduction method capturing a denoised representation of both local and global structure of a dataset (\cref{fig:moon_analysis}b).
For details, see \cref{app:lineage_analysis}.

We then train a $k$-NN classifier ($k=5$) to infer the lineage branch class based on a 30-dimensional PCA embedding of a cell (ESC: 0, neural crest: 1, neuroectoderm: 2, endoderm: 3, mesoderm: 4, other: 5).
We analyze the captured lineage branch heterogeneity of \textsc{GSBflow}'s predictions by computing the lineage branch class of each cell using the $k$-NN classifier. The predicted populations colored by the estimated lineage branch as well as the data with the true lineage branch labels are visualized in \cref{fig:all_results}c. We study a more fine-grained resolution of lineage branches into subbranches in \cref{app:more_results} (\cref{fig:res_evo_subbranches}).
This analysis further demonstrates \textsc{GSBflow}'s ability to learn cells' differentiation into various lineages and to capture biological heterogeneity on a more macroscopic level.

\section{Reproducibility}
An implementation of \textsc{GSBflow} is available at \href{https://github.com/bunnech/gsbflow}{\url{https://github.com/bunnech/gsbflow}}.

\vfill

\end{document}